\documentclass[twocolumn]{svjour3}
\journalname{}
\smartqed  
\usepackage{graphicx}
\usepackage{amsmath}
\usepackage{amssymb}
\usepackage{fix-cm}
\usepackage{authblk}
\usepackage{enumitem}
\usepackage{pgffor}
\usepackage{graphicx}
\usepackage{amsmath}
\usepackage{amssymb}
\usepackage{overpic}
\usepackage{xcolor}
\usepackage{xspace}
\usepackage{multirow}
\usepackage[numbers,square,sort&compress]{natbib}
\usepackage{textcomp}
\usepackage{tikz}
\usepackage{tabularx}
\usepackage{multirow}
\usepackage{listings}
\usepackage[export]{adjustbox}
\usepackage{tikz}
\usepackage{pgffor}
\usepackage{caption}

\usepackage{dcolumn}
\usepackage{times}
\usepackage{enumitem}
\usepackage{pgffor}
\usepackage{graphicx}
\usepackage{amsmath}
\usepackage{amssymb}
\usepackage{overpic}
\usepackage{xcolor}
\usepackage{xspace}
\usepackage{multirow}
\usepackage{epstopdf}
\usepackage{array}
\usepackage[labelformat=empty]{subfig}
\usepackage[numbers,square,sort&compress]{natbib}
\usepackage{hyperref}

\newcommand{\real}{\mathbb{R}}

\newcommand{\bc}{\mathbf{c}}
\newcommand{\bd}{\mathbf{d}}

\newcommand{\bff}{\mathbf{f}}

\newcommand{\bp}{\mathbf{p}}

\newcommand{\bw}{\mathbf{w}}
\newcommand{\bx}{\mathbf{x}}
\newcommand{\by}{\mathbf{y}}

\newcommand{\IFVSIFT}{FV-SIFT\xspace}

\newcommand{\PM}[1]{{\footnotesize $\pm$#1}}
\newcommand{\pmt}[1]{{\tiny $\pm$#1}}
\newcommand{\BF}[1]{\textbf{#1}}

\newcommand{\off}[1]{}

\newcommand{\dcnn}{FV-CNN\xspace}
\newcommand{\rcnn}{FC-CNN\xspace}
\newcommand{\kt}{KTH-TIPS\xspace}
\newcommand{\ktb}{KTH-T2b\xspace}
\newcommand{\kthb}{KTH-T2b\xspace}

\tikzset{
 image label/.style={
   fill=white,
   text=black,
   font=\footnotesize,
   anchor=south east,
   xshift=-0.1cm,
   yshift=0.1cm,
   at={(0,0)}
 }
}

\makeatletter
\DeclareRobustCommand\onedot{\futurelet\@let@token\@onedot}
\def\@onedot{\ifx\@let@token.\else.\null\fi\xspace}

\newcommand{\eg}{\emph{e.g}\onedot}

\newcommand{\ie}{\emph{i.e}\onedot}
\newcommand{\etal}{\emph{et al}\onedot}

\newcolumntype{H}{>{\setbox0=\hbox\bgroup}c<{\egroup}@{}}

\begin{document}
\title{Deep filter banks for texture recognition, description, and segmentation}
\author{
        Mircea Cimpoi \and      
        Subhransu Maji \and
        Iasonas Kokkinos \and
        Andrea Vedaldi
}

\institute{%
Mircea Cimpoi \at
University of Oxford \\
\email{mircea@robots.ox.ac.uk}
\and
Subhransu Maji \at
University of Massachusetts Amherst \\
\email{smaji@cs.umass.edu}
\and
Iasonas Kokkinos \at
CentraleSup\'elec / INRIA-Saclay \\
\email{iasonas.kokkinos@ecp.fr}
\and
Andrea Vedaldi \at
University of Oxford \\
\email{vedaldi@robots.ox.ac.uk}
}

\maketitle
\begin{abstract}
Visual textures have played a key role in image understanding because they convey important semantics of images, and because texture representations that pool local image descriptors in an orderless manner have had a tremendous impact in diverse applications. In this paper we make several contributions to texture understanding. First, instead of focusing on texture instance and material category recognition, we propose a human-interpretable vocabulary of texture attributes to describe common texture patterns, complemented by a new \emph{describable texture dataset} for benchmarking. Second, we look at the problem of recognizing materials and texture attributes in realistic imaging conditions, including when textures appear in clutter, developing corresponding benchmarks on top of the recently proposed OpenSurfaces dataset. Third, we revisit classic texture represenations, including bag-of-visual-words and the Fisher vectors, in the context of deep learning and show that these have excellent efficiency and generalization properties if the convolutional layers of a deep model are used as filter banks. We obtain in this manner state-of-the-art performance in numerous datasets well beyond textures, an efficient method to apply deep features to image regions, as well as benefit in transferring features from one domain to another.
\end{abstract}

\section{Introduction}\label{s:intro}

Visual representations based on orderless aggregations of local features, which were originally developed as texture descriptors, have had a widespread influence in image understanding. These models include cornerstones such as the histograms of vector quantized filter responses of Leung and Malik~\cite{Leung96} and later generalizations such as the bag-of-visual-words model of Csurka~\etal~\cite{csurka04visual} and the Fisher vector of~Perronnin~\etal~\cite{perronnin07fisher}. These and other texture models have been successfully applied to a huge variety of visual domains, including problems closer to ``texture understanding'' such as material recognition, as well as domains such as object categorization and face identification that share little of the appearance of textures.

This paper makes three contributions to texture understanding. The first one is to add a new \emph{semantic dimension} to the problem. We depart from most of the previous works on visual textures, which focused on texture identification and material recognition, and look instead at the problem of \emph{describing generic texture patterns}. We do so by developing a vocabulary of forty-seven \emph{texture attributes} that describe a wide range of texture patterns; we also introduce a large dataset annotated with these attributes which we call the \emph{describable texture dataset} (Sect.~\ref{s:dtd}). We then study whether texture attributes can be reliably estimated from images, and for what tasks are they useful. We demonstrate in particular two applications (Sect.~\ref{s:exp-dtd-attr}): the first one is to use texture attributes as dimensions to organise large collections of texture patterns, such as textile, wallpapers, and construction materials for search and retrieval. The second one is to use texture attributes as a compact basis of visual descriptors applicable to other tasks such as material recognition.

The second contribution of the paper is to introduce new  \emph{data and benchmarks} to study texture recognition in realistic settings. While most of the earlier work on texture recognition was carried out in carefully controlled conditions, more recent benchmarks such as the Flickr material dataset (FMD)~\cite{sharan09material}  have emphasized the importance of testing algorithms ``in the wild'',  for example on Internet images. However, even these datasets are somewhat removed from practical applications as they assume that textures fill the field of view, whereas in applications they are often observed in clutter. Here we leverage the excellent OpenSurfaces dataset~\cite{bell13opensurfaces} to create novel benchmarks for materials and texture attributes where textures appear both in the wild and in clutter (Sect.~\ref{s:clutter}), and demonstrate promising recognition results in these challenging conditions. In \cite{bell15material} the same authors have also investigated material recognition using OpenSurfaces.

The third contribution is technical and revisits classical ideas in texture modeling in the light of modern local feature descriptors and pooling encoders. While texture representations were extensively used in most areas of image understanding, since the breakthrough work of~\cite{krizhevsky12imagenet} they have been replaced by deep Convolutional Neural Networks (CNNs). Often CNNs are applied to a problem by using transfer learning, in the sense that the network is first trained on a large-scale image classification task such as the ImageNet ILSVRC challenge~\cite{deng09imagenet}, and then applied to another domain by exposing the output of a so-called ``fully connected layer'' as a general-purpose image representation. In this work we illustrate the many benefits of truncating these CNNs earlier, at the level of the convolutional layers (Sect.~\ref{s:tex-descr}). In this manner, one obtains  powerful \emph{local image descriptors} that, combined with traditional pooling encoders developed for texture representations, result in state-of-the-art recognition accuracy in a diverse set of visual domains, from material and texture attribute recognition, to coarse and fine grained object categorization and scene classification. We show that a benefit of this approach is that features transfer easily across domains even without fine-tuning the CNN on the target problem. Furthermore, pooling allows us to efficiently evaluate descriptors in image subregions, a fact that we exploit to recognize local image regions without recomputing CNN features from scratch.

A symmetric approach, using SIFT as local features and the IFV followed by fully-connected layers from a deep neural network as a pooling mechanism, was proposed in~\citep{perronnin2015fisher}, obtaining similar results on VOC07.

This paper is the archival version of two previous publications~\cite{cimpoi14describing} and~\cite{cimpoi15deep}. Compared to these two papers, this new version adds a significant number of new experiments and a substantial amount of new discussion.

The code and data for this paper are available on the project page, at \url{http://www.robots.ox.ac.uk/~vgg/research/deeptex}.

\section{Describing textures with attributes}\label{s:dtd}

\begin{figure}[ht]
\begin{center}
\includegraphics[width=\columnwidth, natwidth=414, natheight=292]{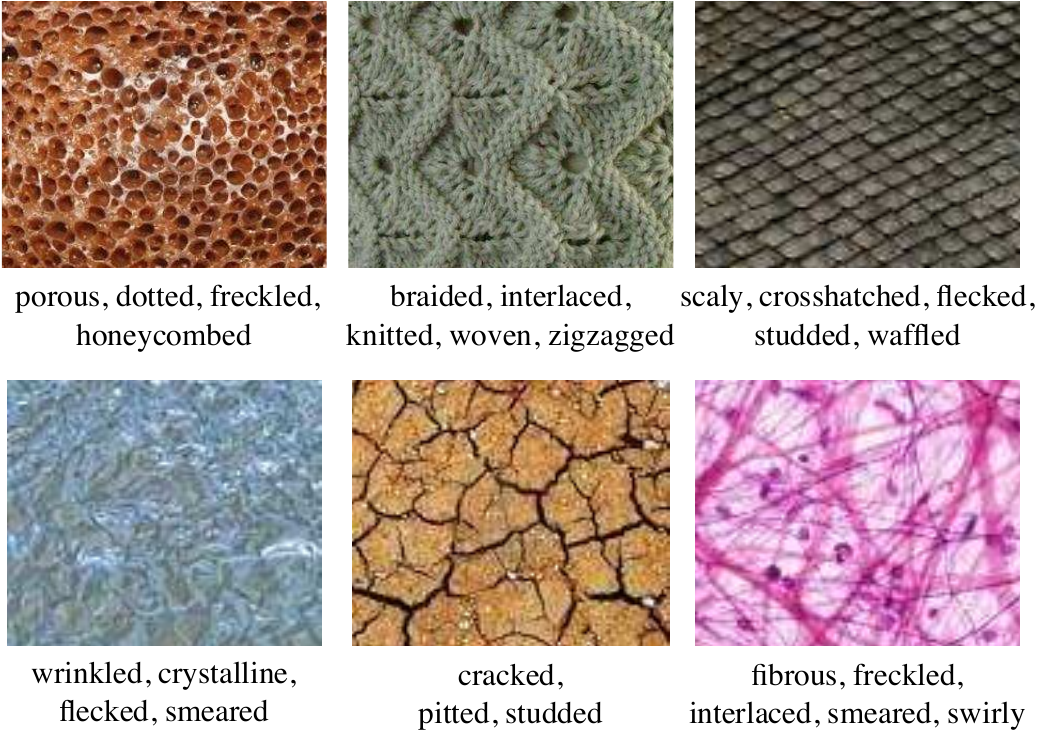}
\end{center}
\caption{We address the problem of {\em describing textures} by associating to them a collection of attributes. Our goal is to understand and generate automatically human-interpretable descriptions such as the examples above.} 
\label{fig:example-output}
\end{figure}

This section looks at the problem of automatically describing texture patterns using a general-purpose vocabulary of human-interpretable texture attributes, in a manner similar to how we can vividly characterize the textures shown in Fig.~\ref{fig:example-output}. The goal is to design algorithms capable of generating and understanding texture descriptions involving a {\em combination of describable attributes} for each texture. Visual attributes have been extensively used in {\em search}, to understand complex user queries, in {\em learning}, to port textual information back to the visual domain, and in image {\em description}, to produce richer accounts of the content of images. Textural properties are an important component of the semantics of images, particularly for objects that are best characterized by a pattern, such as a scarf or the wings of a butterfly~\cite{wang09Blearning}. Nevertheless, the attributes of visual textures have been investigated only tangentially so far. Our aim is to fill this gap.

\begin{figure*}[t]
\raggedright
\foreach \n in {banded, blotchy,braided,bubbly, bumpy,chequered, cobwebbed, cracked, crosshatched, crystalline, dotted, fibrous, flecked, freckled, frilly,gauzy,grid, grooved, honeycombed, interlaced, knitted, lacelike, lined, marbled, matted, meshed, paisley, perforated, pitted, pleated, polka-dotted, porous, potholed, scaly, smeared, spiralled, sprinkled, stained, stratified, striped, studded, swirly, veined, waffled, woven, wrinkled, zigzagged}
{\renewcommand{\baselinestretch}{0.7}\normalsize
\begin{overpic}[width=0.12\textwidth,trim=0 0 -4pt -1pt]{fig_dtd_\n_0002.jpg}
\put(0,0){\includegraphics[trim=0 0 128pt -1pt,clip,width=0.06\textwidth]{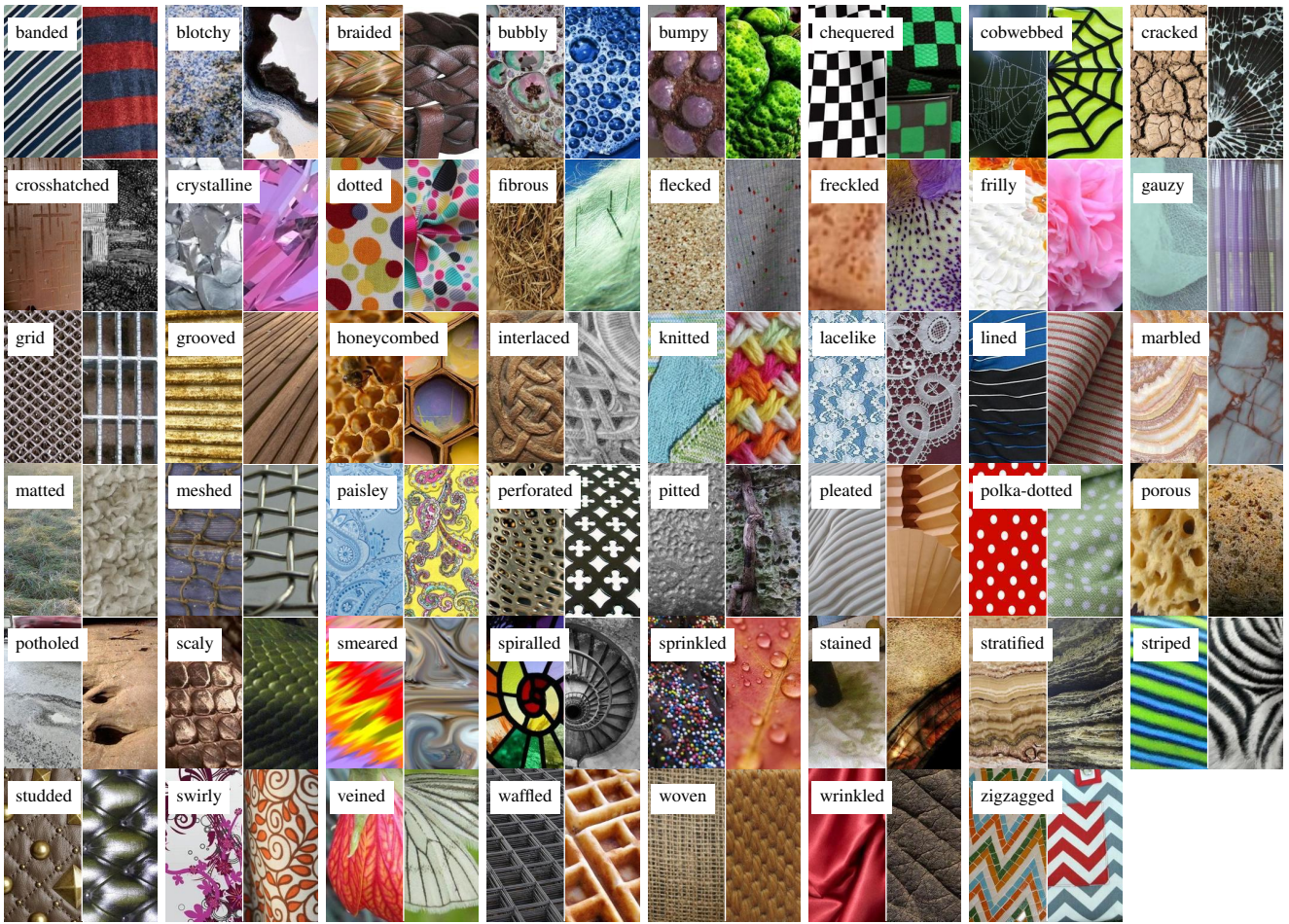}}
\put(50,0){\linethickness{0.5pt}\color{white}\line(0,1){100}}
\put(2,78){\scriptsize\colorbox{white}{\strut\n}}
\end{overpic} }
\caption{The 47 texture words in the \textbf{describable texture dataset} introduced in this paper. Two examples of each attribute are shown to illustrate the significant amount of variability in the data.}\label{f:words}
\end{figure*}

Our first contribution is to introduce the \textbf{Describable Textures Dataset} (DTD)~\cite{cimpoi14describing}, a collection of real-world texture images annotated with one or more adjectives selected in a vocabulary of forty-seven English words. These adjectives, or \emph{describable texture attributes}, are illustrated in Fig.~\ref{f:words} and include words such as {\em banded}, \emph{cobwebbed}, \emph{freckled}, \emph{knitted}, and \emph{zigzagged}. Sect.~\ref{s:dtd-data} describes this data in more detail. Sect.~\ref{s:dtd-collection} discusses the technical challenges we addressed while designing and collecting DTD, including how the forty-seven texture attributes were selected and how the problem of collecting numerous attributes for a vast number of images was addressed. Sect.~\ref{s:dtd-benchmark} defines a number of benchmark tasks in DTD. Finally, Sect.~\ref{s:dtd-related} relates DTD to existing texture datasets.

\subsection{The Describable Texture Dataset}\label{s:dtd-data}

DTD investigates the problem of \emph{texture description}, understood as the recognition of describable texture attributes. This problem is complementary to standard texture analysis tasks such as texture identification and material recognition for the following reasons. While describable attributes are correlated with materials, attributes do not imply materials (\eg \emph{veined} may equally apply to leaves or marble) and materials do not imply attributes (not all marbles are \emph{veined}). This distinction is further elaborated in Sect.~\ref{s:attribute-vs-material}.

\begin{figure*}[t]
\includegraphics[width=0.70\textwidth]{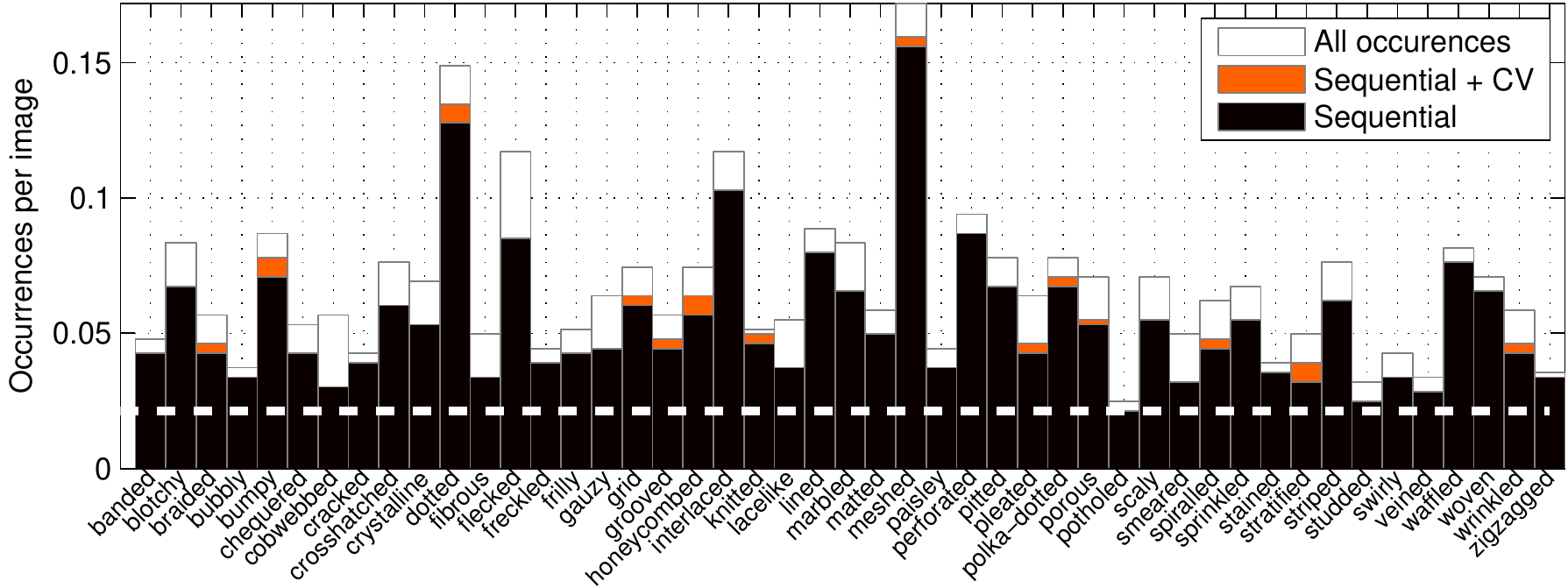}\hfill
\raisebox{0.95in}{
\setlength{\tabcolsep}{3pt}
\begin{tabular}{cc}
\raisebox{0.8in}{$q$} & {\setlength\fboxsep{0pt}\fbox{\includegraphics[width=0.2\textwidth]{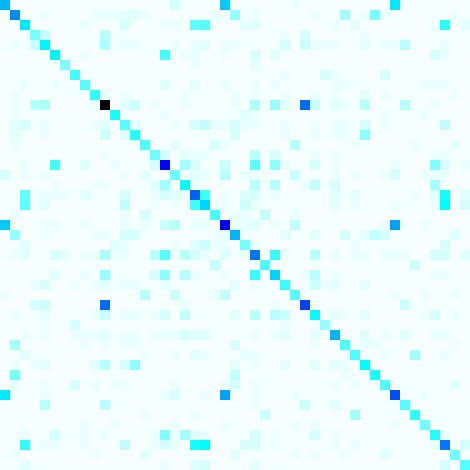}}}\\
& $q'$ \\
\end{tabular}
}\vspace{-1.2em}
\caption{{\bf Quality of sequential joint annotations.} Each bar shows the average number of occurrences of a given attribute in a DTD image. The horizontal dashed line corresponds to a frequency of 1/47, the minimum given the design of DTD (Sect.~\ref{s:dtd-collection}). The black portion of each bar is the amount of attributes discovered by the sequential procedure, using only 10 annotations per image (about one fifth of the effort required for exhaustive annotation). The orange portion shows the additional recall obtained by integrating cross-validation in the process. {\bf Right: co-occurrence of attributes.} The matrix shows the joint probability $p(q,q')$ of two attributes occurring together (rows and columns are sorted in the same way as the left image).}\label{f:ja}\label{f:co-occurence}
\end{figure*}

Describable attributes can be {\em combined} to create rich descriptions (Fig.~\ref{f:co-occurence}; marble can be \emph{veined}, \emph{stratified} and \emph{cracked} at the same time), whereas a typical assumption is that textures are made of a single material. Describable attributes are \emph{subjective} properties that depend on the imaged object as well as on human judgements, whereas materials are objective. In short, attributes capture properties of textures complementary to materials, supporting human-centric tasks where describing textures is important. At the same time, we will show that texture attributes are also helpful in material recognition (Sect.\ref{s:exp-dtd-attr}).

DTD contains {\bf textures in the wild}, \ie texture images extracted from the web rather than captured or generated in a controlled setting. Textures fill the entire image in order to allow studying the problem of texture description independently of texture segmentation, which is instead addressed in Sect.~\ref{s:clutter}. With 5,640 annotated texture images, this dataset aims at supporting real-world applications were the recognition of texture properties is a key component. Collecting images from the Internet is a common approach in categorization and object recognition, and was adopted in material recognition in FMD. This choice trades-off the systematic sampling of illumination and viewpoint variations existing in datasets such as CUReT, \kt, Outex, and Drexel to capture real-world variations, reducing the gap with applications. Furthermore, DTD captures empirically human judgements regarding the invariance of describable texture attributes; this invariance is not necessarily reflected in material properties.

\subsection{Dataset design and collection}\label{s:dtd-collection}

This section discusses how DTD was designed and collected, including: selecting the 47 attributes, finding at least 120 representative images for each attribute, and collecting all the attribute labels for each image in the dataset.

\subsubsection{Selecting the describable attributes}

 Psychological experiments suggest that, while there are a few hundred words that people commonly use to describe textures, this vocabulary is redundant and can be reduced to a much smaller number of representative words. Our starting point is the list of 98 words identified by Bhusan~\etal~\cite{bhushan1997texture}. Their seminal work aimed to achieve for texture recognition the same that color words have achieved for describing color spaces~\cite{berlin1991basic}. However, their work mainly focuses on the cognitive aspects of texture perception, including perceptual similarity and the identification of directions of perceptual texture variability. Since our interest is  in the visual aspects of texture, words such as ``corrugated" that are more related to surface shape or haptic properties were ignored. Other words such as ``messy" that are highly subjective and do not necessarily correspond to well defined visual features were also ignored. After this screening phase we analyzed the remaining words and merged similar ones such as ``coiled", ``spiraled"  and ``corkscrewed" into a single term. This resulted in a set of 47 words, illustrated in Fig.~\ref{f:words}.

\subsubsection{Bootstrapping the key images}

Given the 47 attributes, the next step consisted in collecting a sufficient number (120) of example images representative of each attribute. Initially, a large initial pool of about a hundred-thousand images in total was downloaded from Google and Flickr by entering the attributes and related terms as search queries. Then Amazon Mechanical Turk (AMT) was used to remove low resolution, poor quality, watermarked images, or images that were not almost entirely filled with a texture. Next, detailed annotation instructions were created for each of the 47 attributes, including a dictionary definition of each concept and examples of textures that did and did not match the concept. Votes from three AMT annotators were collected for the candidate images of each attribute and a shortlist of about 200 highly-voted images was further manually checked by the authors to eliminate remaining errors. The result was a selection of 120 {\em key representative images} for each attribute.

\subsubsection{Sequential joint annotation}\label{s:joint-anno}

So far only the key attribute of each image is known while any of the remaining 46 attributes may apply as well. Exhaustively collecting annotations for 46 attributes and 5,640 texture images is fairly expensive. To reduce this cost we propose to exploit the correlation and sparsity of the attribute occurrences (Fig.~\ref{f:co-occurence}). For each attribute $q$, twelve key images are annotated exhaustively and used to estimate the probability $p(q'|q)$ that {\em another} attribute $q'$ could co-exist with $q$. Then for the remaining key images of attribute $q$, only annotations for attributes $q'$ with non negligible probability are collected, assuming that the remaining attributes would not apply. In practice, this requires annotating around 10 attributes per texture instance, instead of 47. This procedure occasionally misses attribute annotations; Fig.~\ref{f:ja} evaluates attribute recall by 12-fold cross-validation on the 12 exhaustive annotations for a fixed budget of collecting 10 annotations per image.

A further refinement is to suggest which attributes $q'$ to annotate not just based on the prior $p(q'|q)$, but also based on the appearance of an image $\bx_i$. This was done by using the attribute classifier learned in Sect.~\ref{s:exp-results}; after Platt's calibration~\cite{platt00probabilistic} on a held-out test set, the classifier score $c_{q'}(\bx_i) \in \real$ is transformed in a probability $p(q'|\bx_i) = \sigma(c_{q'}(\bx))$ where  $\sigma(z) = 1 / (1 + e^{-z})$ is the sigmoid function. By construction, Platt's calibration reflects the prior probability $p(q') \approx p_0 = 1/47$ of $q'$ on the validation set. To reflect the probability $p(q'|q)$ instead, the score is adjusted as
\[
 p(q'|\ell_i, q) \propto
 \sigma(c_{q'}(\ell_i)) \times
 \frac{p(q'|q)} {1 -  p(q'|q)} \times
 \frac{1 - p_0}{p_0}
\]
and used to find which attributes should be annotated for each image. As shown in Fig.~\ref{f:ja}, for a fixed annotation budged this method increases attribute recall. 

Overall, with roughly 10 annotations per image it was possible to recover  all of the attributes for at least 75\% of the images, and miss one out of four (on average) for another 20\%, while keeping the annotation cost to a reasonable level. To put this in perspective,  directly annotating the 5,640 images for 46 attributes and collecting five annotations per attributed would have required 1.2M binary annotations, i.e. roughly 12K USD at the very low rate of 1\textcent{} per annotation. Using the proposed method,  the cost would have been 546 USD.  In practice, we spent around 2.5K USD in order to pay annotators better as well as to account for occasional errors in setting up experiments and the fact that, as explained above, bootstrapping still relies on exhaustive annotations for a subset of the data.

\subsection{Benchmark tasks}\label{s:dtd-benchmark}

DTD is designed as a public benchmark. The data, including images, annotations, and splits, is available on the web at \url{http://www.robots.ox.ac.uk/~vgg/data/dtd}, along with code for evaluation and reproducing the results in Sect.~\ref{s:exp-results}.

DTD defines two challenges. The first one, denoted DTD, is the \emph{prediction of key attributes}, where each image is assigned a single label corresponding to the key attribute defined above. The second one, denoted DTD-J, is the joint \emph{prediction of multiple attributes}. In this case each image is assigned one or more labels, corresponding to all the attributes that apply to that image.

The first task is evaluated both in term of classification accuracy (acc) and in term of mean average precision (mAP), while the second task only in term of mAP due to the possibility of multiple labels. The classification accuracy is normalized per class: if $\hat c(\bx), c(\bx)\in \{1,\dots, C\}$ are respectively the predicted and ground-truth label of image $\bx$, accuracy is defined as
\begin{equation}\label{e:acc}
\operatorname{acc}(\hat c)
=
\frac{1}{C}
\sum_{\bar c=1}^C \frac{|\{\bx:c(\bx) = \bar c \wedge \hat c(\bx)=\bar c\}|}{|\{\bx:c(\bx)=\bar c\}|}.
\end{equation}
We define mAP as per the PASCAL VOC 2008 benchmark onward~\cite{pascal08}.\footnote{PASCAL VOC 2007 uses instead an interpolated version of mAP.}

DTD contains 10 preset splits into equally-sized training, validation and test subsets for easier algorithm comparison. Results on any of the tasks are repeated for each split and average accuracies are reported.

\subsection{Attributes vs materials}\label{s:attribute-vs-material}

 
As noted at the beginning of Sect.~\ref{s:dtd-data} and in~\cite{sharan13recognizing},  texture attributes and materials are correlated, but not equivalent. In this section we verify this quantitatively  on the FMD data~\cite{sharan09material}. Specifically, we manually collected annotations for the 47 DTD attributes for the 1,000 images in the FMD dataset, which span ten different materials. Each of the 47 attributes was considered in turn, using a categorical random variable $C \in \{1,2,\dots,10\}$ to denote the texture material and a binary variable $A \in \{0,1\}$ to indicate whether the attribute applies to the texture or not. On average, the relative reduction in the entropy of the material variable $I(A,C) / H(C)$ given the attribute is of about 14\%; vice-versa, the  relative reduction in the entropy of the attribute variable $I(A,C)/H(A)$ given the material is just $0.5\%$. We conclude that knowing the material or attribute of a texture provides little information on the attribute or material, respectively. Note that  \emph{combinations} of attributes can predict materials much more reliably, although this is difficult to quantify from a small dataset.

\subsection{Related work}\label{s:dtd-related}

\begin{table*}[t!]
\centering
{\small \setlength{\tabcolsep}{3pt}
\begin{tabular}{|l|ccc|ccc|ccc|c|}
\hline
                & \multicolumn{3}{c|}{Size}       & \multicolumn{3}{c|}{Condition}        & \multicolumn{3}{c|}{Content}           & {(I)nstances /}\\
{Dataset}       & {Images} & {Classes}  & Splits  & { Wild} & {Clutter}  & {Controlled}   & {Attributes} & {Materials}& {Objects}  & {(C)ategories}  \\
\hline
{Brodatz}       & 999      &    111     &    --   &         &            &    X           &              &            &     X      &    I   \\
{CUReT}         & 5612     &     61     &    10   &         &            &    X           &              &     X      &            &    I   \\
{UIUC}          & 1000     &     25     &    10   &         &            &    X           &              &     X      &            &    I   \\
{UMD}           & 1000     &     25     &    10   &    X    &            &                &              &            &     X      &    I   \\
{KTH}           & 810      &     11     &    10   &         &            &    X           &              &     X      &            &    I   \\
{Outex}         & --       &     --     &    --   &         &            &    X           &              &     X      &     X      &    I   \\
{Drexel}        & $\sim$40000&   20     &    --   &         &            &    X           &              &     X      &            &    I   \\
{ALOT}          & 25000    &     250    &    10   &         &            &    X           &              &     X      &            &    I   \\
\hline
{FMD}           & 1000     &     10     &    14   &    X    &            &                &              &     X      &            &    C   \\
{\ktb}         & 4752     &     11     &         &         &            &    X           &              &     X      &            &    C   \\
\hline
{DTD}           & 5640     &     47     &    10   &    X    &            &                &     X        &            &            &    C   \\
{OS}            & 10422    &     22     &     1   &    X    &     X      &                &     X (+A)   &     X      &            &    C   \\
\hline
\end{tabular}
}\vspace{-1em}
\caption{Comparison of existing texture datasets, in terms of size, collection condition, nature of the classes to be recognized, and whether each class includes a single object/material instance or several instances of the same category. Note that Outex is a meta-collection of textures spanning different datasets and problems.}
\label{tbl:datasets}
\end{table*}

This section relates DTD to the literature in texture understanding. Textures, due to their ubiquitousness and complementarity to other visual properties such as shape, have been studied in several contexts: texture perception~\cite{adelson01on-seeing,amadasun89textural,gaarding1992shape,forsyth2001shape}, description~\cite{ferrari07learning}, material recognition~\cite{leung2001representing,sharan13recognizing,schwartz13visual,varma2005statistical,ojala2002multiresolution,varma2003texture,leung2001representing}, segmentation~\cite{manjunath1991unsupervised,jain1991unsupervised,jain1991unsupervised,chaudhuri1995texture,manjunath1991unsupervised,dunn1994texture}, synthesis~\cite{efros1999texture,wei2000fast,portilla2000parametric}, and shape from texture~\cite{gaarding1992shape,forsyth2001shape,malik1997computing}. Most related to DTD is the work on texture recognition, summarized below as the recognition of perceptual properties (Sect.~\ref{s:rec-percept}) and recognition of  identities and materials (Sect.~\ref{s:rec-materials})

\subsubsection{Recognition of perceptual properties}\label{s:rec-percept}

The study of perceptual properties of textures originated in computer vision as well as in cognitive sciences. Some of the earliest work on texture perception conducted by Julesz~\cite{julesz81textons} focussed on pre-attentive aspects of perception. It led to the concept of ``textons," primitives such as line-terminators, crossings, intersections, etc., that are responsible for pre-attentive discrimination of textures. In computer vision, Tamura et al.~\cite{tamura78textural} identified six common directions of variability of images in the Broadatz dataset; coarse vs. fine; high-contrast vs. low-contrast; directional vs. non-directional; linelike vs. bloblike; regular vs. irregular; and rough vs. smooth. Similar perceptual attributes of texture~\cite{amadasun89textural,bajcsy73computer} have been found by other researchers.

Our work is motivated by that of Bhusan~\etal~\cite{rao96towards,bhushan1997texture}. Their experiments suggest that there is a strong correlation between the structure of the lexical space and perceptual properties of texture. While they studied the psychological aspects of texture perception, the focus of this paper is the challenge of estimating such properties from images automatically. Their work~\cite{bhushan1997texture}, in particular, identified a set of words sufficient to describe a wide variety of texture patterns; the same set of words was used to bootstrap DTD.

While recent work in computer vision has been focussed on texture identification and material recognition, notable contributions to the recognition of perceptual properties exist. Most of this work is part of the general research on \emph{visual attributes}~\cite{farhadi2009describing,parikh11relative,patterson2012sun,bourdev2011describing,kumar2011describable}.  Texture attributes have an important role in describing objects, particularly for those that are best characterized by a pattern, such as items of clothing and parts of animals such as birds. Notably, the first work on modern visual attributes by Ferrari et al.~\cite{ferrari07learning} focused on the recognition of a few perceptual properties of textures. Later work, such as~\cite{berg2010automatic} that mined visual attributes from images on the Internet, also contain some attributes that describe textures. Nevertheless, so far the attributes of textures have been investigated only tangentially. DTD address the question of whether there exists a ``universal'' set of attributes that can describe a wide range of texture patterns, whether these can be reliably estimated from images, and for what tasks they are useful.

Datasets that focus on the recognition of subjective properties of textures are less common. One example is \emph{Pertex}~\cite{clarke11perceptual}, containing 300 texture images taken in a controlled setting (Lambertian renderings of 3D reconstructions of real materials) as well as a semantic similarity matrix obtained form human similarity judgments. The work most related to ours is probably the one of~\cite{matthews13enriching} that analyzed images in the Outex dataset~\citep{ojala2002multiresolution} using a subset of the texture attributes that we consider. DTD differs in scope (containing more attributes) and, especially, in the nature of the data (controlled vs uncontrolled conditions). In particular, working in uncontrolled conditions allows us to transfer the texture attributes to real-world applications, including material recognition in the wild and in clutter, as shown in the experiments.

\begin{figure}[t]
\begin{center}
\includegraphics[width=1\linewidth]{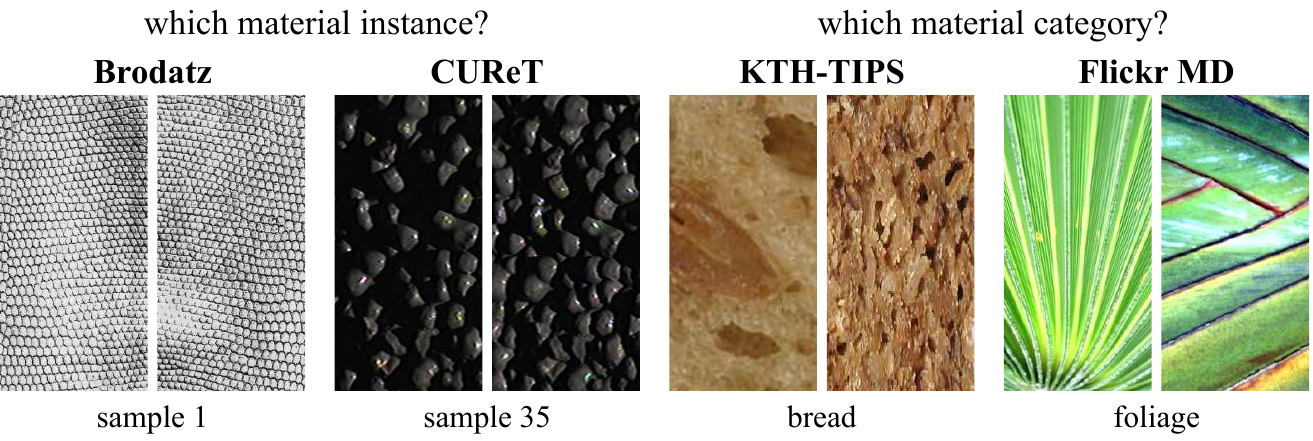}
\end{center}\vspace{-2em}
\caption{\label{fig:fmd} Datasets such as Brodatz~\cite{brodatz66textures} and CUReT~\cite{dana99reflectance} (left) addressed the problem of material instance identification and others such as. \kthb~\cite{hayman04learning} and FMD~\cite{sharan09material} (right) addressed the problem of material category recognition. Our DTD dataset addresses a very different problem: the one of describing a pattern using intuitive attributes (Fig.~\ref{fig:example-output}).}\label{f:datasets}
\end{figure}

\subsubsection{Recognition of texture instances and material categories}\label{s:rec-materials}

Most of the recent work in texture recognition focuses on the recognition of texture instances and material categories, as reflected by the development of corresponding benchmarks (Fig.~\ref{f:datasets}). The \emph{Brodatz}~\cite{brodatz66textures} catalogue was used in early works on textures to study the problem of identifying texture instances (e.g. matching half of the texture image given the other half). Others including \emph{CUReT}~\cite{dana99reflectance}, \emph{UIUC}~\cite{lazebnik05}, \kt~\cite{caputo05class,hayman04learning}, \emph{Outex}~\citep{ojala2002multiresolution}, \emph{Drexel Texture Database}~\cite{oxholm2012texture}, and \emph{ALOT}~ \cite{burghouts09} address the recognition of {\em specific instances of one or more materials}. \emph{UMD}~\cite{xu09viewpoint} is similar, but the imaged objects are not necessarily composed of a single material. As textures are imaged under variable truncation, viewpoint, and illumination, these datasets have stimulated the creation of texture representations that are invariant to viewpoint and illumination changes~\cite{varma2005statistical,ojala2002multiresolution,varma2003texture,leung2001representing}. Frequently, texture understanding is formulated as the problem of recognizing the material of an object rather than a particular texture instance (in this case any two slabs of marble would be considered equal). \kthb~\cite{mallikarjuna05} is one of the first datasets to address this problem by grouping textures not only by the instance, but also by the type of materials (e.g. ``wood''). 

However, these datasets make the simplifying assumption that textures fill images, and often, there is limited intra-class variability, due to \emph{a single or limited number of instances}, captured under controlled scale, view-angle and illumination. Thus, they are not representative of the problem of recognizing materials in natural images, where textures appear under poor viewing conditions, low resolution, and in clutter. Addressing this limitation is the main goal of the \emph{Flickr Material Database} (FMD)~\cite{sharan09material}. FMD samples just one viewpoint and illumination per object, but contains many different object instances grouped in several different material classes. Sect.~\ref{s:clutter} will introduce datasets addressing the problem of clutter as well.

The performance of recognition algorithms on most of this data is close to perfect, with classification accuracies well above 95\%; \kthb and FMD are an exception due to their increased complexity. A review of these datasets and classification methodologies is presented in~\cite{timofte12trainingfree}, who also propose a training-free framework to classify textures, significantly improving on other methods. Table~\ref{tbl:datasets} and Fig.~\ref{f:datasets} provides a summary of the nature and size of various texture datasets that are used in our experiments.

\section{Recognizing textures in clutter}\label{s:clutter}\label{ss:os}

This section looks at the second contribution of the paper, namely studying the recognition of materials and describable textures attributes not only ``in the wild,'' but also ``in clutter''. Even in datasets such as FMD and DTD, in fact, each texture instance fills the entire image, which doest not match most applications. This section removes this limitation and looks at the problem of recognizing textures imaged in the larger context of a complex natural scene, including the challenging task of automatically segmenting textured image regions.


Rather than collecting a new image dataset from scratch,  our starting point is the excellent \emph{OpenSurfaces} (OS) dataset that was recently introduced by Bell~\etal~\cite{bell13opensurfaces}. OS comprises 25,357 images, each containing a number of high-quality texture/material segments. Many of these segments are annotated with additional attributes such as the material, viewpoint, BRDF estimates, and object class. Experiments focus on the 58,928 segments that contain material annotations. Since material classes are highly unbalanced, we  consider only the materials that contain at least 400 examples. This results in 53,915 annotated material segments in 10,422 images spanning 23 different classes.\footnote{The classes and corresponding number of example segments are:
          brick (610),
      cardboard  (423),
     carpet/rug (1,975),
        ceramic (1,643),
       concrete  (567),
   fabric/cloth (7,484),
           food (1,461),
          glass (4,571),
 granite/marble (1,596),
           hair  (443),
          other (2,035),
       laminate  (510),
        leather  (957),
          metal (4,941),
        painted (7,870),
   paper/tissue (1,226),
  plastic/clear  (586),
 plastic/opaque (1,800),
          stone  (417),
           tile (3,085),
      wallpaper  (483),
           wood (9,232).}
Images are split evenly into training, validation, and test subsets with 3,474 images each. Segment sizes are highly variable, with half of them being relatively small, with an area smaller than $64 \times 64$ pixels. One issue with crowdsourced collection of segmentations is that not all the pixels in an image are labelled. This makes it difficult to define a complete background class. For our benchmark several less common materials (including for example segments that annotators could not assign to a material) were merged in an ``other'' class that acts as the background.

This benchmark is similar to the one concurrently proposed by Bell \emph{et al.}~\cite{bell15material}. However, in order to study perceptual properties as well as materials, we also augment the OS dataset with some of the describable attributes of Sect.~\ref{s:dtd}. Since the OS segments do not trigger with sufficient frequency all the 47 attributes, the evaluation is restricted to eleven of them for which it was possible to identify at least 100 matching segments.\footnote{These are: banded, blotchy, checkered, flecked, gauzy, grid, marbled, paisley, pleated, stratified, wrinkled.} The attributes were manually labelled in the 53,915 segments retained for materials. We refer to this data as OSA.

\subsection{Benchmark tasks}\label{s:os-benchmark}

As for DTD, the aim is to define standardized image understanding tasks to be used as public benchmarks. The complete list of images, segments, labels, and splits are publicly available at \url{http://www.robots.ox.ac.uk/~vgg/data/wildtex/}. 

The benchmarks include two tasks on two complementary semantic domains. The first task is the \emph{recognition of texture regions, given the region extent as ground truth information}. This task is instantiated for both material, denoted OS+R, and describable texture attributes, denoted OSA+R. Performance in  OS+R is measured in term of classification accuracy and mAP, using the same definition~\eqref{e:acc} where images are replaced by image regions. Performance in OSA+R uses instead mAP due to the possibility of multiple labels.

The second task is the \emph{segmentation and recognition of texture regions}, which we also instantiate for materials (OS) and describable texture attributes (OSA).  Since not all image pixels are labelled in the ground truth, the performance of a predictor $\hat c$ is measured in term of per-pixel classification accuracy, $\operatorname{pp-acc}(\hat c)$. This is computed using the same formula as \eqref{e:acc} with two modification: first, the images $\bx$ are replaced by pixels $\bp$ (extracted from all images in the dataset); second, the ground truth label $c(\bp)$ of a pixel may take an additional value 0 to denote pixels that are not labelled in the ground truth (the effect is to ignore them in the computation of accuracy). 

In the case of OSA, the per-pixel accuracy is modified such that a class prediction is considered correct if it belongs to any of the ground-truth pixel labels. Furthermore, accuracy is not normalized per class as this is ill-defined, but by the total number of pixels: 
\begin{equation}\label{e:acc-osa}
\operatorname{acc-osa}(\hat c)
=
\frac{|\{ \bp : \hat c(\bp) \in c(\bp) \}|}{|\{ \bp : c(\bp) \not= \phi \}|}.
\end{equation}
where $c(\bp)$ is the set of possible labels of pixel $\bp$ and $\phi$ denotes the empty set.

\section{Texture representations}\label{s:tex-descr}
Having presented our contributions to framing the problem of texture description, we now turn to our technical advances towards addressing the resulting problems. We start by revisiting the concept of texture representation and studies how it relates to modern image descriptors based on CNNs. In general,  a visual representation is a map that takes an image $\bx$ to a vector $\phi(\bx)\in\real^d$ that facilitates understanding the image content. Understanding is often achieved by learning a linear predictor $\langle \phi(\bx), \bw \rangle$ scoring the strength of association between the image and a particular concept, such as an object category.

Among image representations,  this paper is particularly interested in the class of \emph{texture representations} pioneered by the works of~\cite{Mallat89,malik90preattentive,BovikCG90,leung2001representing}. Textures encompass a large diversity of visual patterns, from regular repetitions such as wallpapers, to stochastic processes such as fur, to intermediate cases such as pebbles. Distortions due to viewpoint and other imaging factors further complicate modeling textures. However, one can usually assume that, given a particular texture, appearance variations are statistically independent in the long range and can therefore be eliminated by averaging local image statistics over a sufficiently large texture sample. Hence, the defining characteristic of texture representations is to pool information extracted locally and uniformly from the image, by means of local descriptors, in an orderless manner. 

The importance of texture representations is in the fact that they were found to be applicable well beyond textures. For example, until recently many of the best object categorization methods in challenges such as PASCAL VOC~\cite{everingham07pascal} and ImageNet ILSVRC~\cite{deng09imagenet} were based on variants of texture representations, developed specifically for objects. One of the contributions of this work is to show that these object-optimized texture representations are in fact optimal for a large number of texture-specific problems too (Sect.~\ref{s:results-local}).

More recently, texture representations have been significantly outperformed by \emph{Convolutional Neural Networks} (CNNs) in object categorization~\cite{krizhevsky12imagenet}, detection~\cite{girshick14rich}, segmentation~\cite{hariharan2014simultaneous}, and in fact in almost all domains of image understanding. Key to the success of CNNs is their ability to leverage large \emph{labelled} datasets to learn high-quality features. Importantly, CNN features pre-trained on very large datasets were found to transfer to many other domains with a relatively modest adaptation effort~\cite{jia13caffe,oquab14learning,razavin14cnn-features,chatfield14return,girshick14rich}. Hence, CNNs provide general-purpose image descriptors.

While CNNs generally outperform classical texture representations, it is interesting to ask what is the relation between these two methods and whether they can be fruitfully hybridized.  Standard CNN-based methods such as~\cite{jia13caffe,oquab14learning,razavin14cnn-features,chatfield14return,girshick14rich} can be interpreted as extracting local image descriptors (performed by the the so called ``convolutional layers'') followed by pooling such features in a global image representation (performed by the ``Fully-Connected (FC) layers''). Here we will show that replacing FC pooling with one of the many pooling mechanisms developed in texture representations has several advantages: (i) a much faster computation of the representation for image subregions accelerating applications such as detection and segmentation~\cite{girshick14rich,he14spatial,gong14multi-scale}, (ii) a significantly superior recognition accuracy in several application domains and (iii) the ability of achieving this superior performance without fine-tuning CNNs by implicitly reducing the domain shift problem.

In order to systematically study variants of texture representations $\phi=\phi_e \circ \phi_f$ , we break them into local descriptor extraction $\phi_f$ followed by descriptor pooling $\phi_e$. In this manner, different combinations of each component can be evaluated. Common local descriptors include linear filters, local image patches, local binary patterns, densely-extracted SIFT features, and many others. Since local descriptors are extracted uniformly from the image,  they can be seen as banks of (non-linear) filters; we therefore refer to them as \emph{filter banks} in honor of the pioneering works of~\cite{Mallat89,BovikCG90,FreemanA91,leung2001representing} and others where descriptors were the output of actual linear filters. Pooling methods include bag-of-visual-words, variants using soft-assignment, or extracting higher-order statistics as in the Fisher vector. Since these methods encode the information contained in the local descriptors in a single vector, we refer to them as \emph{pooling encoders}. 

Sect.~\ref{s:texture-local-descr} and Sect.~\ref{s:global-features} discuss filter banks and pooling encoders in detail.

\subsection{Local image descriptors}\label{s:texture-local-descr}

There is a vast choice of local image descriptors in texture representations. Traditionally, these features were handcrafted , but with the latest generation of deep learning methods it is now customary to learn them from data (although often in an implicit form). Representative examples of these two families of local features are discussed in Sect.~\ref{s:handcrafted} and Sect.~\ref{s:learned}, respectively.

\subsubsection{Hand-crafted local descriptors}\label{s:handcrafted}

Some of the earliest local image descriptors were developed as linear filter banks in texture recognition. As an evolution of earlier texture filters~\cite{BovikCG90,malik90preattentive}, the filter bank of \textbf{Leung Malik}~(LM)~\cite{leung01representing} includes 48 filters matching bars, edges and spots, at various scales and orientations. These filters are first and second derivatives of Gaussians at 6 orientations and 3 scales (36), 8 Laplacian of Gaussian (LOG) filters, and 4 Gaussians. Combinations of the filter responses, identified by vector quantisation (Sect.~\ref{s:orderless-pooling}), were used as the computational basis of the ``textons" proposed by Julesz~\cite{julesz83textons}. The filter bank \textbf{MR8} of~\cite{varma2003texture,geusebroek03fast} consists instead of 38 filters, similar to LM. For two of the oriented filters, only the maximum response across the scales is recorded, reducing the number of responses to 8 (3 scales for two oriented filters, and two isotropic -- Gaussian and Laplacian of Gaussian). 

The importance of using linear filters as local features was later questioned by Varma and Zisserman~\cite{varma2003texture}. The~\textbf{VZ} descriptors are in fact small image patches which, remarkably, were shown to outperform LM and MR8 on earlier texture benchmarks such as CuRET. However, as will be demonstrated in the experiments, trivial local descriptors are not competitive in harder tasks.

Another early local image descriptor are the \textbf{Local Binary Patterns} (LBP) of~\cite{ojala96a-comparative,ojala2002multiresolution}, a special case of the texture units of~\cite{wang90texture}. A LBP $\bd_i = (b_1,\dots,b_m)$ computed a pixel $\bp_0$ is the sequence of bits $b_j = [\bx(\bp_i) > \bx(\bp_j)]$ comparing the intensity $\bx(\bp_i)$ of the central pixel to the one of $m$ neighbors $\bp_j$ (usually 8 in a circle). LBPs have specialized quantization schemes; the most common one maps the bit string $\bd_i$ to one of a number of {\em uniform patterns}~\cite{ojala2002multiresolution}. The quantized LBPs can be averaged over the image to build a histogram; alternatively, such histograms can be computed for small image patches and used in turn as local image descriptors.

In the context of object recognition, the best known local descriptor is undoubtedly D. Lowe's \textbf{SIFT}~\cite{lowe99object}. SIFT is the histogram of the occurrences of image gradients quantized with respect to their location within a patch as well to their orientation. While SIFT was originally introduced to match object instances, it was later applied to an impressive diversity of tasks, from object categorization to semantic segmentation and face recognition.

\subsubsection{Learned local descriptors}\label{s:dcnn}\label{s:learned}

Handcrafted image descriptors are nowadays outperformed by features learned using the latest generation of deep CNNs~\cite{krizhevsky12imagenet}. A CNN can be seen as a composition $\phi_K \circ \dots \circ \phi_2 \circ \phi_1$ of $K$ functions or \emph{layers}. The output of each layer $\bx_k = (\phi_k \circ \dots \circ \phi_2 \circ \phi_1)(\bx)$ is a \emph{descriptor field} $\bx_k \in \mathbb{R}^{W_k \times H_k \times D_k}$, where $W_k$ and $H_k$ are the width and height of the field and $D_k$ is the number of feature channels. By collecting the $D_k$ responses at a certain spatial location, one obtains a $D_k$ dimensional descriptor vector. The network is called convolutional if all the layers are implemented as (non-linear) filters, in the sense that they act locally and uniformly on their input. If this is the case, since compositions of filters are filters,  the feature field $\bx_k$ is the result of applying a non-linear filter bank to the image $\bx$.

As computation progresses, the resolution of the descriptor fields decreases whereas the number of feature channels increases. Often, the last several layers $\phi_k$ of a CNN are called ``fully connected'' because, if seen as filters, their support is the same as the size of the input field $\bx_{k-1}$ and therefore lack locality. By contrast, earlier layers that act locally will be referred to as ``convolutional''. If there are $C$ convolutional layers, the CNN $\phi=\phi_e \circ \phi_f$ can be decomposed into a filter bank (local descriptors) $\phi_f = \phi_C \circ \dots \circ \phi_1$ followed by a pooling encoder $\phi_e = \phi_K \circ \dots \circ \phi_{C+1}$.

\subsection{Pooling encoders}\label{s:global-features}\label{s:descr-ifv}

A pooling encoder takes as input the local descriptors extracted from an image $\bx$ and produces as output a single feature vector $\phi(\bx)$, suitable for tasks such as classification with an SVM. A first important differentiating factor between encoders is whether they discard the spatial configuration of input features (orderless pooling; Sect.~\ref{s:orderless-pooling}) or whether they retain it (order-sensitive pooling; Sect.~\ref{s:rcnn}).  A detail of practical importance, furthermore, is the type of post-processing applied to the pooled vectors (post-processing; Sect.~\ref{s:post-processing}).

\subsubsection{Orderless pooling encoders}\label{s:orderless-pooling}

An \emph{orderless pooling encoder} $\phi_e$ maps a sequence $\mathcal{F} = (\bff_1,\dots,\bff_n), \bff_i\in\mathbb{R}^D$ of local image descriptors to a feature vector $\phi_e(\mathcal{F}) \in \mathbb{R}^d$. The encoder is orderless in the sense that the function $\phi_e$ is invariant to permutations of the input $\mathcal{F}$.\footnote{Note that $\mathcal{F}$ cannot be represented as a set as encoders are generally sensitive to repetitions of feature descriptors. It could be defined as a multiset or, as done here, as a sequence $\mathcal{F}$.} Furthermore, the encoder can be applied to any number of features; for example, the encoder can be applied to the sub-sequence $\mathcal{F}' \subset \mathcal{F}$ of local descriptors contained in a target image region without recomputing the local descriptors themselves.

All common orderless encoders are obtained by applying a non-linear \emph{descriptor encoder} $\eta(\bff_i)\in\mathbb{R}^d$ to individual local descriptors and then aggregating the result by using a commutative operator such as average or max. For example, average-pooling yields $\bar \phi_e(\mathcal{F}) = \frac{1}{n}\sum_{i=1}^n \eta(\bff_i)$. The pooled vector $\bar \phi_e(\mathcal{F})$ is post-processed to obtain the final representation $\phi_e(\mathcal{F})$ as discussed later.

The best-known orderless encoder is the \textbf{Bag of Visual Words} (BoVW). This encoder starts by vector-quantizing (VQ) the local features $\bff_i\in\mathbb{R}^D$ by assigning them to their  closest \emph{visual word} in a dictionary $C = \begin{bmatrix} \bc_1 & \dots &\bc_d\end{bmatrix} \in \mathbb{R}^{D \times d}$ of $d$ elements. Visual words can be thought of as ``prototype features'' and are obtained during training by clustering example local features. The descriptor encoder $\eta_1(\bff_i)$ is the one-hot vector indicating the visual word corresponding to $\bff_i$ and average-pooling these one-hot vectors yields the histogram of visual words occurrences.  BoVW was introduced in the work of~\cite{leung01representing} to characterize the distribution of textons, defined as configuration of local filter responses, and then ported to object instance and category understanding by~\cite{Sivic03} and \cite{csurka04visual} respectively. It was then extended in several ways as described below.

The kernel codebook encoder~\cite{Philbin08} assigns each local feature to several visual words, weighted by a degree of membership: $[\eta_{\text{KC}}(\bff_i)]_j \propto \exp\left( - \lambda \|\bff_i-\bc_j \|^2 \right)$, where $\lambda$ is a parameter controlling the locality of the assignment. The descriptor code $\eta_{\text{KC}}(\bff_i)$ is $L^1$ normalized before aggregation, such that $\|\eta_{\text{KC}}(\bff_i)\|_1 = 1$. Several related methods used concepts from \emph{sparse coding} to define the local descriptor encoder~\cite{zhou2010image,liu2011defense}. \textbf{Locality constrained Linear Coding} (LLC) \cite{wang2010locality}, in particular, extends soft assignment by making the assignments reconstructive, local, and sparse: the descriptor encoder $\eta_\text{LLC}(\bff_i) \in \real^d_+$, $\|\eta_\text{LLC}(\bff_i)\|_1=1$, $\|\eta_\text{LLC}(\bff_i)\|_0\leq r$ is computed such that $\bff_i \approx C\eta_\text{LLC} (\bff_i)$ while allowing only the $r \ll d$ visual words closer to $\bff_i$ to have a non-zero coeffcient. 

In the \textbf{Vector of Locally-Aggregated Descriptors} (VLAD) \cite{jegou10aggregating} the descriptor encoder is richer. Local image descriptors are first assigned to their nearest neighbor visual word in a dictionary of $K$ elements like in BoVW; then the descriptor encoder is given by $\eta_{\text{VLAD}}(\bff_i) = (\bff_i - C \eta_1(\bff_i)) \otimes \eta_1(\bff_i)$, where $\otimes$ is the Kronecker product. Intuitively, this subtracts from $\bff_i$ the corresponding visual word $C \eta_1(\bff_i)$ and then copies the difference into one of $K$ possible subvectors, one for each visual word. Hence average-pooling $\eta_{\text{VLAD}}(\bff_i)$ accumulates first-order descriptor statistics instead of simple occurrences as in BoVW.

VLAD can be seen as a variant of the \textbf{Fisher Vector} (FV)~\cite{perronnin07fisher}. The FV differs from VLAD as follows. First, the quantizer is not $K$-means but a Gaussian Mixture Model (GMM) with components $(\pi_k,\mu_k,\Sigma_k),$ $k=1,\dots,K$, where $\pi_k \in \real$ is the prior probability of the component, $\mu_k\in\real^D$ the Gaussian mean and $\Sigma_k \in\real^{D\times D}$ the Gaussian covariance (assumed diagonal). Second, hard-assignments $\eta_1(\bff_i)$ are replaced by soft-assignments $\eta_{\text{GMM}}(\bff_i)$ given by the posterior probability of each GMM component. Third, the FV descriptor encoder $\eta_{\text{FV}}(\bff_i)$ includes both first $\Sigma_k^{-\frac{1}{2}}(\bff_i - \mu_k)$ and second order $\Sigma_k^{-1} (\bff_i - \mu_k) \odot (\bff_i - \mu_k) - \mathbf{1}$ statistics, weighted by $\eta_\text{GMM}(\bff_i)$ (see~\cite{perronnin07fisher,perronnin10improving,chatfield11devil} for details). Hence, average pooling $\eta_{\text{FV}}(\bff_i)$ accumulates both first and second order statistics of the local image descriptors.

All the encoders discussed above use average pooling, except LLC that uses max pooling.

\subsubsection{Order-sensitive pooling encoders}\label{s:rcnn}

An \emph{order-sensitive encoder} differs from an orderless encoder in that the map $\phi_e(\mathcal{F})$ is not invariant to permutation of the input $\mathcal{F}$. Such an encoder can therefore reflect the layout of the local image desctiptors, which may be ineffective or even counter-productive in texture recognition, but is usually helpful in the recognition of objects, scenes, and others.

The most common order-sensitive encoder method is the \textbf{Spatial Pyramid Pooling} (SPP) of~\cite{Lazebnik06}. SSP transforms any orderless encoder into one with (weak) spatial sensitivity by dividing the image in subregions, computing any encoder for each subregion, and stacking the results. This encoder is only be sensitive to reassignments of the local descriptors to different subregions.

The \textbf{Fully-Connected layers} (FC) in a CNN also form an order-sensitive encoder. Compared to the encoders seen above, FC are pre-trained discriminatively, which can be either an advantage or disadvantage, depending on whether the information that they captured can be transferred to the domain of interest.  FC poolers are much less flexible than the encoders seen above as they work only with a particular type of local descriptors, namely the corresponding CNN convolutional layers. Furthermore, a standard FC pooler can only operate on a well defined layout of local descriptors (e.g. a $6 \times 6$), which in turn means that the image needs to be resized to a standard size before the FC encoder can be evaluated. This is particularly expensive when, as in object detection or image segmentation, many image subregions must be considered.

\subsubsection{Post-processing}\label{s:post-processing}

The vector $\by = \bar\phi_e(\mathcal{F})$ obtained by pooling local image descriptors is usually post-processed before being used in a classifier. In the simplest case, this amounts to performing $L^2$ normalization $\phi_e(\mathcal{F}) = \by / \|\by\|_2$. However, this is usually preceded by a non-linear transformation $\phi_K(\by)$ which is best understood in term of kernels. A \emph{kernel} $K(\by',\by'')$ specifies a notion of \emph{similarity} between data points $\by'$ and $\by''$. If $K$ is a positive semidefinite function, then it can always be rewritten as the inner product $\langle \phi_K(\by'),\phi_K(\by'')\rangle$ where $\phi_K$ is a suitable pre-processing function called a \emph{kernel embedding}~\cite{maji08classification,vedaldi10efficient}. Typical kernels include the linear, Hellinger's, additive-$\chi^2$,  and exponential-$\chi^2$ ones, given respectively by:
\begin{align*}
\mbox{}&\langle \by', \by''\rangle, &
\mbox{}&\sum_{i=1}^d \sqrt{y_i' y_i''},
\\
\mbox{}&\sum_{i=1}^d \frac{2 y_i' y_i''}{y_i'+y_i''}, &
\mbox{}&\exp\left[ - \lambda \sum_{i=1}^d \frac{(y_i' -y_i'')^2}{y_i'+y_i''} \right].
\end{align*}
In practice, the kernel embedding $\phi_K$ can be computed easily only in a few cases, including the linear kernel ($\phi_K$ is the identity) and Hellinger's kernel (for each scalar component, $\phi_\text{Hell.}(y) = \sqrt{y}$). In the latter case, if $y$ can take negative values, then the embedding is extended to the so called \emph{signed square rooting}\footnote{This extension generalizes to all homogeneous kernels, including for example $\chi^2$~\cite{vedaldi10efficient}.} $\phi_\text{Hell.}(y) = \operatorname{sign} (y) \sqrt{|y|}$.

Even if $\phi_K$ is not explicitly computed, any kernel can be used to learn a classifier such as an SVM (kernel trick). In this case, $L^2$ normalizing the kernel embedding $\phi_K(\by)$ amounts to normalizing the kernel as
\[
K'(\by,\by'') = \frac{K(\by',\by'')}{\sqrt{K(\by',\by') K(\by'',\by'')}}.
\]

All the pooling encoders discussed above are usually followed by post-processing. In particular, the \emph{Improved Fisher Vector} (IFV)~\cite{perronnin10improving} prescribes the use of the signed-square root embedding followed by $L^2$ normalization. VLAD has several standard variants that differ in the post-processing; here we use the one that $L^2$ normalizes the individual VLAD subvectors (one for each visual word) before $L^2$ normalizing the whole vector~\cite{arandjelovic12three}.

\section{Plan of experiments and highlights}\label{e:summary}

The next several pages contain an extensive set of experimental results. This section provides a guide to these experiments and summarizes the main findings.

The goal of the first block of experiments (Sect.~\ref{s:exp-depth}) is to determine which representations work bests on different problems such as texture attribute, texture material, object, and scene recognition. The main findings are:

\begin{itemize}
\item Orderless pooling of SIFT features (\eg FV-SIFT) performs better than specialized texture descriptors in many texture recognition problems; performance is further improved by switching from SIFT to CNN local descriptors (FV-CNN; Sect.~\ref{s:results-local}). 
\item Orderless pooling of CNN descriptors using the Fisher Vector (FV-CNN) is often significantly superior than fully-connected pooling of the same descriptors (FC-CNN) in texture, scene, and object recognition (Sect.~\ref{s:exp-global-descriptors}). This difference is more marked for deeper CNN architectures (Sect.~\ref{s:exp-deep-feat-variants}) and can be partially explained by the ability of FV pooling to overfit less and to easily integrate information at multiple image scales (Sect.~\ref{s:fv-vs-fc}).
\item FV-CNN descriptors can be compressed to the same dimensionality of FC-CNN descriptors while preserving accuracy (Sect.~\ref{s:pca}).
\end{itemize}

Having determined good representations in Sect.~\ref{s:exp-depth}, the second block of experiments (Sect.~\ref{s:exp-breadth}) compares them to the state of the art in texture, object, and scene recognition. The main findings are:

\begin{itemize}
\item In texture recognition in the wild, for both materials (FMD) and attributes (DTD), CNN-based descriptors substantially outperform existing methods. Depending on the dataset, FV pooling is a little or substantially better than FC pooling of CNN descriptors (Sect.~\ref{s:exp-texture-wild}). When textures are extracted from a larger cluttered scene (instead of filling the whole image), the difference between FV and FC pooling increases (Sect.~\ref{s:exp-texture-clutter}).
\item In coarse object recognition (PASCAL VOC), fine-grained object recognition (CUB-200), scene recognition (MIT Indoor),  and recognition of things \& stuff (MSRC) fine-grained, the FV-CNN representation achieves results that are close and sometimes superior to the state of the art, while using a simple and fully generic pipeline (Sect.~\ref{s:exp-places}).
\item FV-CNN appears to be particularly effective in domain transfer. Sect.~\ref{s:exp-places} shows in fact that FV pooling compensates for the domain gap caused by training a CNN on two very different domains, namely scene and object recognition.
\end{itemize}

Having addressed image classification in Sect.~\ref{s:exp-depth}  and~\ref{s:exp-breadth}, The third block of experiments (Sect.~\ref{s:tex-seg}) compare representations on semantic segmentation. It shows that FV pooling of CNN descriptors can be combined with a region proposal generator to obtain high-quality segmentation of materials in the OpenSurfaces and MSRC data. For example, combined with a post-processing step using a CRF, FV-VGG-VD surpasses the state-of-the-art on the latter dataset. It is also shown that, differently from FV-CNN, FC-CNN is too slow to be practical in this scenario.

\section{Experiments on semantic recognition}\label{s:exp-results}

So far the paper has introduced novel problems in texture understanding as well as a number of old and new texture representations. The goal of this section is to determine, through extensive experiments, what representations work best for which problem.

Representations are labelled as pairs X-Y, where X is a pooling encoder and Y a local descriptor. For example, FV-SIFT denotes the Fisher vector encoder applied to densely extracted SIFT descriptors, whereas BoVW-CNN denotes the bag-of-visual-words encoder applied on top of CNN convolutional descriptors. Note in particular that the CNN-based image representations as commonly extracted in the literature~\cite{jia13caffe,razavin14cnn-features,chatfield14return} implicitly use CNN-based descriptors and the FC pooler,  and therefore are denoted here as \rcnn.


\subsection{Local image descriptors and encoders evaluation}\label{s:exp-depth}

This section compares different local image descriptors and pooling encoders (Sect.~\ref{s:rec-descriptors}) on selected representative tasks in texture recognition, object recognition, and scene recognition (Sect.~\ref{s:rec-datsets}). In particular, Sect.~\ref{s:results-local} compares different local descriptors, Sect.~\ref{s:exp-global-descriptors} different pooling encoders, and Sect.~\ref{s:exp-deep-feat-variants} additional variants of the CNN-based descriptors.

\subsubsection{General experimental setup}\label{s:rec-descriptors}

The experiments are centered around two types of local descriptors. The first type are SIFT descriptors extracted densely from the image (denoted \emph{DSIFT}). SIFT descriptors are sampled with a step of two pixels and the support of the descriptor is scaled such that a SIFT spatial bin has size $8\times 8$ pixels. Since there are $4 \times 4$ spatial bins, the support or ``receptive field'' of each DSIFT descriptor is $40\times 40$ pixels, (including a border of half a bin due to bilinear interpolation).  Descriptors are 128-dimensional~\cite{lowe99object}, but their dimensionality is further reduced to 80 using PCA, in all experiments. Besides improving the classification accuracy, this significantly reduces the size of the Fisher Vector and VLAD encodings.

The second type of local image descriptors are deep convolutional features (denoted \emph{CNN}) extracted from the convolutional layers of CNNs pre-trained on ImageNet ILSVRC data. Most experiments build on the VGG-M model of~\cite{chatfield14return} as this network performs better than standard networks such as the Caffe reference model~\cite{jia13caffe} and AlexNet~\cite{krizhevsky12imagenet} while having a similar computational cost. The VGG-M convolutional features are extracted as the output of the last convolutional layer, directly from the linear filters excluding ReLU and max pooling, which yields a field of 512-dimensional descriptor vectors. In addition to VGG-M, experiments consider the recent  VGG-VD (very deep with 19 layers) model  of Simonyan and Zisserman~\cite{simonyan14very}. The receptive field of CNN descriptors is much larger compared to SIFT: $139\times 139$ pixels for VGG-M and $252 \times 252$ for VGG-VD.

When combined with a pooling encoder,  local descriptors are extracted at multiple scales, obtained by rescaling the image by factors $2^s, s=-3,-2.5,\dots,1.5$ (but, for efficiency, discarding scales that would make the image larger than $1024^2$ pixels). 

The dimensionality of the final representation strongly depends on the encoder type and parameters. For $K$ visual words, BoVW and LLC have $K$ dimensions, VLAD has $KD$ and FV $2KD$, where $D$ is the dimension of the local descriptors. For the FC encoder, the dimensionality is fixed by the CNN architecture; here the representation is extracted from the penultimate FC layer (before the final classification layer) of the CNNs and happens to have 4096 dimensions for all the CNNs considered. In practice, dimensions vary widely, with BoVW, LLC, and FC having a comparable dimensionality, and VLAD and FV a much higher one. For example, FV-CNN has $\simeq 64 \cdot 10^3$ dimensions with $K=64$ Gaussian mixture components, versus the 4096 of FC, BoVW, and LLC (when used with $K=4096$ visual words). In practice, however, dimensions are hardly comparable as VLAD and FV vectors are usually highly compressible~\cite{parkhi14a-compact}. We verified that by using PCA to reduce FV to 4096 dimensions and observing only a marginal reduction in classification performance in the PASCAL VOC object recognition task, as described below.

Unless otherwise specified, learning uses a standard non-linear SVM solver. Initially, cross-validation was used to select the parameter $C$ of the SVM in the range $\{0.1, 1, 10, 100\}$; however, after noting that performance was nearly identical in this range (probably due to the data normalization), $C$ was simply set to the constant 1. Instead, it was found that recalibrating the SVM scores for each class improves classification accuracy (but of course not mAP). Recalibration is obtained by changing the SVM bias and rescaling the SVM weight vector in such a way that the median scores of the negative and positive training samples for each class are mapped respectively to the values $-1$ and $1$.

All the experiments in the paper use the VLFeat library~\cite{Vedaldi10a} for the computation of SIFT features and the pooling embedding (BoVW, VLAD, FV). The MatConvNet~\cite{vedaldi14matconvnet-convolutional} library is used instead for all the experiments involving CNNs. Further details specific to the setup of each experiment are given below as needed.

\subsubsection{Datasets and evaluation measures}\label{s:rec-datsets}

The evaluation is performed on a diversity of tasks: the new describable attribute and material recognition benchmarks in DTD and OpenSurfaces, existing ones in FMD and \kthb, object recognition in PASCAL VOC 2007, and scene recognition in MIT Indoor. All experiments follow standard evaluation protocols for each dataset, as detailed below. 


{\bf DTD} (Sect.~\ref{s:dtd}) contains 47 texture classes, one per visual attribute, containing 120 images each. Images are equally spilt into train, test and validation, and include experiments on the prediction of ``key attributes'' as well as ``joint attributes'', as as defined in Sect.~\ref{s:dtd-data}, and reports accuracy averaged over the 10 default splits provided with the datasets. 
%
{\bf OpenSurfaces}~\cite{bell13opensurfaces} is used in the setup described in Sect.~\ref{s:clutter} and contains 25,357 images, out of which we selected 10,422 images, spanning across 21 categories. When segments are provided, the dataset is referred to as OS+R, and recognition accuracy is reported on a per-segment basis. We also annotated the segments with the attributes from DTD, and called this subset OSA (and OSA+R for the setup when segments are provided). For the recognition task on OSA+R we report mean average precision, as this is a multi-label dataset.

{\bf FMD}~\cite{sharan09material} consists of 1,000 images with 100 for each of ten material categories. The standard evaluation protocol of \cite{sharan09material} uses 50 images per class for training and the remaining 50 for testing, and reports classification accuracy averaged over 14 splits.
{\bf \kthb}~\cite{mallikarjuna05} contains 4,752 images, grouped into 11 material categories. For each material category, images of four samples were captured under various conditions, resulting in 108 images per sample. Following the standard procedure~\cite{caputo05class, timofte12trainingfree}, images of one material sample are used to train the model, and the other three samples for evaluating it, resulting in four possible splits of the data, for which average per-class classification accuracy is reported.
{\bf MIT Indoor Scenes}~\cite{quattoni09} contains 6,700 images divided in 67 scene categories. There is one split of the data into train (80\%) and test (20\%), provided with the dataset, and the evaluation metric is average per-class classification accuracy.
{\bf PASCAL VOC 2007}~\cite{everingham07pascal} contains 9,963 images split across 20 object categories. The dataset provides a standard split in training, validation and test data. Performance is reported in term of mean average precision (mAP) computed using the TRECVID 11-point interpolation scheme~\cite{everingham07pascal}.\footnote{The procedure for computing the AP was changed in later versions of the benchmark.}

\subsubsection{Local image descriptors and kernels comparison}\label{s:results-local}

\begin{table}[t]
\centering
{\small \setlength{\tabcolsep}{3pt}
\begin{tabular}{|l|cccc|}
\hline
                     & \multicolumn{4}{c|}{{Kernel}}                                                         \\
   {Local descr.}    & {Linear}            & {Hellinger}         & add-$\chi^2$        & exp-$\chi^2$        \\
\hline
{MR8}                & 20.8 $\pm$ 0.9      & 26.2 $\pm$ 0.8      & 29.7 $\pm$ 0.9      & 34.3 $\pm$ 1.1      \\
{LM}                 & 26.7 $\pm$ 0.9      & 34.8 $\pm$ 1.2      & 39.5 $\pm$ 1.4      & 44.0 $\pm$ 1.4      \\
{Patch$_{3\times3}$} & 15.9 $\pm$ 0.5      & 24.4 $\pm$ 0.7      & 27.8 $\pm$ 0.8      & 30.9 $\pm$ 0.7      \\
{Patch$_{7\times7}$} & 20.7 $\pm$ 0.8      & 30.6 $\pm$ 1.0      & 34.8 $\pm$ 1.0      & 37.9 $\pm$ 0.9      \\
{LBP$^{u}$}          &  8.5 $\pm$ 0.4      &  9.3 $\pm$ 0.5      & 12.5 $\pm$ 0.4      & 19.4 $\pm$ 0.7      \\
{LBP-VQ}             & 26.2 $\pm$ 0.8      & 28.8 $\pm$ 0.9      & 32.7 $\pm$ 1.0      & 36.1 $\pm$ 1.3      \\
{SIFT}               & \bf{45.2 $\pm$ 1.0} & \bf{49.1 $\pm$ 1.1} & \bf{50.9 $\pm$ 1.0} & \bf{52.3 $\pm$ 1.2} \\
{Conv VGG-M}         & \bf{55.9 $\pm$ 1.3} & \bf{61.7 $\pm$ 0.9} & \bf{61.9 $\pm$ 1.0} & \bf{61.2 $\pm$ 1.0} \\
{Conv VGG-VD}        & \bf{64.1 $\pm$ 1.3} & \bf{68.8 $\pm$ 1.3} & \bf{69.0 $\pm$ 0.9} & \bf{68.8 $\pm$ 0.9} \\
\hline
\end{tabular}
}
\caption{Comparison of local features and kernels on the DTD data. The table reports classification accuracy, averaged over the predefined ten splits, provided with the dataset.
We marked in bold the best performing descriptors, SIFT and convolutional features, which we will cover in the following experiments and discussions.}
\label{tbl:res-local-feats}
\end{table}

\begin{figure*}
\includegraphics[width=\textwidth,height=0.5\textwidth]{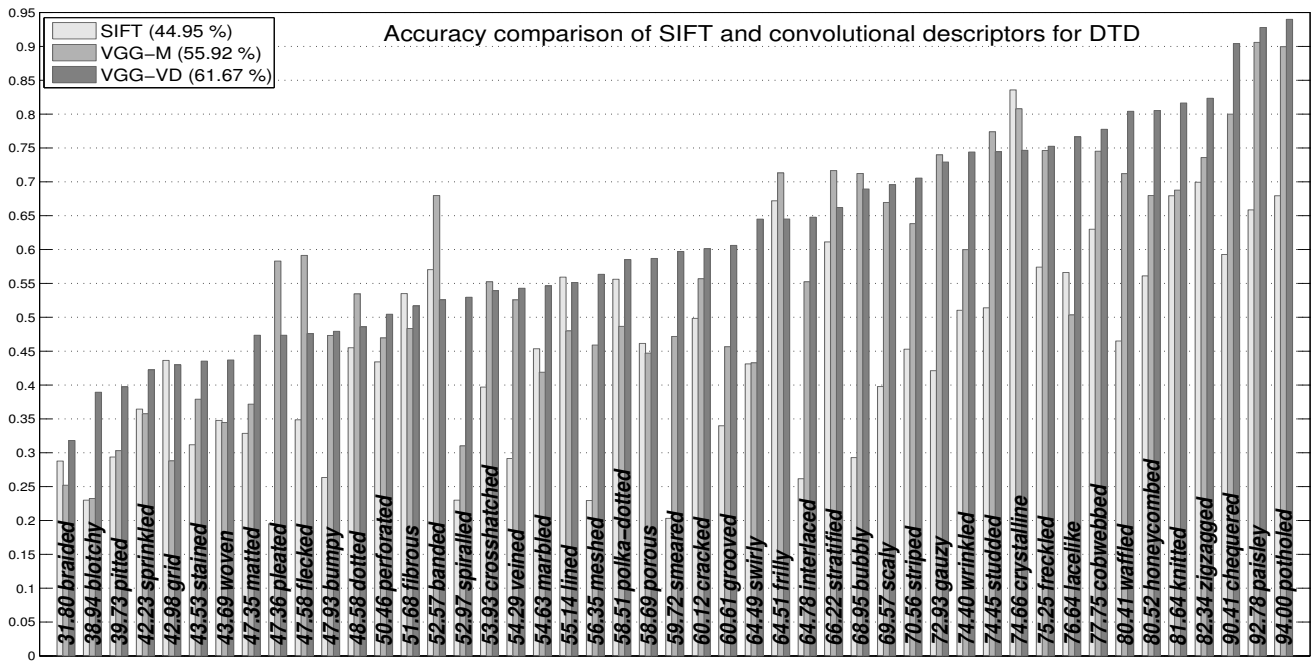}
\caption{Per class classification accuracy in the DTD data comparing three local image descriptors: SIFT, VGG-M, and VGG-VD. For all three local descriptors, BoVW with 4096 visual words was used. Classes are sorted by increasing BoVW-CNN-VD accuracy (this number is reported along each bar).}\label{fig:dtd-per-class-acc}
\end{figure*}

The goal of this section is to establish which local image descriptors work best in a texture representation. The question is relevant because: (i) while SIFT is the de-facto standard handcrafted-feature in object and scene recognition, most authors use specialized descriptors for texture recognition and (ii) learned convolutional features in CNNs have not yet been compared when used as local descriptors (instead, they have been compared to classical image representations when used in combination with their FC layers).

The experiments are carried on the the task of recognizing describable texture attributes in DTD (Sect.~\ref{s:dtd}) using the BoVW encoder. As a byproduct, the experiments determine the relative difficulty of recognizing the different 47 perceptual attributes in DTD.

\paragraph{Experimental setup.} The following local image descriptors are compared: the linear filter banks of \emph{Leung and Malik} (LM)~\cite{leung2001representing} (48D descriptors) and \emph{MR8} (8D descriptors)~\cite{varma2005statistical,geusebroek03fast}, the $3 \times 3$ and $7\times 7$ raw image patches of~\cite{varma2003texture} (respectively 9D and 49D), the {\em local binary patterns} (LBP) of~\cite{ojala2002multiresolution} (58D), SIFT (128D), and CNN features extracted from \emph{VGG-M} and \emph{VGG-VD} (512D).

After the BoVW representation is extracted, it is used to train a 1-vs-all SVM using the different kernels discussed in Sect.~\ref{s:post-processing}: linear, Hellinger, additive-$\chi^2$, and exponential-$\chi^2$. Kernels are normalized as described before. The exponential-$\chi^2$ kernel requires choosing the parameter $\lambda$; this is set as the reciprocal of the mean of the $\chi^2$ distance matrix of the training BoVW vectors. Before computing the exponential-$\chi^2$ kernel, furthermore, BoVW vectors are $L^1$ normalized.  An important parameter in BoVW is the number of visual words selected. $K$ was varied in the range of 256, 512, 1024, 2048, 4096 and performance evaluated on a validation set. Regardless of the local feature and embedding, performance was found to increase with $K$ and to saturate around $K=4096$ (although the relative benefit of increasing $K$ was larger for  features such as SIFT and CNNs). Therefore $K$ was set to this value in all experiments.

\paragraph{Analysis.} Table~\ref{tbl:res-local-feats} reports the classification accuracy for 47 1-vs-all SVM attribute classifiers, computed as~\eqref{e:acc}. As often found in the literature, the best kernel was found to be exponential-$\chi^2$, followed by additive-$\chi^2$, Hellinger's, and linear kernels. Among the hand-crafted descriptors, dense SIFT significantly outperforms the best specialized texture descriptor on the DTD data (52.3\% for BoVW-exp-$\chi^2$-SIFT vs 44\% for BoVW-exp-$\chi^2$-LM). CNN local descriptors handily outperform handcrafted features by a 10-15\% recognition accuracy margin. It is also interesting to note that the choice of kernel function has a much stronger effect for image patches and linear filters (\eg accuracy nearly doubles moving from BoVW-linear-patches to BoVW-exp-$\chi^2$-patches) and an almost negligible effect for the much stronger CNN features.

Fig.~\ref{fig:dtd-per-class-acc} reports the classification accuracy for each attribute in DTD for the BoVW-SIFT, BoVW-VGG-M, and BoVW-VGG-VD descriptors and the additive-$\chi^2$ kernel. As it may be expected, concepts such as \emph{chequered}, \emph{waffled}, \emph{knitted}, \emph{paisley} achieve nearly perfect classification, while others such as \emph{blotchy}, \emph{smeared} or \emph{stained} are far harder.

\paragraph{Conclusions.}
The conclusions are that (i) SIFT descriptors outperform significantly texture-specific descriptors such as linear filter banks, patches, and LBP on this texture recognition task, and that (ii) learned convolutional local descriptors significantly surpass SIFT.

\subsubsection{Pooling encoders}\label{s:exp-global-descriptors}

{

\newcolumntype{d}{D{.}{.}{2.3}}
\makeatletter
\newcolumntype{b}{>{\boldmath\DC@{.}{.}{2.3}}c<{\DC@end}}
\makeatother

\renewcommand{\BF}[1]{\multicolumn{1}{b}{#1}}
\newcommand{\BFc}[1]{\multicolumn{1}{b|}{#1}}
\newcommand{\hd}[1]{\multicolumn{1}{c}{#1}}
\newcommand{\hdc}[1]{\multicolumn{1}{c|}{#1}}
\renewcommand{\PM}{\scriptscriptstyle\pm}
\setlength{\tabcolsep}{5.1pt}

\begin{table*}[th]
\begin{center}
\begin{tabular}{|l||c|dddd|dddd|d|}
\hline
\multirow{2}{*}{Dataset} & meas. & \multicolumn{4}{c|}{SIFT} & \multicolumn{4}{c|}{VGG-M} & \hdc{VGG-M} \\
             & (\%)  & \hd{BoVW}    & \hd{LLC}     & \hd{VLAD}    & \hdc{IFV}    & \hd{BoVW}    & \hd{LLC}     & \hd{VLAD}         & \hdc{IFV}          & \hdc{FC}      \\
\hline                                                                                                                                                   
\hline                                                                                                                                                   
{DTD}        & acc   & 49.0 \PM 0.8 & 48.2 \PM 1.4 & 54.3 \PM 0.8 & 58.6 \PM 1.2 & 61.2 \PM 1.3 & 64.0 \PM 1.3 & \BF{67.6 \PM 0.7} & \BFc{66.8 \PM 1.5} & 58.7 \PM 0.9 \\
{OS+R}       & acc   & 30.0         & 30.8         & 32.5         & 39.8         & 41.3         & 45.3         & 49.7              & \BFc{52.5}         & 41.3         \\
{\ktb}       & acc   & 57.6 \PM 1.5 & 56.8 \PM 2.0 & 64.3 \PM 1.3 & 70.2 \PM 1.6 & 73.6 \PM 2.8 & 74.0 \PM 3.3 & 72.2 \PM 4.7      & \BFc{73.3 \PM 4.8} & 71.0 \PM 2.3 \\
{FMD}        & acc   & 50.5 \PM 1.7 & 48.4 \PM 2.2 & 54.0 \PM 1.3 & 59.7 \PM 1.6 & 67.9 \PM 2.2 & 71.7 \PM 2.1 & \BF{74.2 \PM 2.0} & \BFc{73.5 \PM 2.0} & 70.3 \PM 1.8 \\
\hline                                                                                                                                                   
{VOC07}      & mAP11 & 51.2         & 47.8         & 56.9         & 59.9         & 72.9         & 75.5         & \BF{76.8}         & \BFc{76.4}         & \BFc{76.8}   \\
 \hline                                                                                                                                                   
{MIT Indoor} & acc   & 47.7         & 39.2         & 51.0         & 54.9         & 69.1         & 68.9         & 71.2              & \BFc{74.2}         & 62.5         \\
\hline
\end{tabular}
\end{center}
\caption{Pooling encoder comparisons. The table compares the orderless pooling encoders BoVW, LLC, VLAD, and IFV with either SIFT local descriptors and VGG-M CNN local descriptors (\dcnn). It also compares pooling convolutional features with the CNN fully connected layers (\rcnn). The table reports classification accuracies for all datasets except VOC07 and OS+R for which mAP-11~\cite{everingham07pascal} and mAP are reported, respectively.}
\label{tbl:encodings-sift-vs-cnn-rcnn}
\end{table*}
}

The previous section established the primacy of SIFT and CNN local image descriptors on alternatives. The goal of this section is to determine which pooling encoders (Sect.~\ref{s:global-features}) work best with these descriptors, comparing the orderless BoVW, LLC, VLAD, FV  encoders and the order-sensitive FC encoder. The latter, in particular, reproduces the CNN transfer learning setting commonly found in the literature where CNN features are extracted in correspondence to the FC layers of a network.

\paragraph{Experimental setup.}
The experimental setup is similar to the previous experiment: the same SIFT and CNN VGG-M descriptors are used; BoVW is used in combination with the Hellinger kernel (the exponential variant is slightly better, but much more expensive); the same $K=4096$ codebook size is used with LLC. VLAD and FV use a much smaller codebook as these representations multiply the dimensionality of the descriptors (Sect.~\ref{s:rec-descriptors}). Since SIFT and CNN features are respectively 128 and 512-dimensional, $K$ is set to 256 and 64 respectively. The impact of varying the number of visual words in the FV representation is further analyzed in Sect.~\ref{s:exp-deep-feat-variants}.

Before pooling local descriptors with a FV, these are usually de-correlated by using PCA whitening. Here PCA is applied to SIFT, additionally reducing its dimension to 80, as this was empirically shown to improve recognition performance. The effect of PCA-reduction to the convolutional features is studied in Section~\ref{s:pca}. The improved version of the FV is used in all the experiments (Sect.~\ref{s:clutter}), and, similarly, for VLAD, we applied signed square root to the resulting encoding, which is then normalized component-wise (Sect.~\ref{s:post-processing}).

\paragraph{Analysis.}
Results are reported in Table~\ref{tbl:encodings-sift-vs-cnn-rcnn}. In term of orderless encoders, BoVW and LLC result in similar performance for SIFT, while the difference is slightly larger and in favor of LLC for CNN features.  Note that BoVW is used with the Hellinger kernel, which contributes to reducing the gap between BoVW and LLC. IFV and VLAD significantly outperform BoVW and LLC in almost all tasks; FV is definitely better than VALD with SIFT features and about the same with CNN features. CNN features maintain a healthy lead on SIFT features regardless of the encoder used. Importantly, VLAD and FV (and to some extent BoVW and LLC) perform either substantially better or as well as the original FC encoders. Some of these observations can are confirmed by other experiments such as Table~\ref{tbl:deep-net-flavors-soa}.

Next, we compare using CNN features with an orderless encoder (\dcnn) as opposed to the standard FC layer (\rcnn). As seen in Table~\ref{tbl:encodings-sift-vs-cnn-rcnn} and Table~\ref{tbl:deep-net-flavors-soa}, in PASCAL VOC and MIT Indoor the \rcnn descriptor performs very well but in line with previous results for this class of methods~\cite{chatfield14return}. \dcnn performs similarly to \rcnn in PASCAL VOC, \ktb and FMD, but substantially better for DTD, OS+R, and MIT Indoor (e.g. for the latter $+5\%$ for VGG-M and $+13\%$ for VGG-VD).

As a sanity check, results are within 1\% of the ones reported in~\cite{chatfield11devil} and~\cite{chatfield14return} for matching experiments on FV-SIFT and FC-VGG-M. The differences in case of SIFT LLC and BoVW are easily explained by the fact that, differently from \cite{chatfield11devil},  our present experiments do not use SPP and image augmentation.

\paragraph{Conclusions.}
The conclusions of these experiments are that: (i) IFV and VLAD are preferable to other orderless pooling encoders, that (ii) orderless pooling encoders such as the FV are at least as good and often significantly better than FC pooling with CNN features.

\subsubsection{CNN descriptor variants comparison}\label{s:exp-deep-feat-variants}

\begin{table*}[ht]
\setlength{\tabcolsep}{2pt}
\centering
\begin{tabular}{|cl|c|c|ccc|ccc|ccc|c|c|}
\hline
\multicolumn{2}{|c|}{dataset} &  meas. & SIFT & \multicolumn{3}{c|}{AlexNet} & \multicolumn{3}{c|}{VGG-M} & \multicolumn{3}{c|}{VGG-VD} & FV-SIFT & \multirow{2}{*}{SoA} \\
\multicolumn{2}{|c|}{} &              (\%)  &        FV            & FC            & FV                 & FC+FV              & FC            & FV                 & FC+FV              & FC            & FV                 & FC+FV              & FC+FV-VD           & \\

\hline                                                                                                                                      
\hline                                                                                                                                      
\multirow{6}{*}{(a)} & CUReT              & acc      & 99.0\pmt{0.2}      & 94.4\pmt{0.4} & 98.5\pmt{0.2}      & 99.0\pmt{0.2}      & 94.2\pmt{0.3} & 98.7\pmt{0.2}      & 99.1\pmt{0.2}      & 94.5\pmt{0.4} & 99.0\pmt{0.2}      & 99.2\pmt{0.2}      & \bf{99.7\pmt{0.1}} & 99.8\pmt{0.1} \cite{sifre13rotation} \\
                     & UMD                & acc      & 99.1\pmt{0.5}      & 95.9\pmt{0.9} & \bf{99.7\pmt{0.2}} & \bf{99.7\pmt{0.3}} & 97.2\pmt{0.9} & \bf{99.9\pmt{0.1}} & \bf{99.8\pmt{0.2}} & 97.7\pmt{0.7} & \bf{99.9\pmt{0.1}} & \bf{99.9\pmt{0.1}} & \bf{99.9\pmt{0.1}} & 99.7\pmt{0.3} \cite{sifre13rotation} \\
                     & UIUC               & acc      & 96.6\pmt{0.8}      & 91.1\pmt{1.7} & 99.2\pmt{0.4}      & 99.3\pmt{0.4}      & 94.5\pmt{1.4} & \bf{99.6\pmt{0.4}} & 99.6\pmt{0.3}      & 97.0\pmt{0.7} & \bf{99.9\pmt{0.1}} & \bf{99.9\pmt{0.1}} & \bf{99.9\pmt{0.1}} & 99.4\pmt{0.4} \cite{sifre13rotation} \\
                     & KT                 & acc      & \bf{99.5\pmt{0.5}} & 95.5\pmt{1.3} & \bf{99.6\pmt{0.4}} & \bf{99.8\pmt{0.2}} & 96.1\pmt{0.9} & \bf{99.8\pmt{0.2}} & \bf{99.9\pmt{0.1}} & 97.9\pmt{0.9} & \bf{99.8\pmt{0.2}} & \bf{99.9\pmt{0.1}} & \bf{100}           & 99.4\pmt{0.4} \cite{sifre13rotation} \\
                     & ALOT               & acc      & 94.6\pmt{0.3}      & 86.0\pmt{0.4} & \bf{96.7\pmt{0.3}} & \bf{97.8\pmt{0.2}} & 88.7\pmt{0.5} & \bf{97.8\pmt{0.2}} & \bf{98.4\pmt{0.1}} & 90.6\pmt{0.4} & \bf{98.5\pmt{0.1}} & \bf{99.0\pmt{0.1}} & {99.3\pmt{0.1}}    & 95.9\pmt{0.5} \cite{sulc14fast}      \\
\hline                                                                                                                                                                                                                                                              
\multirow{3}{*}{(b)} & \ktb               & acc      & 70.8\pmt{2.7}      & 71.5\pmt{1.3} & 69.7\pmt{3.2}      & 72.1\pmt{2.8}      & 71\pmt{2.3}   & 73.3\pmt{4.7}      & 73.9\pmt{4.9}      & 75.4\pmt{1.5} & \bf{81.8\pmt{2.5}} & \bf{81.1\pmt{2.4}} & \bf{81.5\pmt{2.0}} & 76.0\pmt{2.9} \cite{sulc14fast}                          \\
                     & FMD                & acc      & 59.8\pmt{1.6}      & 64.8\pmt{1.8} & 67.7\pmt{1.5}      & 71.4\pmt{1.7}      & 70.3\pmt{1.8} & 73.5\pmt{2.0}      & 76.6\pmt{1.9}      & 77.4\pmt{1.8} & 79.8\pmt{1.8}      & \bf{82.4\pmt{1.5}} & \bf{82.2\pmt{1.4}} & 57.7\pmt{1.7} \cite{sharan13recognizing} \\
                     & OS+R               & acc      & 39.8               & 36.8          & 46.1               & 49.8               & 41.3          & 52.5               & 54.9               & 43.4          & 59.5               & 60.9               & 58.7               & --                                                       \\
\hline                                                                                                                                                                                                                                                              
\multirow{4}{*}{(c)} & DTD                & acc      & 58.6\pmt{1.2}      & 55.1\pmt{0.6} & 62.9\pmt{1.4}      & 66.5\pmt{1.1}      & 58.8\pmt{0.8} & \bf{66.8\pmt{1.6}} & \bf{69.8\pmt{1.1}} & 62.9\pmt{0.8} & \bf{72.3\pmt{1.0}} & \bf{74.7\pmt{1.0}} & \bf{75.5\pmt{0.8}} & --                                                       \\
                     & DTD                & mAP      & 61.3\pmt{1.1}      & 57.7\pmt{0.9} & 66.5\pmt{1.4}      & 70.5\pmt{1.2}      & 62.1\pmt{0.9} & \bf{70.8\pmt{1.2}} & \bf{74.2\pmt{1.1}} & 67.0\pmt{1.1} & \bf{76.7\pmt{0.8}} & 
\bf{79.1\pmt{0.8}}   & \bf{80.4\pmt{0.9}} & --                                                       \\
                     & DTD-J              & mAP      & 59.6\pmt{0.6}      & 58.4\pmt{0.7} & 65.0\pmt{0.9}      & 68.3\pmt{0.9}      & 62.8\pmt{0.7} & \bf{69.8\pmt{0.9}} & \bf{72.9\pmt{0.9}} & 67.3\pmt{0.9} & \bf{75.8\pmt{0.6}} & \bf{77.5\pmt{0.8}} & \bf{78.9\pmt{0.7}} & --                                                       \\
                     & OSA+R              & mAP      & 56.5               & 53.9          & 62.1               & 64.6               & 54.3          & 65.2               & 67.9               & 49.7          & 67.2               & 67.9               & 68.2               & --                                                       \\
\hline                                                                                                                                                                                                                                                              
\multirow{5}{*}{(d)} & MSRC+R             & acc      & 85.7               & 83.6          & 91.7               & 94.9               & 85.0          & 95.4               & 96.9               & 79.4          & 97.7               & 98.8               & 99.1               & --                                       \\
                     & MSRC+R             & msrc-acc & 92.0               & 84.1          & 95.0               & 97.3               & 84.0          & 97.6               & 98.1               & 82.0          & 99.2               & 99.6               & 99.5               & --                                       \\
                     & VOC07              & mAP11    & 59.9               & 74.0          & 73.1               & 76.8               & 76.8          & 76.4               & 79.5               & 81.7          & 84.9               & 85.1               & 84.5               & 85.2 \cite{wei14cnn:}                    \\
                     & VOC07              & mAP      & 60.2               & 76.0          & 75.0               & 79.0               & 79.2          & 78.7               & 82.3               & 84.6          & 88.6               & 88.5               & 87.9               & 85.2 \cite{wei14cnn:}                    \\

                     & MIT Ind.           & acc      & 54.9               & 58.6          & 69.7               & 71.6               & 62.5          & 74.2               & 74.4               & 67.6          & 81.0               & 80.3               & 80.0               & 70.8 \cite{zhou14learning}               \\
                     & CUB                & acc      & 17.5               & 45.8          & 49                 & 54.1               & 46.1          & 49.9               & 54.9               & 54.6          & \bf{66.7}          & \bf{67.3}          & \bf{65.4}          & 73.9$^*$ \cite{zhang14part-based} \\
                     & CUB+R              & acc      & 27.7               & 54.5          & 62.6               & 65.2               & 56.5          & 65.5               & 68.1               & 62.8          & 73.0               & 74.9               & 73.6               & 76.37 \cite{zhang14part-based}           \\
\hline
\end{tabular}
\caption{State of the art texture descriptors. The table compares \rcnn, \dcnn on three networks trained on ImageNet -- VGG-M, VGG-VD and AlexNet, and IFV on dense SIFT.
We evaluated these descriptors on (a) texture datasets -- in controlled settings, (b) material datasets (FMD, \ktb, OS+R), (c) texture attributes (DTD, OSA+R) and (d) general categorisation datasets (MSRC+R, VOC07, MIT Indoor) and fine grained categorisation (CUB, CUB+R). For this experiment the region support is assumed to be known (and equal to the entire image for all the datasets except OS+R and MSRC+R and for CUB+R, where it is set to the bounding box of a bird). $^*$using a model without parts like ours the performance is 62.8\%.}
\label{tbl:deep-net-flavors-soa}
\end{table*}

\begin{figure*}[t]
\includegraphics[width=\textwidth]{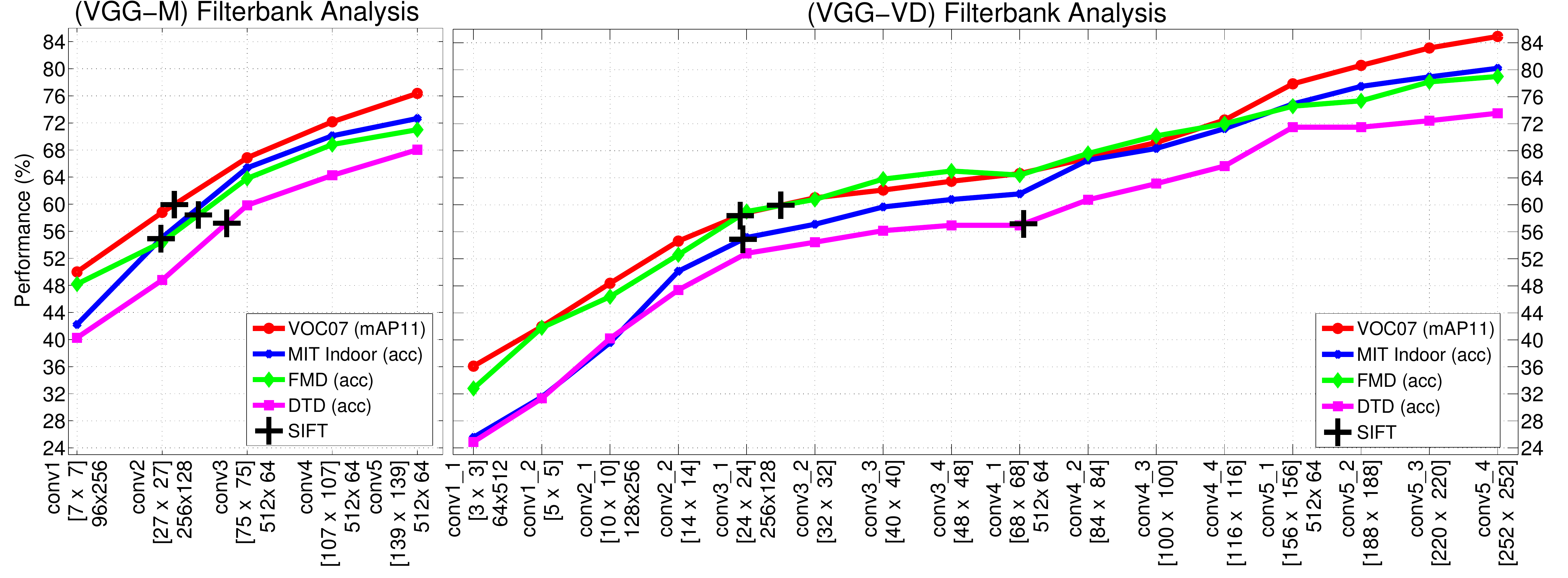}
\caption{\textbf{Effect of the depth on CNN features.} The figure reports the performance of VGG-M (left) and VGG-VD (right) local image descriptors pooled with the FV encoder. For each layer the figures shows the size of the receptive field of the local descriptors (denoted $[N\times N]$]), as well as, for some of the layers, the dimension $D$ of the local descriptors and the number $K$ of visual words in the FV representation (denoted as $D \times K$). Curves for PASCAL VOC, MIT Indoor, FMD, and DTD are reported; the performance of using SIFT as local descriptors is reported as a plus (+) mark.}
\label{f:filterbank}
\end{figure*} 

\begin{figure}[t]
\centering
\includegraphics[width=\linewidth]{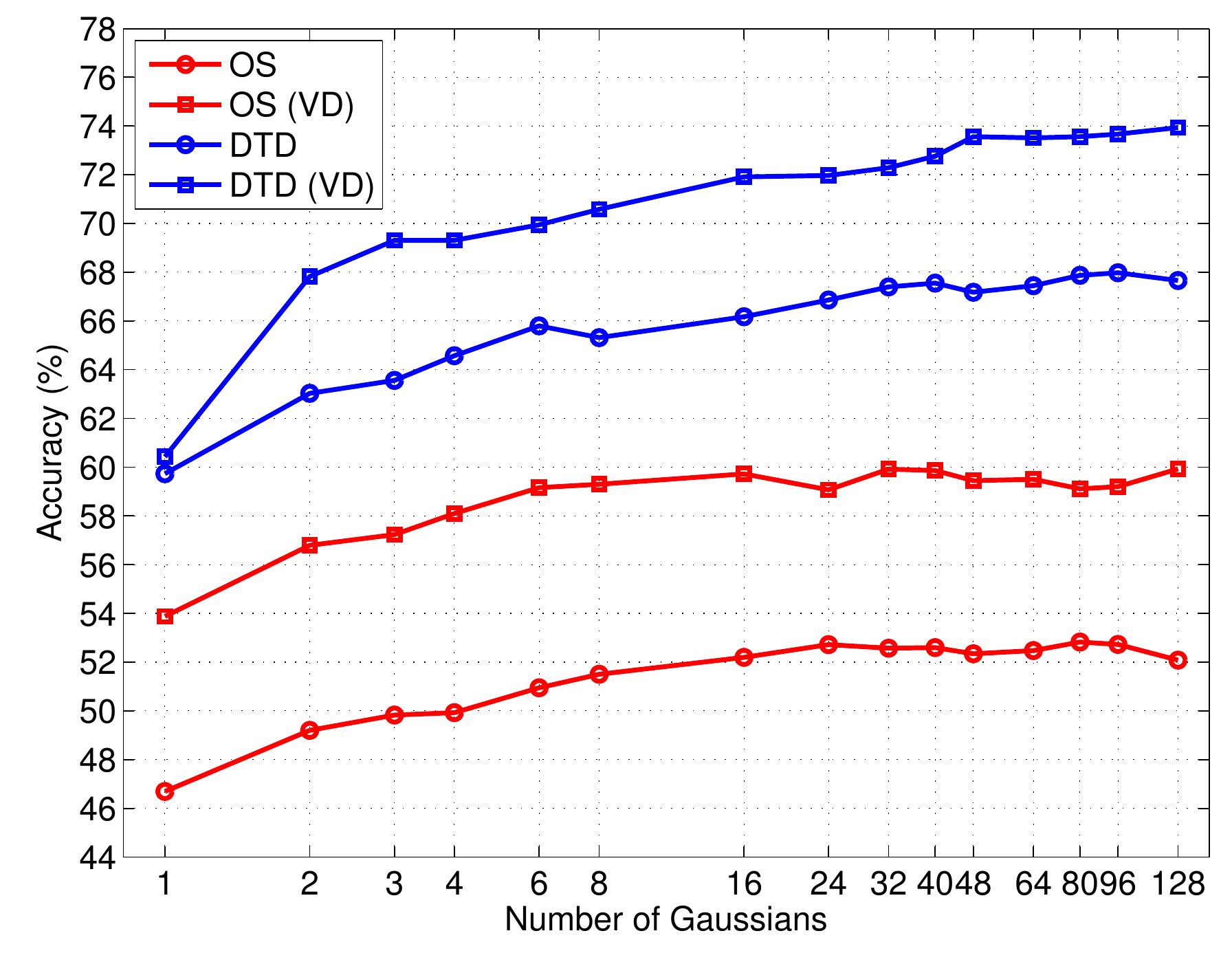}
\caption{\textbf{Effect of the number of Gaussian components in the FV encoder.}  The figure shows the performance of the FV-VGG-M and FV-VGG-VD representations on the OS and DTD datasets when the number of Gaussians components in the GMM is varied from 1 to 128. Note that the abscissa is scaled logarithmically.}\label{f:num-gmm}
\end{figure}

This section conducts additional experiments on CNN local descriptors to find the best variants.

\paragraph{Experimental setup.} The same setup of the previous section is used. We compare the performance of \rcnn and \dcnn local descriptors obtained from VGG-M, VGG-VD as well as the simpler AlexNet \cite{krizhevsky12imagenet} CNN which is widely adopted in the literature.

\paragraph{Analysis.} Results are presented in detail in Table~\ref{tbl:deep-net-flavors-soa}. Within that table, the analysis here focuses mainly on texture and material datasets, but conclusions are similar for the other datasets. In general, VGG-M is better than AlexNet and VGG-VD is substantially better than VGG-M (e.g. on FMD, FC-AlexNet obtains 64.8\%, FC-VGG-M obtains 70.3\% (+5.5\%), FC-VGG-VD obtains 77.4\% (+7.1\%)). However, switching from FC to FV pooling improves the performance more than switching to a better CNN (\eg on DTD going from FC-VGG-M to FC-VGG-VD yields a 7.1\% improvement, while going from FC-VGG-M to FV-VGG-M yields a 11.3\% improvement). Combining \dcnn and \rcnn (by stacking the corresponding image representations) improves the accuracy by 1-2\% for VGG-VD, and up to 3-5\% for VGG-M. There is no significant benefit from adding FV-SIFT as well, as the improvement is at most 1\%, and in some cases (MIT, FMD) it degrades the performance.


Next, we analyze in detail the effect of depth on the convolutional features. Fig.~\ref{f:filterbank} reports the accuracy of VGG-M and VGG-VD on several datasets for features extracted at increasing depths. The pooling method is fixed to FV and the number of Gaussian centers $K$ is set such that the overall dimensionality of the descriptor $2KD_k$ is constant. For both VGG-M and VGG-VD, the improvement with increasing depth is substantial and the best performance is obtained by the deepest features (up to 32\% absolute accuracy improvement in VGG-M and up to 48\% in VGG-VD).  Performance increases at a faster rate up to the third convolutional layer (\texttt{conv3}) and then the rate tapers off somewhat. The performance of the earlier levels in VGG-VD is much worse than the corresponding layers in VGG-M. In fact, the performance of VGG-VD matches the performance of the deepest (fifth) layer in VGG-M in correspondence of \texttt{conv5\_1}, which has depth 13.

Finally, we look at the effect of the number of Gaussian components (visual words) in the \dcnn representation, testing possible values in the range 1 to 128 in small (1-16) increments. Results are presented in Fig.~\ref{f:num-gmm}. While there is a substantial improvement in moving from one Gaussian component to about 64 (up to +15\% on DTD and up to 6\% on OS), there is little if any advantage at increasing the number of components further.

\paragraph{Conclusions.}
The conclusions of these experiments are as follows: (i) deeper models substantially improve performance; (ii) switching from FC to FV pooling has an ever more substantial impact, particularly for deeper models; (iii) combining FC and FV pooling has a modest benefit and there is no benefit in integrating SIFT features; (iv) in very deep models, most of the performance gain is realized in the very last few layers.

\subsubsection{FV pooling vs FC pooling}\label{s:fv-vs-fc}

\begin{table*}[ht]
\setlength{\tabcolsep}{2pt}
\centering
\begin{tabular}{|l|c|cccc|cccc|}
\hline
{dataset} &  meas. &\multicolumn{4}{c|}{VGG-M} & \multicolumn{4}{c|}{VGG-VD}  \\
 &              (\%)  &        FC (SS)           & FC  (MS)            & FV  (SS)               & FV (MS)              & FC (SS)           & FC (MS)            & FV  (SS)               & FV (MS)               \\

\hline                                                                                                                                      
\hline                                                                                                                                      
\ktb        & acc   & 71\pmt{2.3}    & 68.9\pmt{3.9}  &  69.0\pmt{2.8} & 73.3\pmt{4.7}   &  75.4\pmt{1.5} & 75.1\pmt{3.8} & 74.5\pmt{4.4} & 81.8\pmt{2.5} \\
FMD         & acc   & 70.3\pmt{1.8}  & 69.3\pmt{1.8}  &  71.6\pmt{2.4} & 73.5\pmt{2.0}   &  77.4\pmt{1.8} & 78.1\pmt{1.7} & 79.4\pmt{2.5} & 79.8\pmt{1.8} \\
DTD         & acc   & 58.8\pmt{0.8}  & 59.9\pmt{1.1}  & 62.8 \pmt{1.5} & 66.8\pmt{1.6}   &  62.9\pmt{0.8} & 65.3\pmt{1.5} & 69.2\pmt{0.8} & 72.3\pmt{1.0} \\
VOC07       & mAP11 & 76.8           & 78             & 74.8           & 76.4            &  81.7          & 83.2          & 84.7          & 84.9 \\
MIT Ind.    & acc   & 62.5           & 66.1           & 68.1           & 74.2            &  67.6          & 75.3          & 76.8          & 81.0 \\
\hline
\end{tabular}
\caption{The table the single and multi-scale variants of \rcnn and \dcnn using two deep CNN, VGG-M and VGG-VD, trained on the ImageNet ILSVRC, and a number of representative target datasets.  The single scale variants are denoted FC (SS) and FV (SS) and the multi-scale variants as FC (MS) and FV (MS).}
\label{tbl:res-fv-ss}
\end{table*}

In the previous section, we have seen that switching from FC to FV pooling may have a substantial impact in certain problems. We could find three reasons that can explain this difference.

The first reason is that orderless pooling in FV can be more suitable for {\bf texture modeling} than the order-sensitive pooling in FC. However, this explains the advantage of FV in texture recognition but not in object recognition.

The second reason is that FV pooling may reduce {\bf overfitting in domain transfer}.  Pre-trained FC layers could be too specialized for the source domain (e.g. ImageNet ILSVRC) and there may not be enough training data in the target domain to retrain them properly. On the contrary, a linear classifier built on FV pooling is less prone to overfitting as it encodes a simpler, smoother classification function than a sequence of FC layers in a CNN. This is further investigated in~Sect.~\ref{s:exp-places}.

The third reason is the ability to easily incorporate information from {\bf multiple image scales}. 

 In order to investigate this hypothesis, we evaluated FV-CNN by pooling CNN descriptors at a single scale instead of multiple ones, for both VGG-M and VGG-VD models. For datasets like FMD, DTD and MIT Indoor, FV-CNN at a single scale still generally outperforms FC-CNN (columns FC (SS) and FV (SS) in Table~\ref{tbl:res-fv-ss}), by up to 5.6\% for VGG-M, and by up to 9.1\% for VGG-VD; however, the difference is less marked as using a single scale in FV-CNN looses up to 3.8\% accuracy points and and in some cases the representations is overtaken by FC-CNN.
 
 The complementary experiment, namely using multiple scales in FC pooling, is less obvious as, by construction, FC-CNN resizes the input image to a fixed resolution. However, we can relax this restriction by computing multiple FC representations in a sliding-window manner  (also know as a ``fully-convolutional'' network). Then individual representations computed at multiple locations and, after resizing the image, at multiple scales can be averaged in a single representation vector. We refer to this as multi-scale FC pooling. Multi-scale FC codes perform slightly better than single-scale FC in most (but not all) cases; however, the benefit of using multiple scales is not as large as for multi-scale FV pooling, which is still significantly better than multi-scale FC.

\subsubsection{Dimensionality reduction of the CNN descriptors}\label{s:pca}

\begin{figure*}[t]
\includegraphics[width=0.49\textwidth]{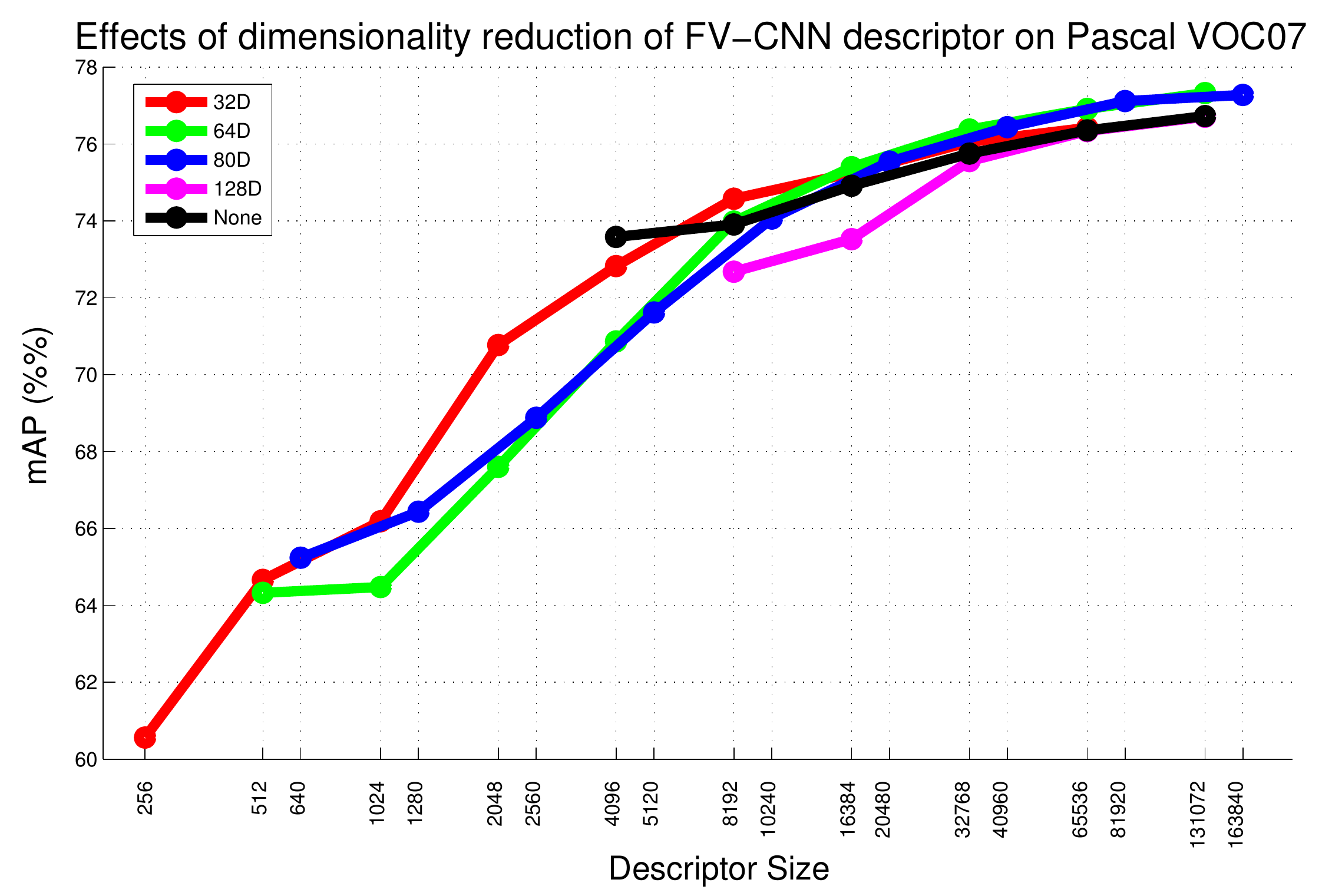}
\includegraphics[width=0.49\textwidth]{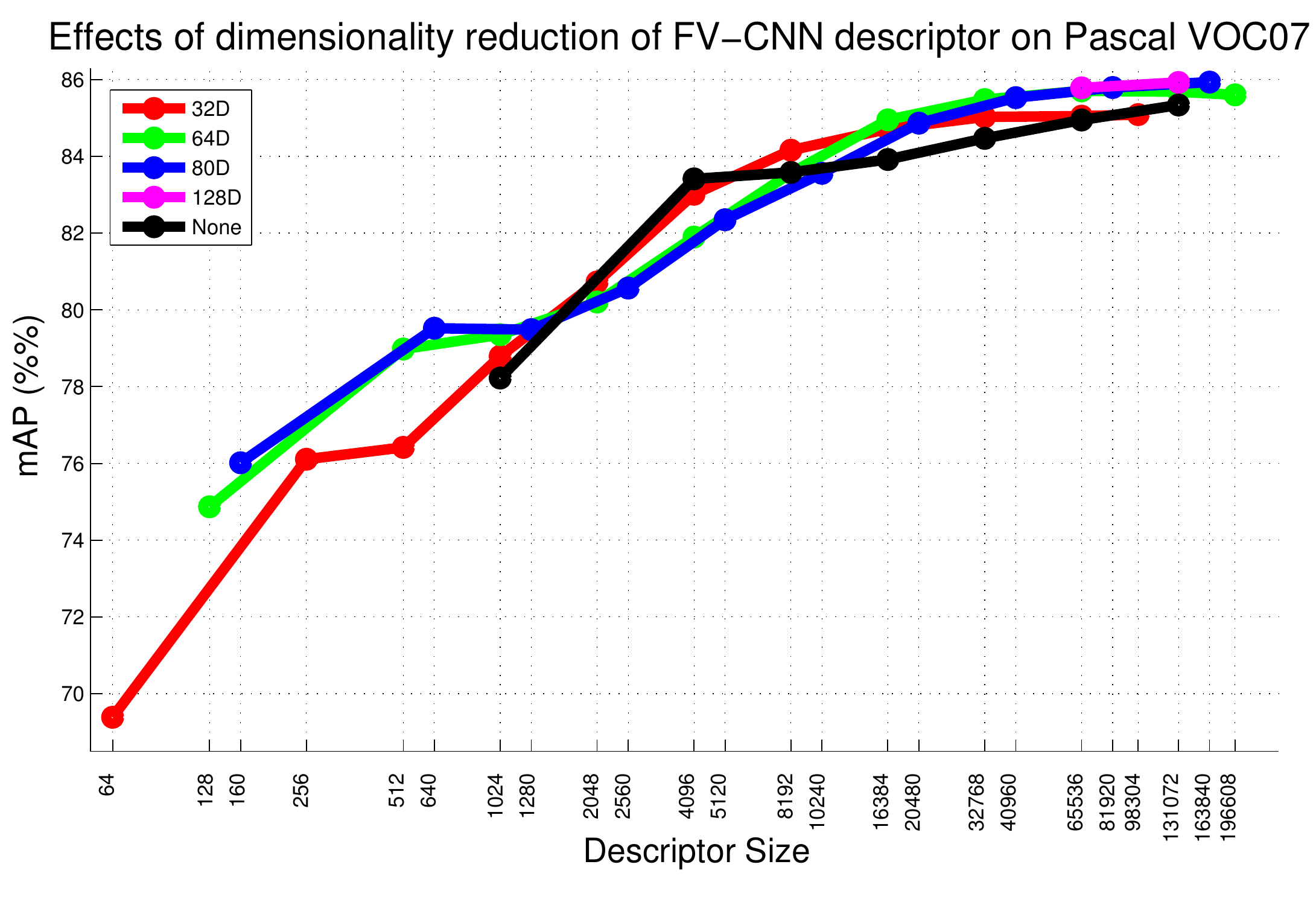}
\caption{\textbf{PCA reduced \dcnn}. The figure reports the performance of VGG-M (left) and VGG-VD (right) local descriptors, on PASCAL VOC 2007, when reducing their dimensionality from 512 to up to 32 using PCA in combination with a variable number of GMM components. The horizontal axis report the total descriptor dimensionality, proportional to the dimensionality of the local descriptors by the number of GMM components.}
\label{f:pca}
\end{figure*}

This section explores the effect of applying dimensionality reduction to the CNN local descriptors before FV pooling.

This experiment investigates the effect of two parameters, the number of Gaussians in the mixture model used by the FV encoder, and the dimensionality of the convolutional features, which we reduce using PCA. Various  local descriptor dimensions are evaluated, from 512 (no PCA) to 32, reporting mAP on PASCAL VOC 2007, as a function of the pooled descriptor dimension. The latter is equal to $2KD$, where $K$ is the number of Gaussian centers, and $D$ the dimensionality of the local descriptor after PCA reduction.

Results are presented in Figure~\ref{f:pca} for VGG-M and VGG-VD. It can be noted that, for similar values of the total representation dimensionality $2KD$, the performance of PCA-reduced descriptors is a little better than not using PCA, provided that this is compensated by a large number of GMM components. In particular,  similar to what was observed for SIFT in~\cite{perronnin10improving}, using PCA does improve the performance by 1-2\% mAP point; furthermore, reducing descriptors to 64 or 80 dimensions appears to result in the best performance.



\subsubsection{Visualization of descriptors}

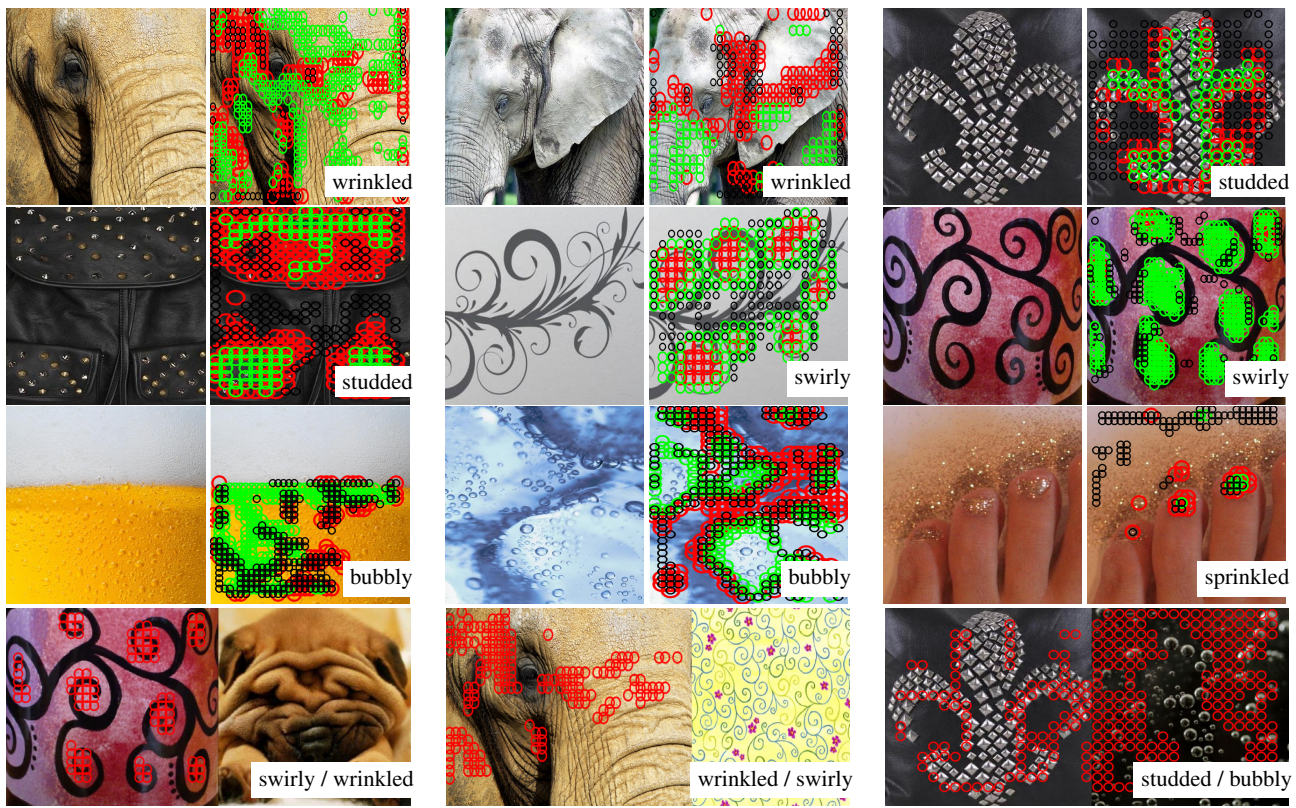
\begin{figure*}[t]
\newcommand{\puti}[2]
{
\begin{tikzpicture}
\node[anchor=south east,inner sep=0] at (0,0) {%
\includegraphics[width=0.15\textwidth, height=0.15\textwidth]{fig_fv_#2_o.jpg}
\includegraphics[width=0.15\textwidth, height=0.15\textwidth]{fig_fv_#2_top3.png} \hspace{0.01\textwidth}
};
\node[image label]{#1};
\end{tikzpicture}
}
\newcommand{\putci}[2]
{
\begin{tikzpicture}
\node[anchor=south east,inner sep=0] at (0,0) {%
\includegraphics[width=0.305\textwidth, height=0.15\textwidth]{fig_fv_#2.png} \hspace{0.01\textwidth}
};
\node[image label]{#1};
\end{tikzpicture}
}
\raggedright
\puti{wrinkled}{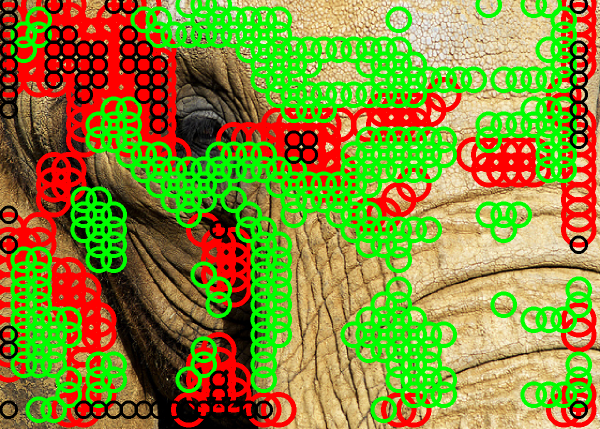}
\puti{wrinkled}{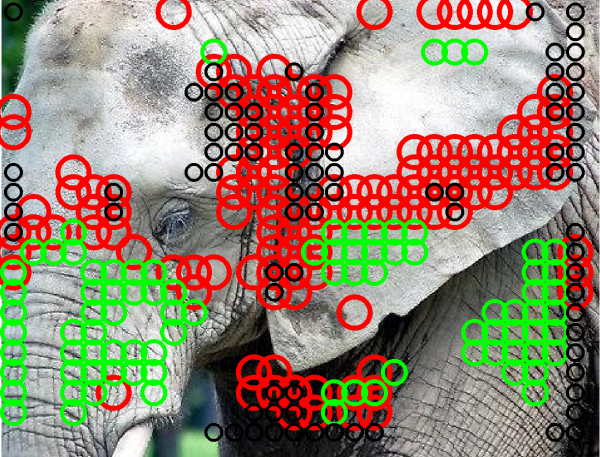}
\puti{studded}{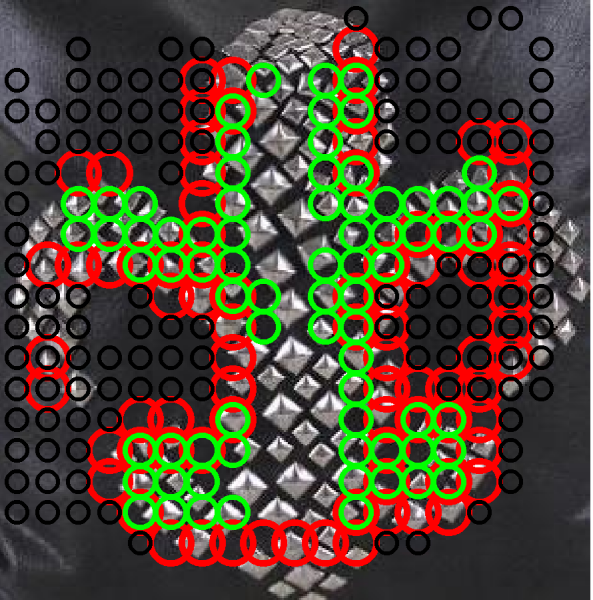}
\puti{studded}{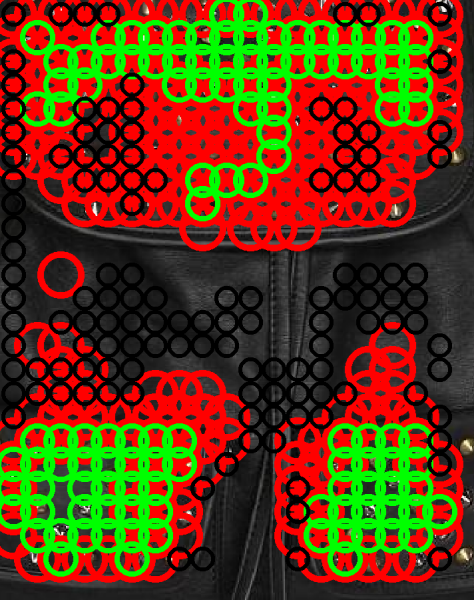}
\puti{swirly}{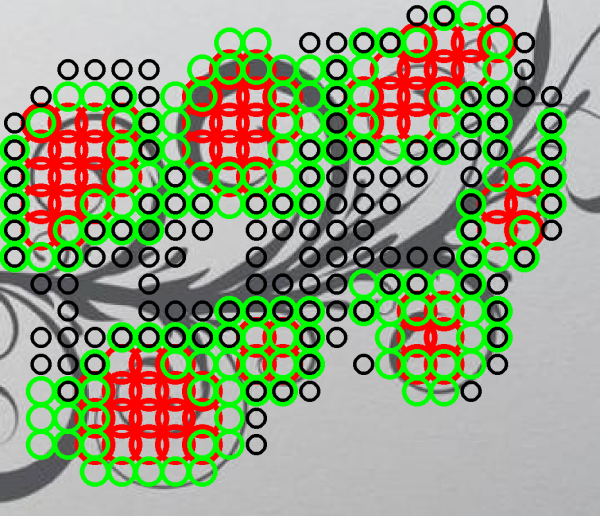}
\puti{swirly}{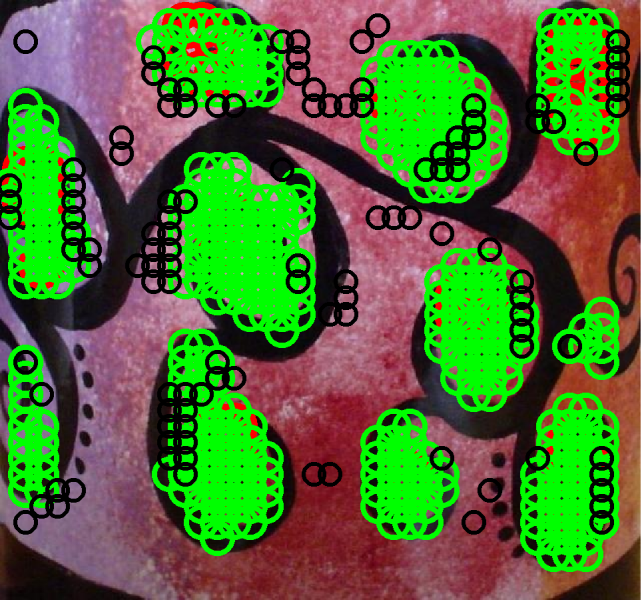}
\puti{bubbly}{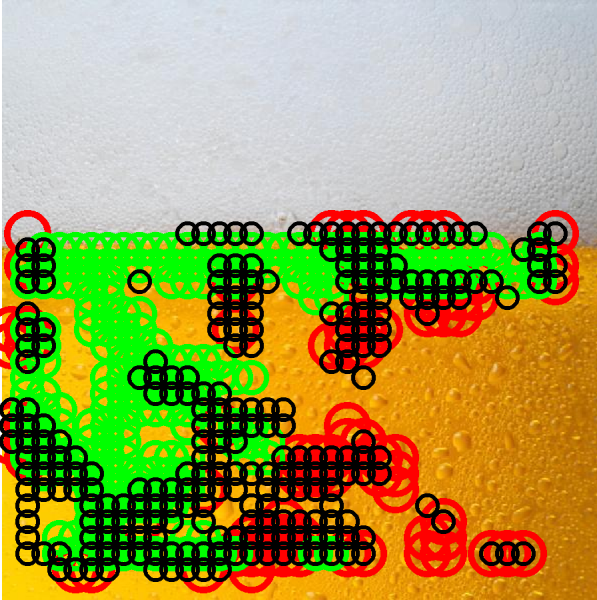}
\puti{bubbly}{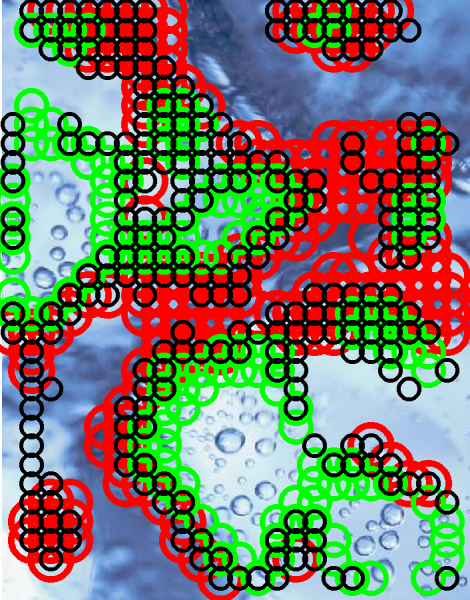}
\puti{sprinkled}{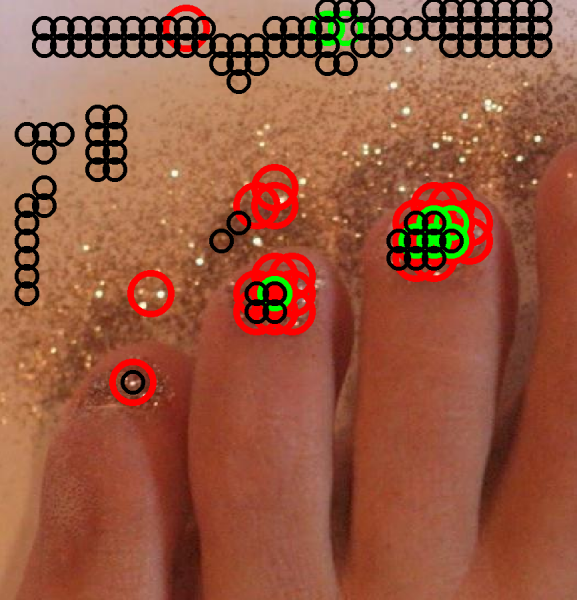}
\hfill \vspace{0.1em}
\putci{swirly / wrinkled}{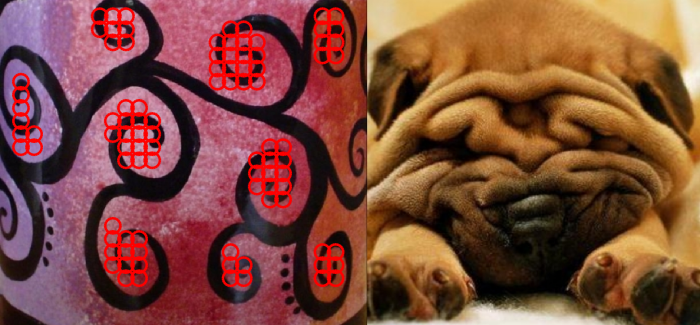}
\putci{wrinkled / swirly}{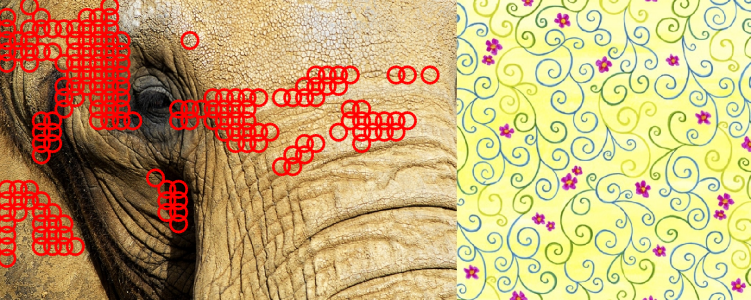}
\putci{studded / bubbly}{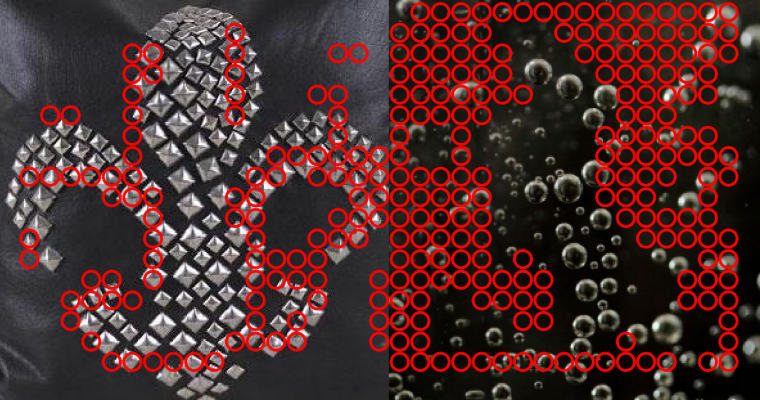}

\caption{\textbf{FV-CNN descriptor visualization.} First three rows: Each image shows the location of the CNN local descriptors that map to the FV-CNN components most strongly associated with the ``wrinkled'', ``studded'',  ``swirly'',  ``bubbly'', and ``sprinkled'' classes for a number of example images in DTD. Red, green and black marks correspond to the top three components selected as described in the text.
Last row: Each image was obtained by combining two images, \eg swirly and wrinkled, and we marked the CNN local descriptors associated with the first class. Swirly descriptors do not fire on the selected wrinkled images. The last pair, studded and bubbly is a harder, as the two images are visually similar, and the descriptors corresponding to \emph{studded} appear on the bubbly image as well. In order to improve visibility, in these images, we show only the most discriminative FV component. }
\label{f:fv-visualize}
\end{figure*}

In this experiment we are interested in understanding which GMM components in the FV-CNN representation code for a particular concept, as well as in determining which areas of the input image contribute the most to the classification score. 

In order to do so, let $\bw$ be the weight vector learned by a SVM classifier for a target class using the FB-CNN representation as input. We  partition $\bw$ in subvectors $\bw_k$, one for each GMM component $k$, and rank components by decreasing value $\|\bw_k\|$, matching the intuition that the GMM component that is predictive  of the target class will result in larger weights. Having identified the top components for a target concept,  the CNN local descriptors are then extracted from a test image, the descriptors that are assigned to a top component are selected, and their location is marked on the image. To simplify the visualization, features are extracted at a single scale.

As can be noted in Fig.~\ref{f:fv-visualize} for some indicative texture types in DTD, the strongest GMM components do tend to fire in correspondence to the characteristic features of each texture. Hence, we conclude that GMM components, while trained in an unsupervised manner, contain clusters that can consistently localize features that capture distinctive characteristics of different texture types.

\subsection{Evaluating texture representations on different domains}\label{s:exp-breadth}

The previous section established optimal combinations of local image descriptors and pooling encoders in texture representations. This section investigates the applicability of these representations to a variety of  domains, from texture (Sect.~\ref{s:exp-texture}) to object and scene recognition (Sect.~\ref{s:exp-places}). It also emphasizes several practical advantages of orderless pooling compared to fully-connected pooling, including helping with the problem of domain shift in learned descriptors. This section focuses on problems where the goal is to either classify an image as a whole or a \emph{known} region of an image, while texture segmentation is looked at later in Sect.~\ref{s:exp-texture-seg}.
 
\subsubsection{Texture recognition}\label{s:exp-texture}

Experiments on textures are divided in recognition in controlled conditions (Sect.~\ref{s:exp-texture-controlled}), where the main sources of variability are viewpoint and illumination, recognition in the wild (Sect.~\ref{s:exp-texture-wild}), characterized by larger intra-class variations, and recognition in the wild and clutter (Sect.~\ref{s:exp-texture-clutter}), where textures are a small portion of a larger scene.

\paragraph{Datasets and evaluation measures.} In addition to the datasets evaluated in Sect.~\ref{s:exp-depth}, DTD, OS+R, FMD and \kthb, we consider here also the standard benchmarks for texture recognition. CUReT~\cite{dana99reflectance} (5612 images, 61 classes), UIUC~\cite{lazebnik05} (1000 images, 25 classes), \kt~\cite{burghouts09} (810 images, 10 classes) are collected in controlled conditions, by photographing the same instance of a material, under varying scale, viewing angle and illumination. UMD~\cite{xu09viewpoint} consists of 1000 images, spread across 25 classes, but  was collected in uncontrolled conditions. For these datasets, we follow the standard evaluation procedures, that is, we are using half of the images for training, and the remaining half for testing, and we are reporting accuracy, averaged over 10 random splits. The ALOT dataset~\cite{burghouts09} is similar to the existing texture datasets, but significantly larger, having 250 categories. For our experiments we used the protocol of  \cite{sulc14fast},  using 20 images per class for training and the rest for testing.

\paragraph{Experimental setup.} For the recognition tasks described in the following subsections, we compare SIFT, VGG-M, and VGG-VD local descriptors and the FC and FV pooling encoders as these were determined before to be some of the best representative texture descriptors. Combinations of such descriptors are evaluated as well.

\paragraph{Texture recognition in controlled conditions.}\label{s:exp-texture-controlled}

This paragraph evaluates texture representations on datasets which are collected under controlled condition (Table~\ref{tbl:deep-net-flavors-soa}, section a). 

For instance recognition, CUReT, UIMD, UIUC are saturated by modern techniques such as~\citep{sifre13rotation,sharma12local,sulc14fast}, with accuracies above $\geq 99\%$. There is little difference between methods, and FV-SIFT, \dcnn, and \rcnn behave similarly. KT is also saturated, although \rcnn looses about (3\%) accuracy compared to \dcnn. 

In material recognition, \ktb and ALOT offer a somewhat more interesting challenge. First, there is a significant difference between \rcnn and \dcnn (3-6\% absolute difference in \ktb and 8-10\% in ALOT), consistent across all CNN evaluated.  Second, CNN descriptors are significantly better than SIFT on \ktb and ALOT with absolute accuracy gains of up to 11\%. 

Compared to the state of the art, FV-SIFT is generally very competitive. In \ktb, FV-SIFT outperforms all recent methods~\cite{cimpoi14describing} with the exception of~\cite{sulc14fast} which is based on a variant of LBP. The latter is very strong in ALOT too, but in this case FV-SIFT is virtually as good. In the case of \ktb, \cite{sulc14fast} is  better than most of the deep descriptors as well, but it is still significantly bested by FV-VGG-VD (+5.5\%). Nevertheless, this is an example in which a specialized texture descriptor can be competitive with deep features, although of course deep features apply unchanged to several other problems.

On ALOT, \dcnn with VGG-VD is on par with the result obtained by \cite{badri14fast} -- 98.45\% -- but their model was trained with 30 images per class instead of 20. The same paper reports even better results, but when training with 50 images per class or by integrating additional synthetic training data. 

\paragraph{Texture recognition in the wild.}\label{s:exp-texture-wild}
%
This paragraph evaluates the texture representations on two texture datasets collected ``in the wild'': FMD (materials) and DTD (describable attributes).

Texture recognition in the wild is more comparable, in term of the type of intra-class variations, to object recognition than to texture recognition in controlled conditions. Hence, one can expect larger gains in moving from texture-specific descriptors to general-purpose descriptors. This is confirmed by the results. SIFT is competitive with AlexNet and VGG-M features in FMD (within 3\% accuracy), but it is significantly worse in DTD (+4.3\% for FV-AlexNet and +8.2\% for FV-VGG-M). \dcnn is a little better than \rcnn ($\sim$3\%) on FMD and substantially better in DTD ($\sim$8\%). Different CNN architectures exhibit very different performance; moving from AlexNet to VGG-VD, the accuracy absolute improvement is more than 11\% across the board.

Compared to the state of the art, FV-SIFT is generally very competitive, outperforming the specialized texture descriptors developed by~\cite{qi14pairwise,sharan13recognizing} in FMD (and this without using ground-truth texture segmentations as used by~\cite{sharan13recognizing}). Yet FV-VGG-VD is significantly better than all these descriptors (+24.7\%).
 
In term of complementarity of the features, the combination of \rcnn and \dcnn improves performance by about 3\% across the board, but including FV-SIFT (labelled FV-SIFT/FC+FV-VD in the table) as well does not seem to improve performance further. This is in contrast with the fact that SIFT was found to be fairly complementary to \rcnn on a variant of AlexNet in~\cite{cimpoi14describing}.

\paragraph{Texture recognition in clutter.}\label{s:exp-texture-clutter}
%
This section evaluates texture representations on recognizing texture materials and describable attributes in clutter. Since there is no standard benchmark for this setting, we introduce here the first analysis of this kind using the the OS+R and OSA+R datasets of Sect.~\ref{s:os-benchmark}. Recall that the +R suffix indicates that, while textures are imaged in clutter, the classifier is given the ground-truth region segmentation; therefore, the goal of this experiment is to evaluate the effect of realistic viewing conditions on texture recognition, but the problem of segmenting the textures is evaluated later, in Sect.~\ref{s:exp-texture-seg}.

Results are reported in Table~\ref{tbl:deep-net-flavors-soa} in sections b and c. As before, performance improves with the depth of CNNs. For example, in material recognition (OS+R) accuracy starts at about $39.1\%$ for FV-SIFT, is about the same for FC-VGG-M ($41.3\%$) and a little better for FC-VGG-VD ($43.4\%$). However, the benefit of switching from FC encoding to FV encoding is now even more dramatic. For example, on OS+R FV-VGG-M has accuracy $52.5\%$ ($+11.2\%$) while FV-VGG-VD $59.5\%$ ($+16.1\%$). This clearly demonstrates the advantage of orderless pooling of CNN local descriptors on FC pooling when regions of different sizes and shapes must be evaluated. There is also a significant computational advantage (evaluated further in Sect.~\ref{s:exp-places}) if, as it is typical, several regions must be classified: in that case, CNN features need not to be recomputed for each region.  Results on OSA+R are entirely analogous.

\subsubsection{Object and scene recognition}\label{s:exp-places}

\begin{table}[t]
\centering
\begin{tabular}{|r|ccc|}
\hline
 & \multicolumn{3}{c|}{Accuracy (\%)}\\
CNN  & \rcnn & \dcnn & FC+\dcnn \\
\hline
     PLACES       & 65.0 &  67.6 &   73.1 \\
     CAFFE        & 58.6 &  69.7 &   71.6 \\
\hline
     VGG-M        & 62.5 &  74.2 &   74.4 \\
     VGG-VD       & 67.6 &  81.0 &   80.3 \\
\hline
\end{tabular}
\caption{\textbf{Accuracy of various CNNs on the MIT indoor dataset.} PLACES and CAFFE are the same CNN architecture (``AlexNet") but trained on different datasets (PLACES and ImageNet resp.). The domain specific advantage of training on PLACES dissapears when the convolutional features are used with FV pooling. For all architectures FV CNN outperformns FC and better architectures lead to better overall performance.}\label{t:places-cnn}
\end{table}

This section evaluates texture descriptors on tasks other than texture recognition, namely coarse and fine-grained object categorization, scene recognition, and semantic region recognition.

\paragraph{Datasets and evaluation measures.} In addition to the datasets seen before, here we experiment with fine grained recognition in the CUB~\cite{WahCUB_200_2011} dataset. This dataset contains 11788 images, representing 200 species of birds. The images are split approximately into half for training and half for testing, according to the list that accompanies the dataset.  Image representations are either applied to the whole image (denoted CUB) or on the region counting the target bird using ground-truth bounding boxes (CUB+R). Performance in CUB and CUB+R is reported as per-image classification accuracy. For this dataset, the local descriptors are again extracted at multiple scales, but now only for the smaller range $\{0.5, 0.75, 1\}$ which was found to work better for this task.

Performance is also evaluated on the MSRC dataset, designed to benchmark semantic segmentation algorithms. The dataset contains 591 images, for which some pixels are labelled with one of the 23 classes. In order to be consistent with the results reported in the literature, performance is reported in term of per-pixel classification accuracy, similar to the measure used for the OS task as defined in Sect.~\ref{s:os-benchmark}. However, this measure is further modified such that it is \emph{not} normalized per class:
\begin{equation}\label{e:acc-msrc}
\operatorname{acc-msrc}(\hat c)
=
\frac{|\{ \bp : c(\bp) = \hat c(\bp)\}|}{|\{ \bp : c(\bp) \not= 0\}|}.
\end{equation}

\paragraph{Analysis.}
Results are reported in Table~\ref{tbl:deep-net-flavors-soa} section d. On PASCAL VOC, MIT Indoor, CUB, and CUB+R the relative performance of the different descriptors is similar to what has been observed above for textures. Compared to the state-of-the-art results in each dataset, \rcnn and particularly the \dcnn descriptors are very competitive. The best result obtained in PASCAL VOC is comparable to the current state-of-the-art set by the deep learning method of~\cite{wei14cnn:} ($85.2\%$ vs $84.9\%$ mAP), but using a much more straightforward pipeline.  In MIT Places the best performance is also substantially superior ($+10\%$) to the current state-of-the-art using deep convolutional networks learned on the MIT Place dataset~\cite{zhou14learning} (this is discussed further below). In the CUB dataset, the best performance is  short ($\sim 6\%$) of the state-of-the-art results of~\cite{zhang14part-based}. However, \cite{zhang14part-based} uses a category-specific part detector and corresponding part descriptor as well as a CNN fine-tuned on the CUB data; by contrast, \dcnn and \rcnn are used here as \emph{global image descriptors} which, furthermore, \emph{are the same for all the datasets considered}. Compared to the results of \cite{zhang14part-based} without part-based descriptors (but still using a part-based object detector), the best of our global image descriptors perform substantially better ($62.1\%$ vs $67.3\%$).

Results on MSRC+R for semantic segmentation are entirely analogous; it is worth noting that, although ground-truth segments are used in this experiment and hence this number is not comparable with other reported in the literature, the best model achieves an outstanding $99.1\%$ per-pixel classification rate in this dataset.

\paragraph{Conclusions.}
The conclusion of this section is that \dcnn, although inspired by texture representations, are superior to many alternative descriptors in object and scene recognition, including more elaborate constructions. Furthermore, \dcnn is significantly superior to \rcnn in this case as well.

\subsubsection{Domain transfer}\label{s:exp-places}

This section investigates in more detail the problem of domain transfer in CNN-based features. So far, the same underlying CNN features, trained on the ImageNet's ILSVCR data, were used in all cases. To investigate the effect of the source domain on performance, this section consider, in addition to these networks, new ones trained on the PLACES dataset~\cite{zhou14learning} to recognize scenes on a dataset of about 2.5 million labeled images. \cite{zhou14learning} showed that, applied to the task of scene recognition in MIT Indoor, these features outperform similar ones trained on ILSVCR (denoted CAFFE~\cite{jia13caffe} below) -- a fact explained by the similarity of domains. We repeat this experiment using FC- and \dcnn descriptors on top of VGG-M, VGG-VD, PLACES, and CAFFE.

Results are shown in Table~\ref{t:places-cnn}. The \rcnn performance is in line with those reported in~\cite{zhou14learning} -- in scene recognition with \rcnn the same CNN architecture performs better if trained on the Places dataset instead of the ImageNet data ($58.6\%$ vs $65.0\%$ accuracy\footnote{\cite{zhou14learning} report $68.3\%$ for PLACES applied to MIT Indoor, a small difference explained by implementation details such as the fact that, for all the methods, we do not perform data augmentation by jittering.}). Nevertheless, stronger CNN architectures such as VGG-M and VGG-VD can approach and outperform PLACES even if trained on ImageNet data ($65.0\%$ vs $62.5\%/67.6\%$).

However, when it comes to using the filter banks with \dcnn, conclusions are very different. First, \dcnn outperforms \rcnn in all cases, with substantial gains up to $\sim 11-12\%$ in correspondence of a domain transfer from ImageNet to MIT Indoor.  The gap between \rcnn and \dcnn is the highest for VGG-VD models (67.6\% vs 81.0\%, nearly 14\% difference), a trend also exhibited by other datasets as seen in Table~\ref{tbl:deep-net-flavors-soa}. Second, \emph{the advantage of using domain-specific CNNs disappears}. In fact, the same CAFFE model that is 6.4\% worse than PLACES with \rcnn, is actually 2.1\% \emph{better} when used in \dcnn. The conclusion is that \dcnn appears to be immune, or at least substantially less sensitive, to domain shifts.  

Our explanation of this  phenomenon is that the convolutional features are substantially less committed to a specific dataset than the fully connected layers. Hence, by using those, \dcnn  tends to be a lot more general than \rcnn. A second explanation is that PLACES CNN may learn filters that tend to capture the overall spatial structure of the image, whereas CNNs trained on ImageNet tend to focus on localized attributes which may work well with orderless pooling.

Finally, we compare \dcnn to alternative CNN pooling techniques in the literature. The closest method is the one of~\cite{gong14multi-scale}, which uses a similar underlying CNN to extract local image descriptors and VLAD instead of FV for pooling. Notably, however, \dcnn results on MIT Indoor are markedly better than theirs for both VGG-M and VGG-VD (68.8\% vs 74.2\% / 81.0\% resp.) and marginally better (69.7\% -- Table~\ref{tbl:deep-net-flavors-soa} and~\ref{t:places-cnn}) when the same CAFFE CNN is used. Also, when using VLAD instead of FV for pooling the convolutional layer descriptors, the performance of our method is still better (68.8\% vs 71.2\%), as seen in Table~\ref{tbl:encodings-sift-vs-cnn-rcnn}. The key difference is that \dcnn pools convolutional features, whereas \cite{gong14multi-scale} pools fully connected descriptors extracted from square image patches. Thus, even without spatial information as used by~\cite{gong14multi-scale}, \dcnn is not only substantially faster -- 8.5$\times$ speedup when using the same network and three scales, but at least as accurate.

\section{Experiments on semantic segmentation}\label{s:tex-seg}

The previous sections considered the problem of  recognizing given image regions. This section explores instead the problem of automatically recognizing as well as segmenting such regions in the image.

\subsection{Experimental setup}

Inspired by Cimpoi~\etal~\cite{cimpoi14describing} that successfully ported object description methods to texture descriptors, here we propose a segmentation technique building on ideas from object detection. An increasingly popular method for object detection, followed for example by \rcnn~\cite{girshick14rich}, is to first propose a number of candidate object regions using low-level image cues, and then verifying a shortlist of such regions using a powerful classifier. Applied to textures, this requires a low-level mechanism to generate textured region proposals, followed by a region classifier. A key advantage of this approach is that it allows applying object- (\rcnn) and texture-like (\dcnn) descriptors alike. After proposal classification, each pixel can be assigned more than one label; this is solved with a simple voting schemes, also inspired by object detections methods.


The paper explores two such region generation methods: the \emph{crisp regions} of~\cite{isola14crisp} and the \emph{Multi-scale Combinatorial Grouping} (MCG) of~\cite{arbelaez2014multiscale}. In both cases, region proposals are generated using low-level image cues, such as color or texture consistency, as specified by the original methods. It would of course be possible to incorporate \rcnn and \dcnn among these energy terms to potentially strengthen the region generation mechanism itself. However, this contradicts partially  the logic of the scheme, which breaks down the problem into cheaply generating tentative segmentations and then verifying them using a more powerful (and likely expensive) model. Furthermore, and more importantly, these cues focus on separating texture \emph{instances}, as presented in each particular image, whereas \rcnn and \dcnn are meant to identify a texture class. It is reasonable to expect instance-specific cues (say the color of a painted wall) to be better for segmentation.

The crisp region method generates a single partition of the image; hence, individual pixels are labelled by transferring the label of the corresponding region, as determined by the learned predictor. By contrast,  MCG generates many thousands overlapping region proposals in an image and requires a mechanism to resolve potentially ambiguous pixel labelings. This is done using the following simple scheme. For each proposed region, its label is set to the the highest scoring class based on the multi-class SVM, and its score to the corresponding class score divided by the region area. Proposals are then sorted by increasing score and ``pasted'' to the image sequentially. This has the effect of considering larger regions before smaller ones and more confident regions after less confident ones for regions of the same area. 

\newcommand{\mycomment}{}
\subsection{Dense-CRF post-processing}
The  segmentation results delivered by the previous methods can potentially be hampered by the occasional failures of the respective front-end superpixel segmentation modules. But  we can see the front-end segmentation as providing as a convenient way of pooling discriminative information, which can then be refined post-hoc through a pixel-level segmentation algorithm.

In particular, a series of recent works \cite{ChenPKMY14,bell14material,ZhengJRVSDHT15} have reported that substantial gains can be obtained by combining CNN classification scores with  the densely-connected
Conditional Random Field (Dense-CRF) of \cite{krahenbuhl2011efficient}.
Apart from its ability to incorporate information pertaining to image boundaries and color similarity, the Dense-CRF is particularily effiecient when used in conjunction with approximate probabilistic
inference: 
the message passing updates under a
fully decomposable mean field approximation can be expressed as convolutions with a Gaussian kernel in feature
space, implemented efficiently using  high-dimensional filtering  \citep{adams2010fast}.

Inspired by these advances, we have  employed the Dense-CRF segmentation algorithm {\em post-hoc},  
with the aim of enhancing our algorithm's ability to localize region boundaries by taking context and low-level image information into account. For this we turn the superpixel classification scores into pixel-level unary terms, interpeting the SVM classifier's scores as  indicating the negative energy associated to labelling each pixel with the respective labels. Even though Platt scaling could be used to turn the SVM scores into log-probability estimates, we prefer to estimate the transformation by jointly cross-validating the SVM-Dense-CRF cascade's parameters. 
In particular, similarly to \cite{krahenbuhl2011efficient,ChenPKMY14}, we
set the dense CRF hyperparameters by cross-validation, performing grid
search to find the values that perform best on a validation set.

\subsection{Analysis}\label{s:exp-texture-seg}

\begin{table*}
\centering
\begin{tabular}{|cc|ccc|cccc|c|}
\hline
& & \multicolumn{3}{c|}{VGG-M}  & \multicolumn{4}{c|}{VGG-VD}   & \\
dataset  & measure (\%) & \rcnn & \dcnn       & FV+FC-CNN & \rcnn & \dcnn       & FC+FV-CNN & CRF & SoA                       \\
\hline
       OS &    pp-acc & 36.0 & 48.6 (46.9) & 49.8 &  38.5 &  55.5 (55.7) &   55.9 & 56.5 &   -- \\
       OSA &     acc-osa~\eqref{e:acc-osa} & 42.8 & 66.0 & 63.4 &  42.1 &  67.9 &   64.6   & 68.9 &                     --\\
     MSRC &   acc-msrc~\eqref{e:acc-msrc} & 56.1 & 82.3        & 75.2 &  57.7 &  86.9        &   81.5 & 90.2 & 86.5~\cite{ladicky10graph} \\
\hline
\end{tabular}
\caption{Segmentation and recognition using crisp region proposals of materials (OS) and things \& stuff (MSRC). Per-pixel accuracies are reported, using the MSRC variant (see text) for the MSRC dataset. Results using MCG proposals~\cite{arbelaez2014multiscale} are reported in brackets for FV-CNN.} \label{f:seg-table}
\end{table*}
\begin{figure*}
\newcommand{\inc}[2]{%
\includegraphics[width=0.166\textwidth,frame]{fig_os_crisp_result-#1-seg-figures_#2}}
\includegraphics[width=\textwidth]{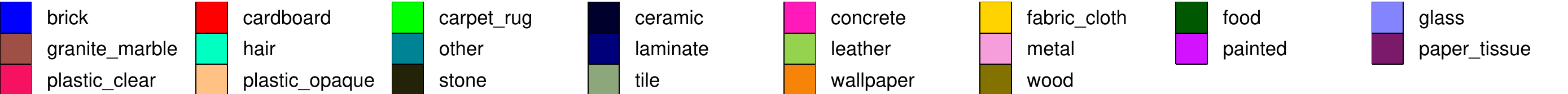}
\newcommand{\xinc}[1]{%
\inc{rcnnvd}{#1}%
\inc{rcnnvd}{#1-rcnnvd-gt}%
\inc{rcnnvd}{#1-rcnnvd-pred}%
\inc{rcnnvd}{#1-rcnnvd-err}%
\inc{dcnnvd}{#1-dcnnvd-pred}%
\inc{dcnnvd}{#1-dcnnvd-err}}\\
\null\hfill(a)\hfill\hfill(b)\hfill\hfill(c)%
\hfill\hfill(d)\hfill\hfill(e)\hfill\hfill(f)\hfill\null\\
\xinc{009626}
\xinc{008294}
\xinc{116511}
\xinc{112263}
\caption{{\bf OS material recognition results.} Example test image with material recognition and segmentation on the OS dataset. (a) original image. (b) ground truth segmentations from the OpenSurfaces repository (note that not all pixels are annotated). (c) \rcnn and crisp-region proposals segmentation results. (d) correctly (green) and incorrectly (red) predicted pixels (restricted to the ones annotated). (e-f) the same, but for \dcnn.}\label{f:os-examples}
\end{figure*}
\begin{figure*}
\newcommand{\inc}[2]{%
\includegraphics[width=0.1249\textwidth,frame]{fig_msrc_crisp_result-#1-seg-figures_#2}}
\includegraphics[width=\textwidth]{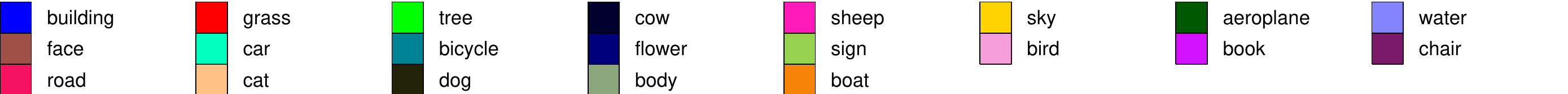}
\newcommand{\xinc}[1]{%
\inc{rcnnvd}{#1}%
\inc{rcnnvd}{#1-rcnnvd-gt}%
\inc{rcnnvd}{#1-rcnnvd-pred}%
\inc{rcnnvd}{#1-rcnnvd-err}%
\inc{dcnnvd}{#1-dcnnvd-pred}%
\inc{dcnnvd}{#1-dcnnvd-err}%
\inc{crf}{#1-crf-pred}%
\inc{crf}{#1-crf-err}}
\null\hfill(a)\hfill\hfill(b)\hfill\hfill(c)%
\hfill\hfill(d)\hfill\hfill(e)\hfill\hfill(f)%
\hfill\hfill(g)\hfill\hfill(h)\hfill\null\\
\xinc{18_29_s}
\xinc{4_20_s}
\xinc{7_10_s}
\xinc{8_19_s}
\caption{{\bf MSRC object segmentation results.} (a) image, (b) ground-truth, (c-d) \rcnn segmentation and errors, (e-f) \dcnn segmentation and errors (in red),  (g-h) segmentation and errors after Dense CRF post-processing.}
\label{f:msrc-examples}
\end{figure*}


Results are reported in Table~\ref{f:seg-table}. Two datasets are evaluated: OS for material recognition and MSRC for things \& stuff. Compared to OS+R, classifying crisp regions results in a drop of about $10\%$ per-pixel classification accuracy for all descriptors. 
 At the same time, it shows that there is ample space for future improvements. In MSRC, the best accuracy is $87.0\%$, just a hair above the best published result $86.5\%$~\cite{ladicky10graph}. Remarkably, these algorithms do not use any dataset-specific training, nor CRF-regularised semantic inference: they simply greedily classify regions as obtained from a general-purpose segmentation algorithms. CRF post-processing improves the results even further, up to $90.2\%$ in MSRC. Qualitative segmentation results (sampled at random) are given in Fig.~\ref{f:os-examples} and~\ref{f:msrc-examples}.

Results using \dcnn shown in Table~\ref{f:seg-table} in brackets (due to the requirement of computing CNN features from scratch for every region, it was impractical to use \rcnn with MCG proposals). The results are comparable to those using crisp regions, resulting in $55.7\%$ accuracy on the OS dataset. Other schemes such as non-maximum suppression of overlapping regions that are quite successful for object segmentation~\cite{hariharan2014simultaneous} performed rather poorly in this case. This is probably because, unlike objects, texture information is fairly localized and highly irregularly shaped in an image.

While for recognizing textures, materials or objects covering the entire image, the advantage in computational cost of \dcnn on \rcnn and was not significant, the latter consisting in evaluating few layers less, the advantage of \dcnn becomes clear for segmentation tasks, as \rcnn requires recomputing the features for every region proposal.

\section{Applications of describable texture attributes}\label{s:exp-dtd-attr}

This section explores two applications of the DTD attributes: using them as general-purpose texture descriptors (Sect.~\ref{s:exp-dtd-attr}) and as a tool for search and visualization (Sect.\ref{s:exp-visualize}).

\subsection{Describable attributes as generic texture descriptors}\label{s:exp-dtd-attr}

\begin{table*}[t!]
\begin{center}
\small
\begin{tabular}{|l|c|c|c|c|}
\hline
  DTD Classifier     & \multicolumn{2}{c|}{\ktb} & \multicolumn{2}{c|}{FMD}    \\
   Method                &     Linear   &   RBF        &   Linear     &   RBF        \\                       
  \hline
\IFVSIFT            & 64.74\pmt{2.36} & 67.75\pmt{2.89} & 49.24\pmt{1.73} & 52.53\pmt{1.26}  \\ 
\dcnn               & 67.39\pmt{3.75} & 67.66\pmt{3.30} & 62.81\pmt{1.33} & 64.69\pmt{1.41}  \\ 
\dcnn-VD            & 74.59\pmt{2.45} & 74.71\pmt{1.96} & 70.81\pmt{1.39} & 73.09\pmt{1.35}  \\ 
\IFVSIFT + \rcnn    & 73.98\pmt{1.24} & 74.53\pmt{1.14} & 64.20\pmt{1.65} & 67.13\pmt{1.95}  \\ 
\IFVSIFT + \rcnn-VD & \bf{74.52\pmt{2.31}} & \bf{77.14\pmt{1.36}} & 69.21\pmt{1.77} & 72.17\pmt{1.66}  \\ 
 \hline
 Previous best      &  \multicolumn{2}{c|}{76.0~\pmt{2.9}\cite{sulc14fast}} & \multicolumn{2}{c|}{57.7\pmt{1.7}~\cite{sharan13recognizing,qi14pairwise}} \\
 \hline
\end{tabular}
\end{center}
\caption{{\bf DTD for material recognition.} Accuracy on material recognition on the  \kthb and FMD benchmarks obtained by using as image representation the predictions of the 47 DTD attributes by different methods: FV-SIFT, FV-CNN (using either VGG-M or VGG-VD) or combinations.  Accuracies are compared to published state of the art results.}
\label{tbl:dtd-descr-results}
\end{table*}

This section explores using the 47 describable attributes of Sect.~\ref{s:dtd} as a general-purpose texture descriptor. The first step in this construction is to learn a multi-class predictor for the 47 attributes; this predictor is trained on DTD using a texture representation of choice and a multi-class linear SVM as before. The second step is to evaluate the multi-class predictor to obtain a 47-dimensional descriptor (of class scores) for each image in a target dataset. In this manner, one obtains a novel and very compact representation which is then used to learn a multi-class non-linear SVM classifier, for example for material recognition.

Results are reported in Table~\ref{tbl:dtd-descr-results} for material recognition in FMD and \kthb. There are two important factors in this experiment. The first one is the choice of the DTD attributes predictor. Here the best texture representations found before are evaluated: FV-SIFT, \rcnn, and \dcnn (using either VGG-M or VGG-VD local descriptors), as well as their combinations. The second one is the choice of classifier used to predict a texture material based on the 47-dimensional vector of describable attributes. This is done using either a linear or RBF SVM.

Using a linear SVM and \IFVSIFT to predict the DTD attributes yields promising results: 64.7\% classification accuracy on \ktb and 49.2\% on FMD. The latter outperforms the specialized aLDA model of~\citep{sharan13recognizing} combining color, SIFT and edge-slice features, whose accuracy is 44.6\%.  Replacing SIFT with CNN image descriptors (\dcnn) improves results significantly for FMD (49.2\% vs 62.8\% for VGG-M and 70.8\% for VGG-VD) as well as \ktb (64.7\% vs 67.4\% and 74.6\% respectively). While these results are not as good as using the best texture representations directly on these datasets, remarkably the dimensionality of the DTD descriptors is \emph{two orders of magnitude smaller} than all the other alternatives. 

An advantage of the small dimensionality of the DTD descriptors is that using an RBF classifier instead of the linear one is relatively cheap. Doing so improves the performance by 1-3\% on both FMD and \ktb across experiments. Overall, the best result of the DTD features on \kthb is 77.1\% accuracy, slightly better than the state-of-the-art accuracy rate of 76.0\% of~\cite{sulc14fast}. On FMD the DTD features outperform significantly the state of the art~\cite{}: 72.17\% accuracy vs.\ 57.7\%, an improvement of about 15\%.

The final experiment compares the semantic attributes of~\cite{matthews13enriching} on the Outex data. Using \IFVSIFT and a linear classifier to predict the DTD attributes,  performance on the retrieval experiment of~\cite{matthews13enriching} is 49.82\% mAP which is not competitive with their result of 63.3\% obtained using $\text{LBP}^u$ (Sect.~\ref{s:texture-local-descr}). To verify whether this was due to $\text{LBP}^u$ being particularly optimized for the Outex data,  the DTD attributes where trained again using FV on top of the $\text{LBP}^u$ local image descriptors; by doing so, using the 47 attributes on Outex results in an accuracy of 64.5\% mAP; at the same time, Table~\ref{tbl:res-local-feats} shows that $\text{LBP}^u$ is not a competitive predictor on DTD itself. This confirms the advantage of the $\text{LBP}^u$ on the Outex dataset.


\subsection{Search and visualization}\label{s:exp-visualize}

\renewcommand{\dim}{.09\textwidth}
\setlength{\tabcolsep}{0.3pt}
\newcommand{\incf}[2]{\footnotesize #1\vphantom{y} \includegraphics[width=\dim]{#2}}
\begin{figure*}[!t]
\begin{tabular}{
>{\raggedright\centering}p{\dim}
>{\raggedright\centering}p{\dim}
>{\raggedright\centering}p{\dim}
>{\raggedright\centering}p{\dim}
>{\raggedright\centering}p{\dim}
>{\raggedright\centering}p{\dim}
>{\raggedright\centering}p{\dim}
>{\raggedright\centering}p{\dim}
>{\raggedright\centering}p{\dim}
>{\raggedright\centering}p{\dim}
>{\raggedright\centering}p{\dim}}
\incf{aluminum}{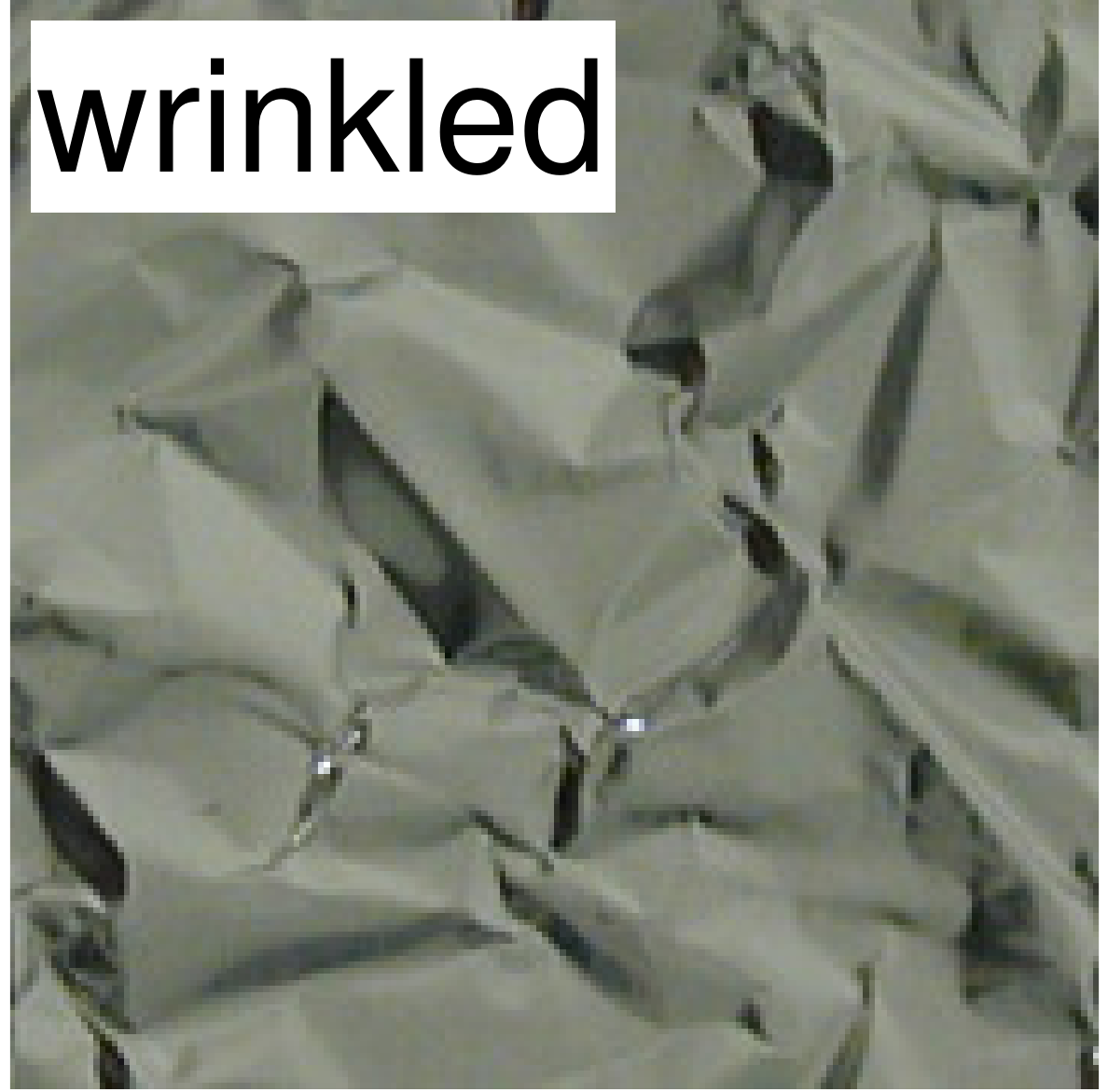} &
\incf{brown bread}{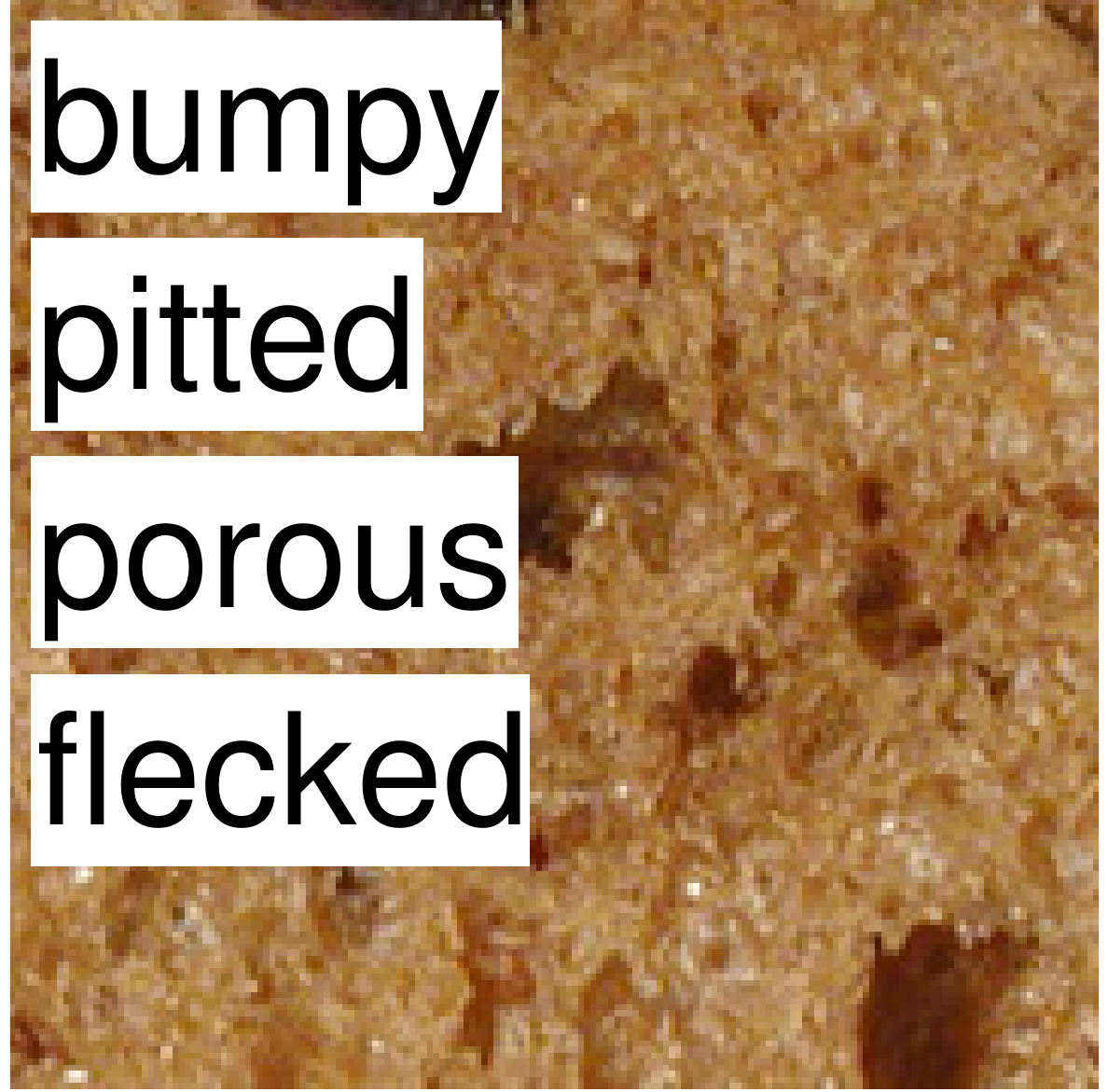} &
\incf{corduroy}{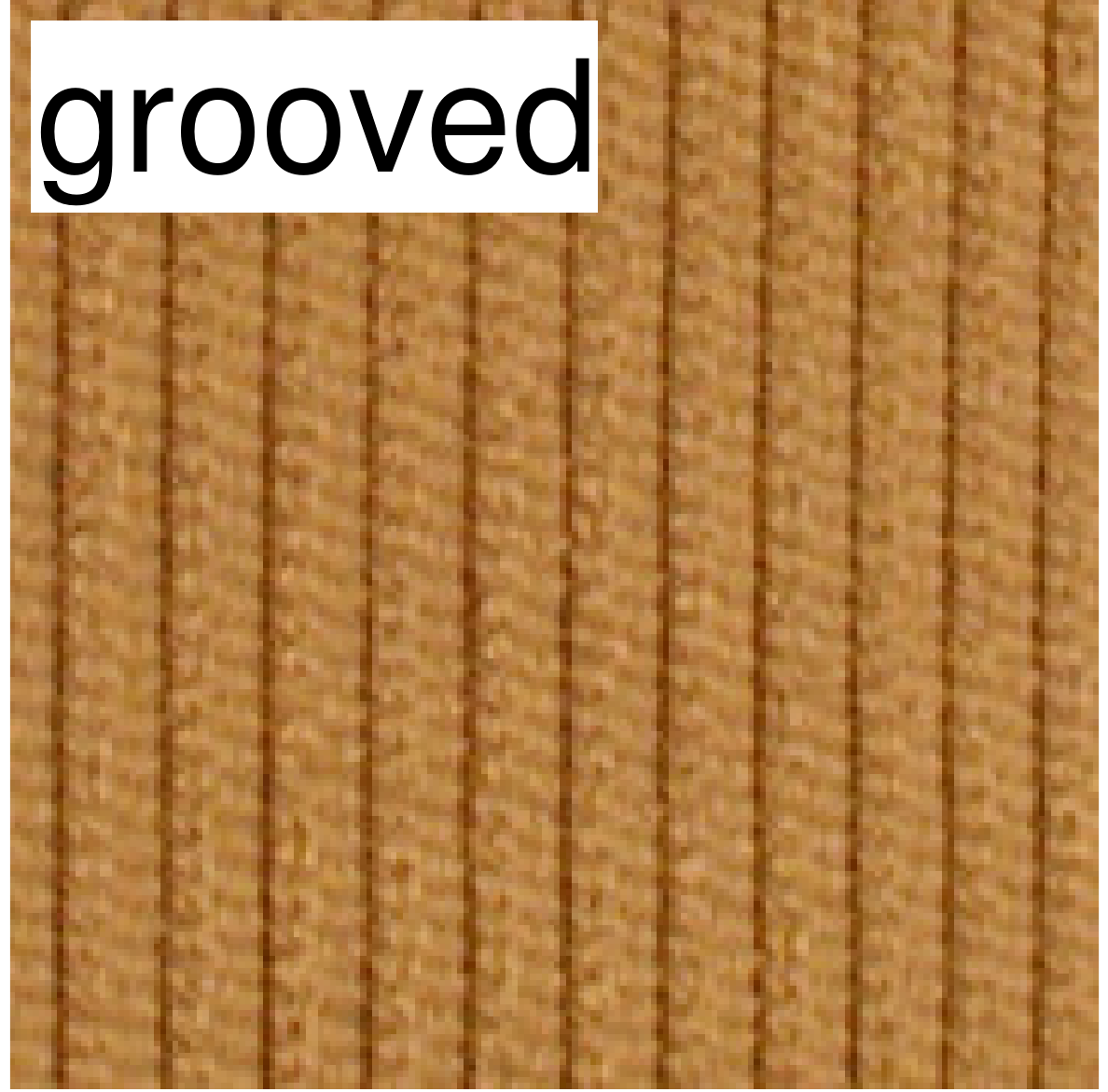} &
\incf{cork}{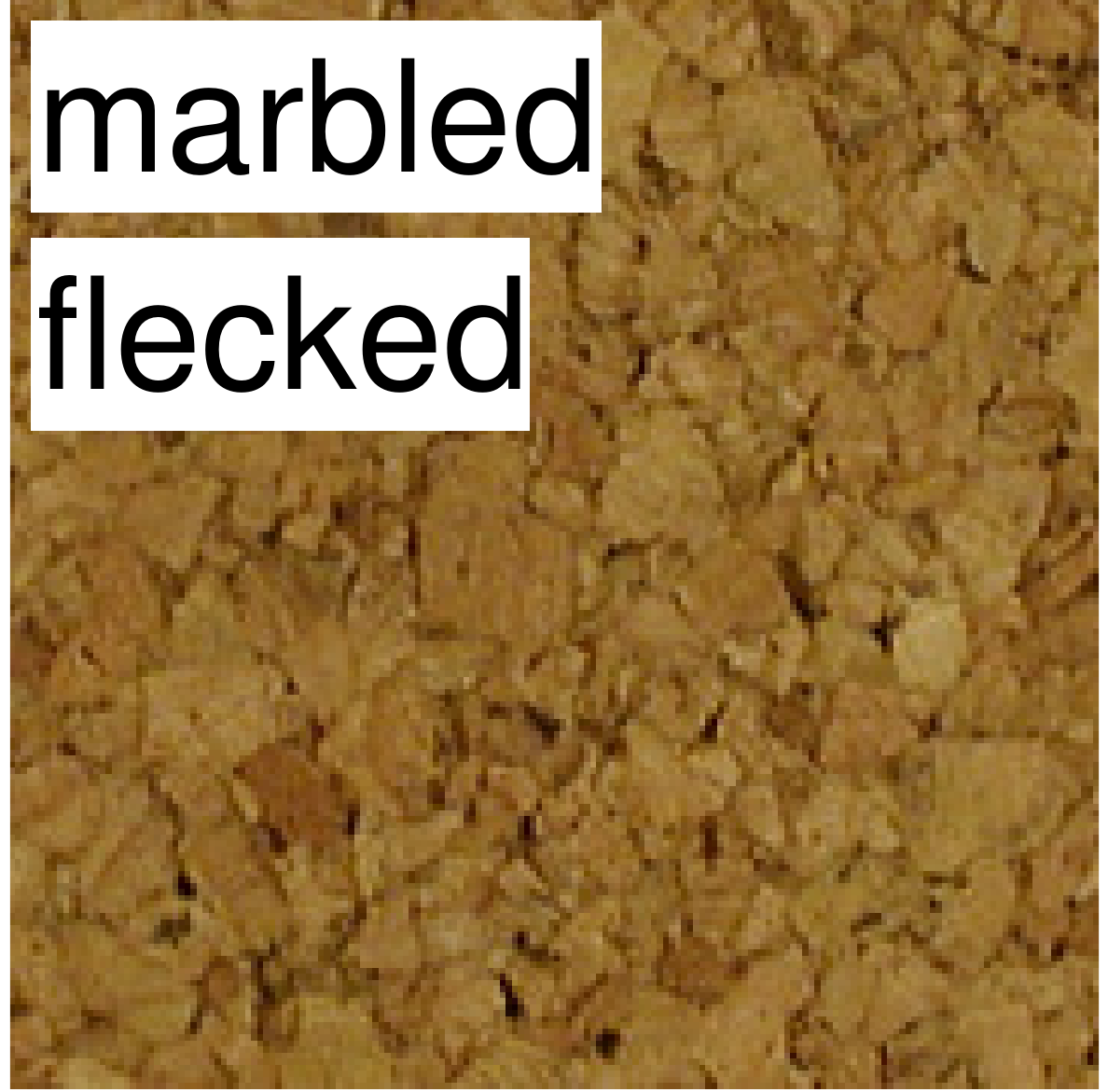} &
\incf{cotton}{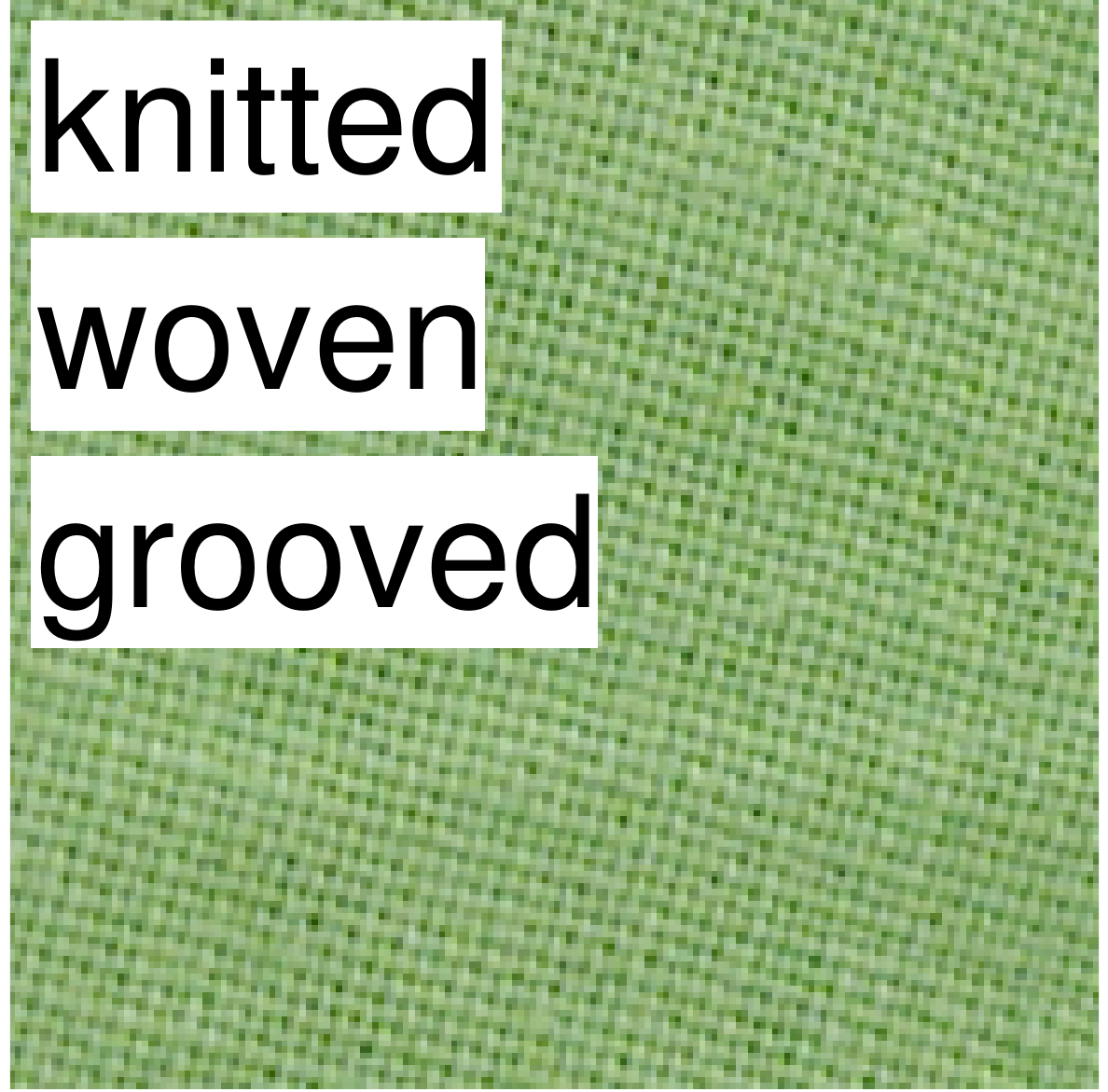}&
\incf{cracker}{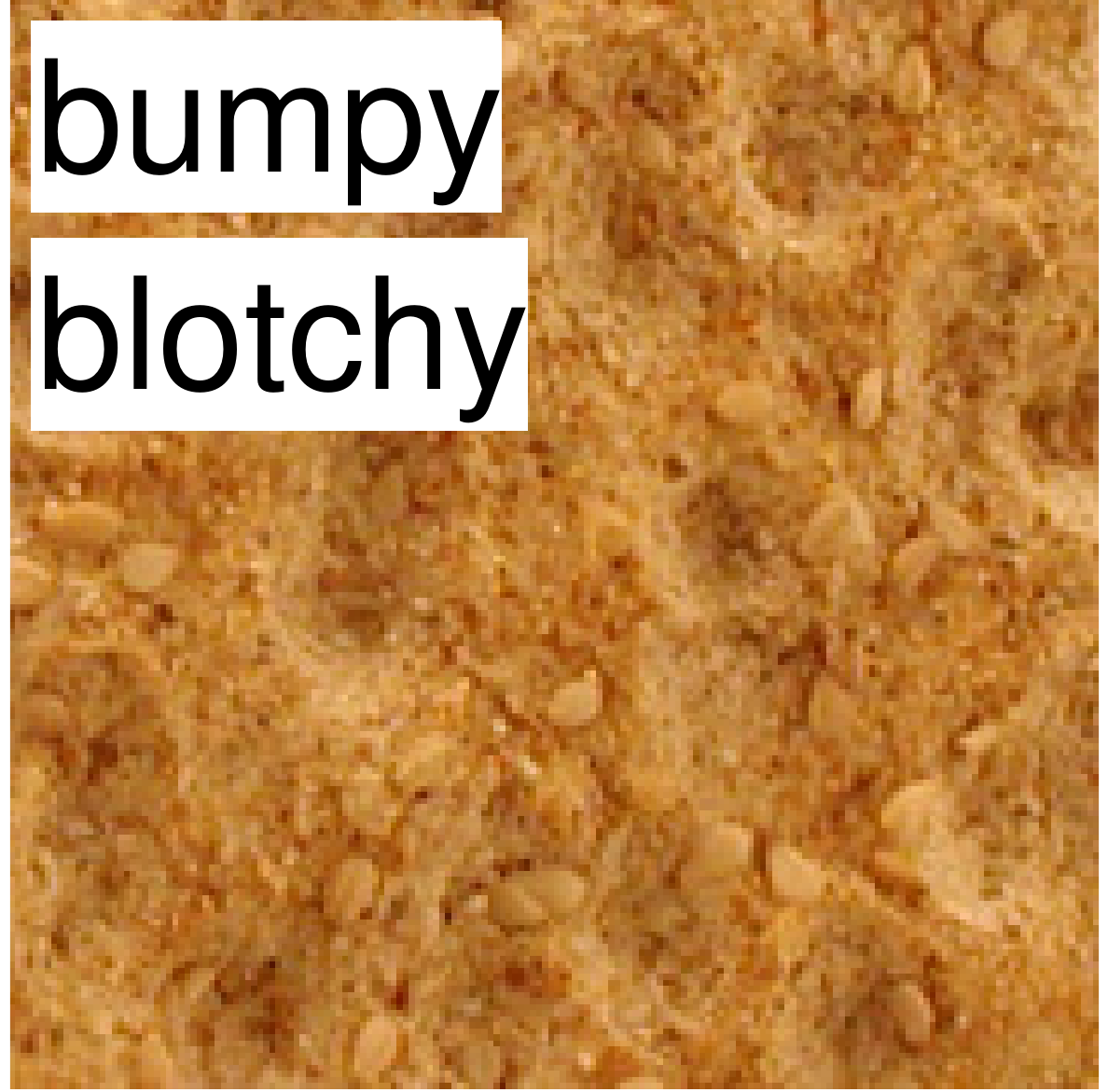}&
\incf{lettuce leaf}{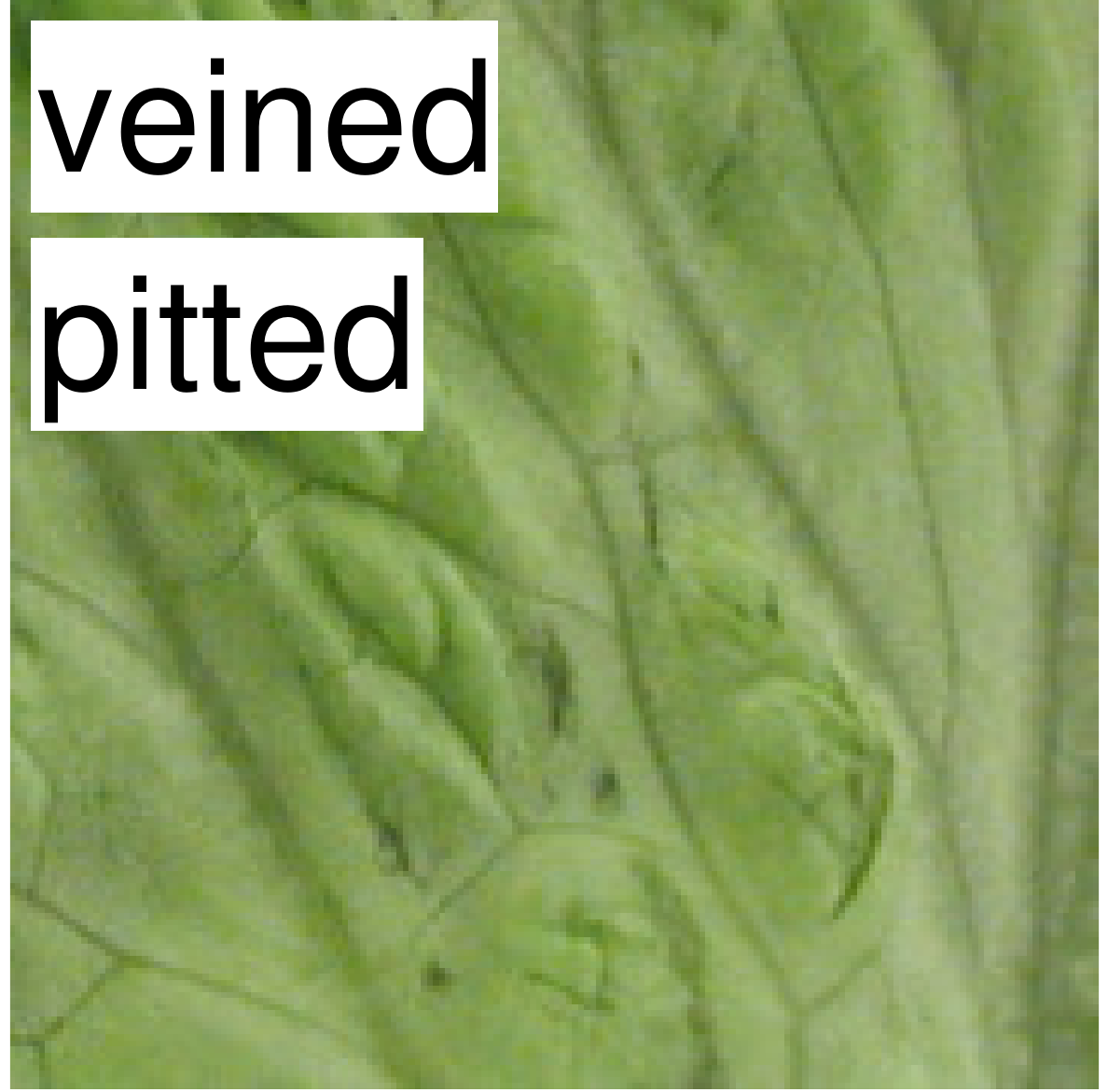}&
\incf{linen}{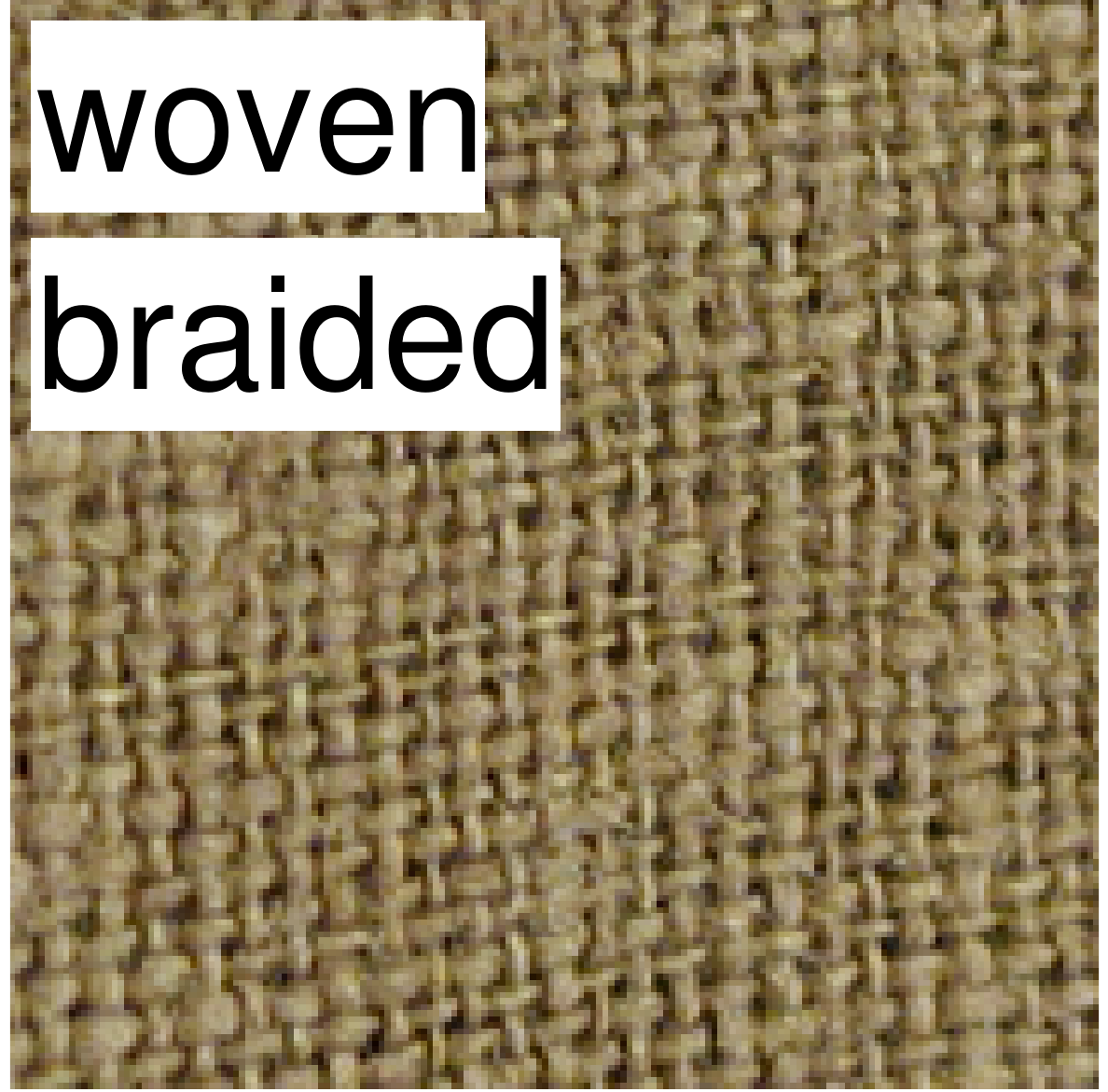}&
\incf{white bread}{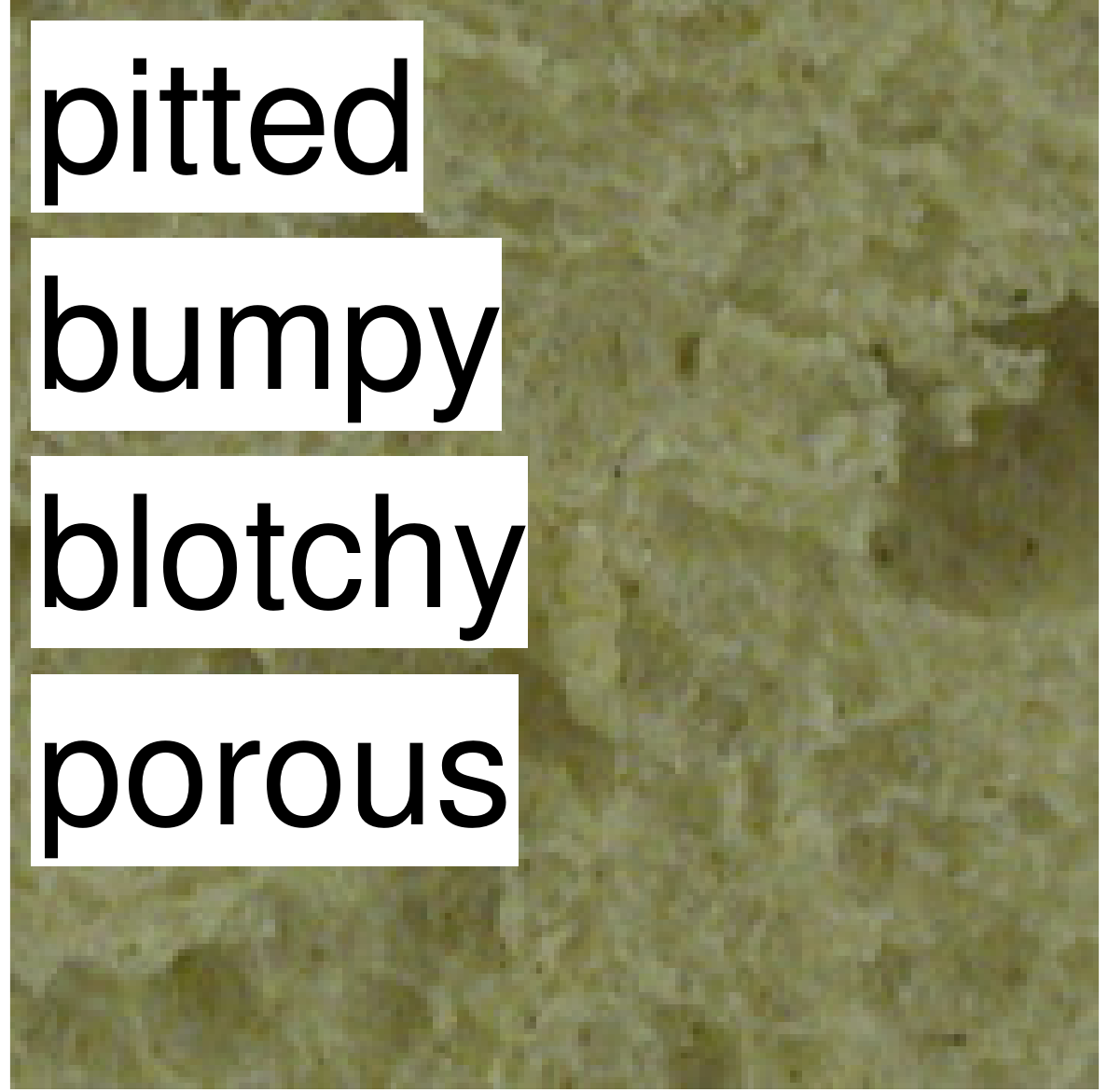}&
\incf{wood}{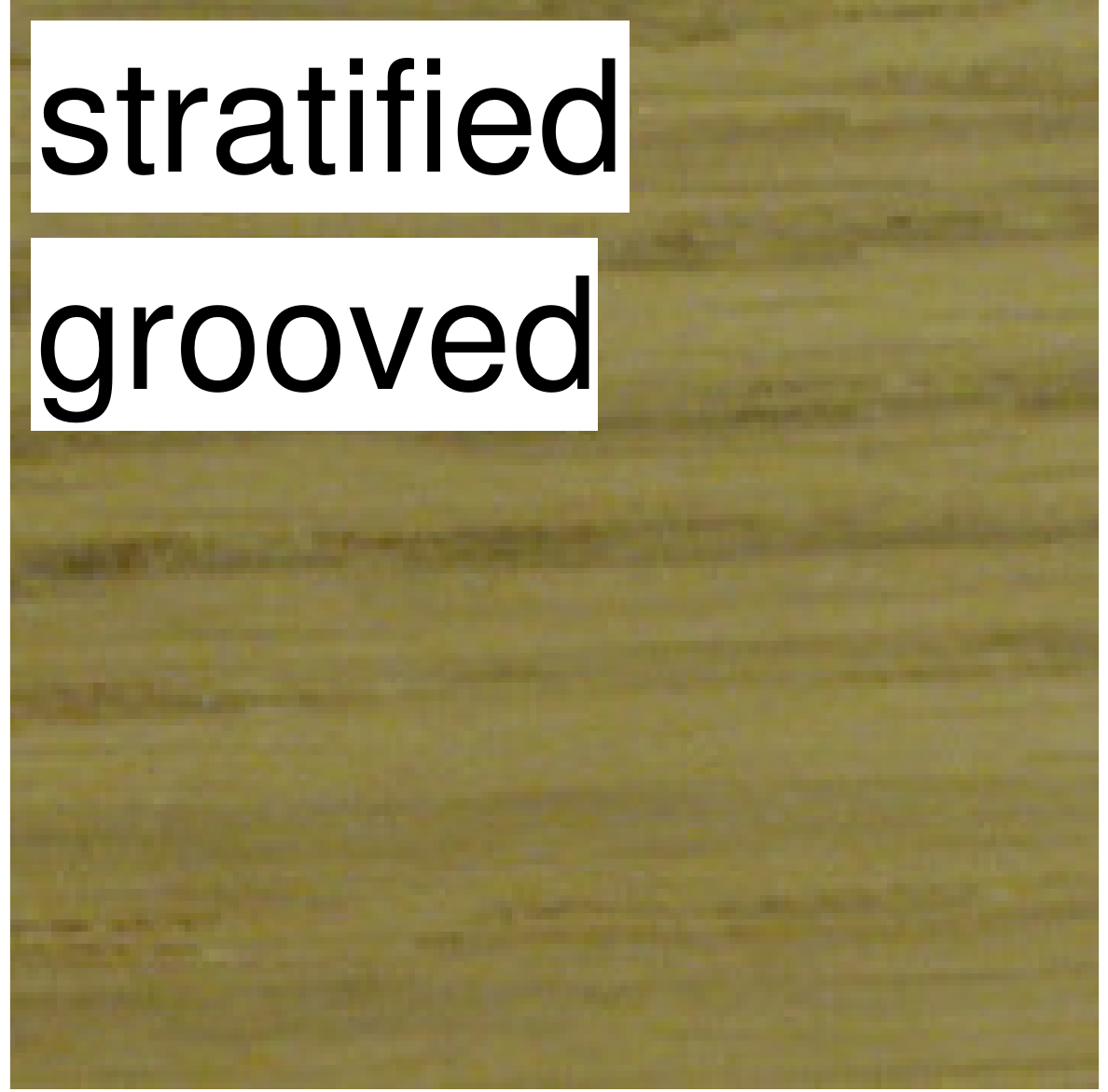}&
\incf{wool}{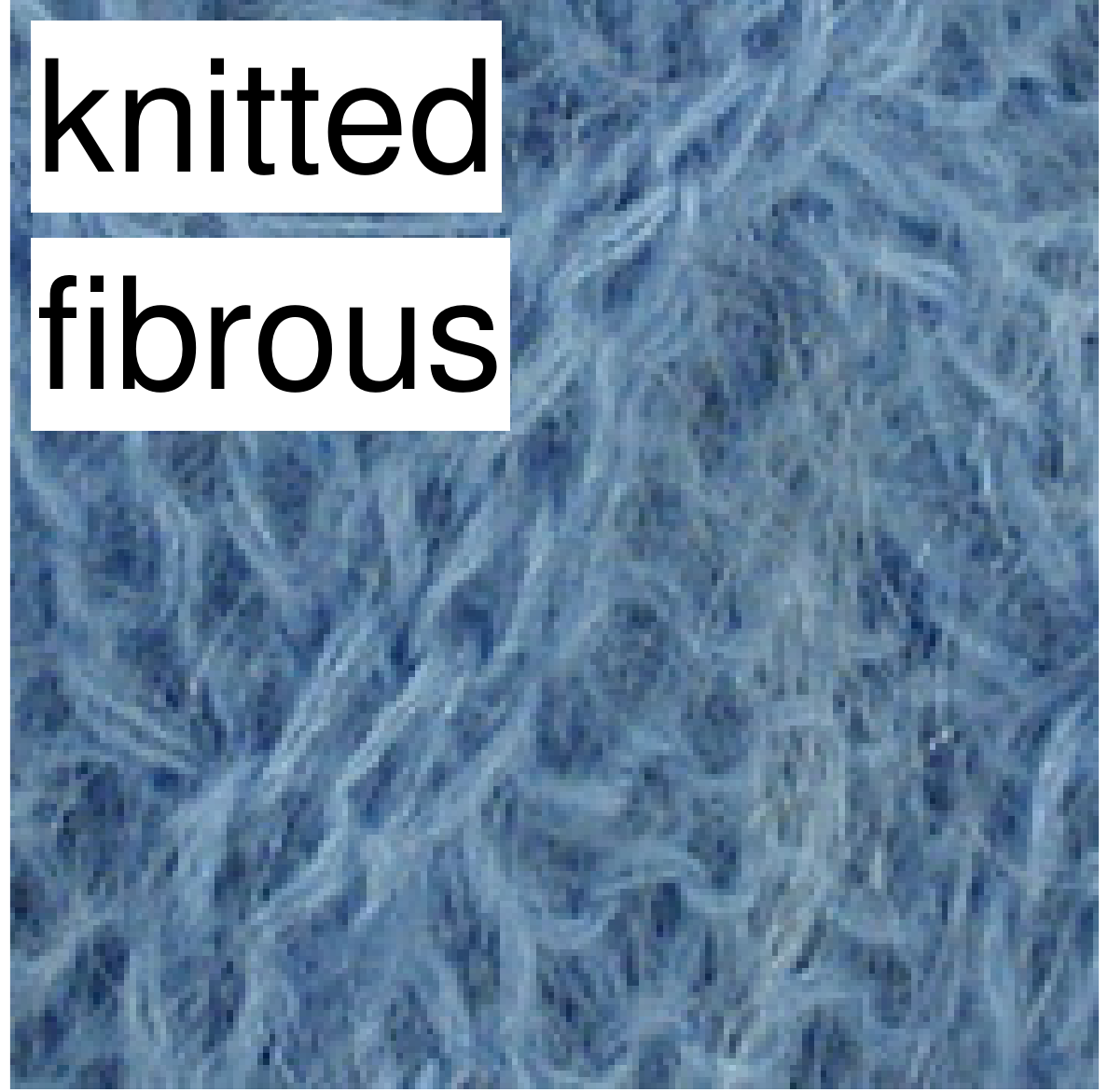}
\end{tabular}
\vspace{-1.5em}
\caption{Descriptions of materials from \kthb dataset. These words are the most frequent top scoring texture attributes (from the list of 47 we proposed), when classifying the images from the \kthb dataset. The descriptions are obtained by considering the whole material category, while a single image per material is shown for visualization.}
\vspace{-0.1in}
\label{f:kth-tips-described}
\end{figure*}

\begin{figure*}[t]
\begin{center}
\includegraphics[width=0.5\textwidth]{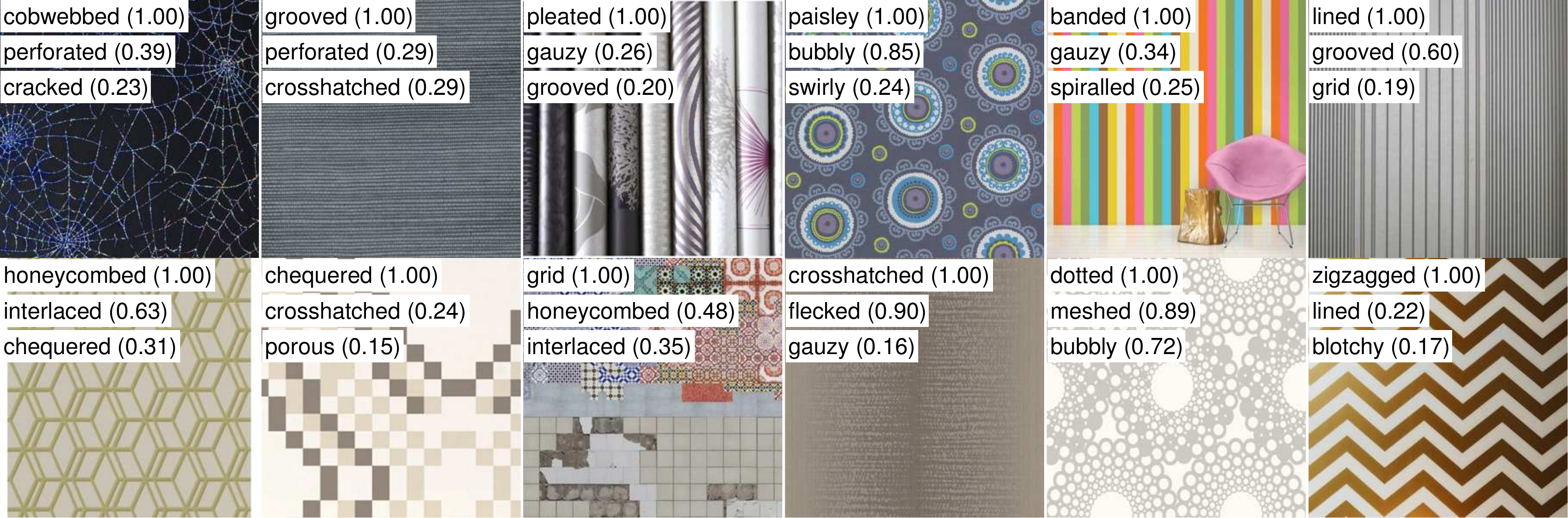}%
\includegraphics[width=0.5\textwidth]{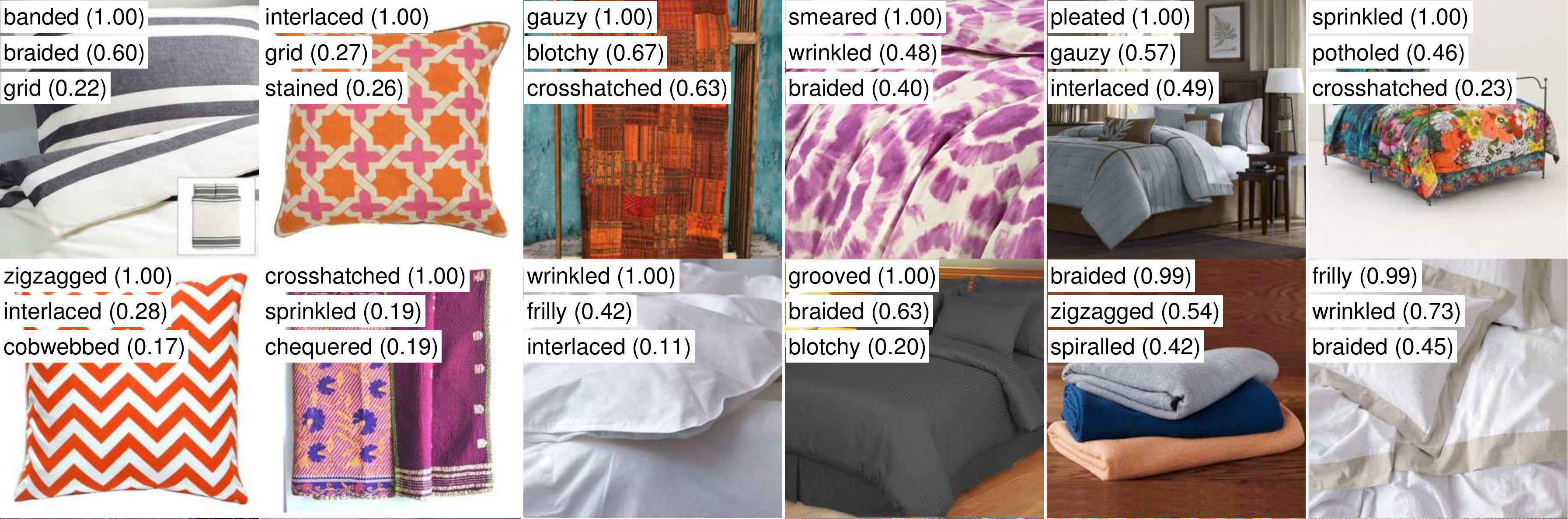}
\caption{Bedding sets (top) and wallpapers (bottom) with the top $3$ attributes predicted by our classifier and normalized classification score in brackets.}
\label{fig:example-bedding-wallpaper}
\end{center}
\end{figure*}

This section includes a short qualitative evaluation of the DTD attributes. Perhaps their most appealing property is interpretability; to verify that semantics transfer in a reasonable way across domains, Fig.~\ref{f:kth-tips-described} shows an excellent semantic correlation between the ten categories in \kthb and the attributes in DTD. For example, aluminum foil is found to be \emph{wrinkled}, while bread is found be \emph{bumpy}, \emph{pitted}, \emph{porous} and \emph{flecked}.

As an additional application of our describable texture attributes we compute them on a large dataset of 10,000 wallpapers and bedding sets from \url{houzz.com}. The 47 attribute classifiers are learned as in Sect.~\ref{s:exp-results} using the \IFVSIFT representation and then applied to the 10,000 images to predict the strength of association of each attribute and image. Classifier scores are re-calibrated on the target data and converted to probabilities by rescaling the scores to have a maximum value of one on the whole dataset. Fig.~\ref{fig:example-bedding-wallpaper} shows some example attribute predictions, selecting for each of a number of attributes an image that has a score close to 1 (excluding images used for calibrating the scores), and then including additional top two attribute matches. The top two matches tend to be a very good description of each texture or pattern, while the third is a good match in about half of the cases.


\section{Conclusions}
In this paper we have introduced a dataset of 5,640 images collected ``in the wild'' that have been jointly labelled with 47 describable texture attributes and  have used this dataset to study the problem of extracting semantic properties of textures and patterns, addressing real-world human-centric applications. We have also introduced a novel analysis of material and texture attribute recognition in a large dataset of textures in clutter derived from the excellent OpenSurfaces dataset. Finally,  we have analyzed texture representation in relation to modern deep neural networks. The main finding is that orderless pooling of convolutional neural network features is a remarkably good texture descriptor, sufficiently versatile to dub as a scene and object descriptor too and resulting in the new state-of-the-art performance in several benchmarks.

\vskip 1em
\noindent{\bf Acknowledgments.} The development of DTD is based on work done at the 2012 CLSP Summer Workshop, and was partially supported by NSF Grant \#1005411, ODNI via the JHU HLTCOE and Google Research. Mircea Cimpoi was supported by the ERC grant VisRec no\@. 228180 and the XRCE UAC grant. Iasonas Kokkinos was supported by EU Projects RECONFIG FP7-ICT-600825 and MOBOT FP7-ICT-2011-600796.

\bibliographystyle{spmpsci}
\bibliography{bibliography}
\end{document}